\documentclass[11pt, a4paper, logo, copyright]{googlecloud}

\pdftrailerid{redacted}

\makeatletter
\renewcommand\bibentry[1]{\nocitep{#1}{\frenchspacing\@nameuse{BR@r@#1\@extra@b@citeb}}}
\makeatother

\usepackage{kantlipsum, lipsum}
\usepackage{dsfont}

\definecolor{ourred}{HTML}{F19C99}
\definecolor{ourblue}{HTML}{7EA6E0}
\definecolor{plotred}{HTML}{f77189}
\definecolor{plotgreen}{HTML}{33b07a}
\definecolor{plotpurple}{HTML}{cc7af4}
\definecolor{plotmagenta}{HTML}{f565cc}
\definecolor{plotazure}{HTML}{38a9c5}

\definecolor{codegreen}{rgb}{0,0.6,0}
\definecolor{codegray}{rgb}{0.5,0.5,0.5}
\definecolor{codepurple}{rgb}{0.58,0,0.82}
\definecolor{backcolour}{rgb}{0.95,0.95,0.92}

\definecolor{Gray}{gray}{0.92}

\definecolor{stage1}{HTML}{34A853}
\definecolor{stage2}{HTML}{A680B8}
\definecolor{stage3}{HTML}{009999}

\usepackage{listings}

\lstdefinestyle{mystyle}{
    backgroundcolor=\color{backcolour},   
    commentstyle=\color{codegreen},
    keywordstyle=\color{magenta},
    numberstyle=\tiny\color{codegray},
    stringstyle=\color{codepurple},
    basicstyle=\ttfamily\scriptsize,
    breakatwhitespace=false,         
    breaklines=true,                 
    captionpos=b,                    
    keepspaces=true,                 
    numbersep=5pt,                  
    showspaces=false,                
    showstringspaces=false,
    showtabs=false,                  
    tabsize=2,
    frame=none,
    aboveskip=1pt,
    belowskip=1pt,
}
\usepackage[most]{tcolorbox}
\tcbset{enhanced,
boxrule=0pt,frame hidden,
borderline west={4pt}{0pt}{green!75!black},
top=0pt,bottom=0pt,
colback=green!10!white,
sharp corners}

\usepackage{colortbl}
\usepackage{float}

\usepackage{multirow}

\usepackage{xcolor}
\usepackage[authoryear, sort&compress, round]{natbib}

\usepackage[utf8]{inputenc} %
\usepackage[T1]{fontenc}    %
\usepackage{hyperref}       %
\hypersetup{
  colorlinks,
  citecolor=MidnightBlue,
  linkcolor=red,
  urlcolor=blue}
\usepackage{url}            %
\usepackage{booktabs}       %
\usepackage{amsfonts}       %
\usepackage{nicefrac}       %
\usepackage{microtype}      %
\usepackage{wrapfig} %
\usepackage{graphicx} %
\usepackage{soul} %
\usepackage{enumitem}
\usepackage{amssymb,amsmath}
\usepackage{tikz, adjustbox}
\usepackage[most]{tcolorbox}
\usepackage{subcaption}
\usepackage{multirow}
\usepackage{arydshln}
\usepackage{float}
\usepackage{algorithm}
\usepackage{algorithmic}
\usepackage[capitalize,noabbrev]{cleveref}
\usepackage{fancyvrb}
\usepackage{float}
\usepackage{tcolorbox}

\usepackage{amsmath,amsfonts,bm}

\def\eqref#1{equation~\ref{#1}}

\def\1{\bm{1}}

\DeclareMathAlphabet{\mathsfit}{\encodingdefault}{\sfdefault}{m}{sl}
\SetMathAlphabet{\mathsfit}{bold}{\encodingdefault}{\sfdefault}{bx}{n}

\lstdefinestyle{mystyle}{
    backgroundcolor=\color{backcolour},   
    commentstyle=\color{codegreen},
    keywordstyle=\color{magenta},
    numberstyle=\tiny\color{codegray},
    stringstyle=\color{codepurple},
    basicstyle=\ttfamily\scriptsize,
    breakatwhitespace=false,         
    breaklines=true,                 
    captionpos=b,                    
    keepspaces=true,                 
    numbers=left,                    
    numbersep=5pt,                  
    showspaces=false,                
    showstringspaces=false,
    showtabs=false,                  
    tabsize=2,
    frame=none,
    aboveskip=1pt,
    belowskip=1pt,
}
\lstdefinestyle{plainins}{
    backgroundcolor=\color{white},   
    commentstyle=\color{codegreen},
    keywordstyle=\color{magenta},
    numberstyle=\tiny\color{codegray},
    stringstyle=\color{codepurple},
    basicstyle=\ttfamily\scriptsize,
    breakatwhitespace=false,         
    breaklines=true,                 
    captionpos=b,                    
    keepspaces=true,                 
    numbers=none,                    
    numbersep=5pt,                  
    showspaces=false,                
    showstringspaces=false,
    showtabs=false,                  
    tabsize=2,
    aboveskip=0pt,
    belowskip=0pt,
    frame=single
}
\lstdefinestyle{plainexam}{
    backgroundcolor=\color[HTML]{FFFCF3},   
    commentstyle=\color{codegreen},
    keywordstyle=\color{magenta},
    numberstyle=\tiny\color{codegray},
    stringstyle=\color{codepurple},
    basicstyle=\ttfamily\scriptsize,
    breakatwhitespace=false,         
    breaklines=true,                 
    captionpos=b,                    
    keepspaces=true,                 
    numbers=none,                    
    numbersep=5pt,                  
    showspaces=false,                
    showstringspaces=false,
    showtabs=false,                  
    tabsize=2,
    aboveskip=0pt,
    belowskip=0pt
}

\title{On the Role of Feedback in Test-Time Scaling of Agentic~AI~Workflows}

\correspondingauthor{schakra3@umd.edu, \{yiwensong,hamidpalangi\}@google.com}

\renewcommand{\today}{}

\author[1 2 **]{Souradip Chakraborty}
\author[1]{Mohammadreza Pourreza}
\author[1]{Ruoxi Sun}
\author[1]{Yiwen Song}
\author[1]{Nino Scherrer}
\author[2]{\\Furong Huang}
\author[3]{Amrit Singh Bedi}
\author[1]{Ahmad Beirami}
\author[1]{Jindong Gu}
\author[1 *]{Hamid Palangi}
\author[1 *]{Tomas Pfister}

\affil[1]{Google}
\affil[2]{University of Maryland}
\affil[3]{University of Central Florida}

\begin{abstract}
  Agentic AI workflows (systems that autonomously plan and act) are becoming widespread, yet their task-success rate on complex tasks remains low. A promising solution is \textit{inference-time alignment}, which uses extra compute at test time to improve performance. Inference-time alignment relies on three components: \textit{sampling}, \textit{evaluation}, and \textit{feedback}. While most prior work studies sampling and automatic evaluation, feedback remains under-explored. To study the role of feedback, we introduce \textbf{Iterative Agent Decoding (IAD)}, a procedure that repeatedly inserts feedback extracted from different forms of critiques (reward models or AI-generated textual feedback) between decoding steps. Through IAD, we analyze feedback along four dimensions: (1) its role in the accuracy–compute trade-offs with limited inference budget, (2) quantifying the gains over diversity-only baselines such as best-of-\textit{N} sampling, (3) effectiveness of composing feedback from reward models versus textual critique, and (4) robustness to noisy or low-quality feedback. Across Sketch2Code, Text2SQL, Intercode, and WebShop, we show that IAD with proper integration of high fidelity feedback leads to consistent gains up to 10\% absolute performance improvement over various baselines such as best-of-N. Our findings underscore feedback as a crucial knob for inference-time alignment of agentic AI workflows with limited inference budget.
\end{abstract}

\begin{document}

\maketitle

\section{Introduction}

\begin{wrapfigure}[10]{r}{0.45\textwidth}   
    \vspace{-16pt}                       
    \centering
    \includegraphics[width=0.9\linewidth]{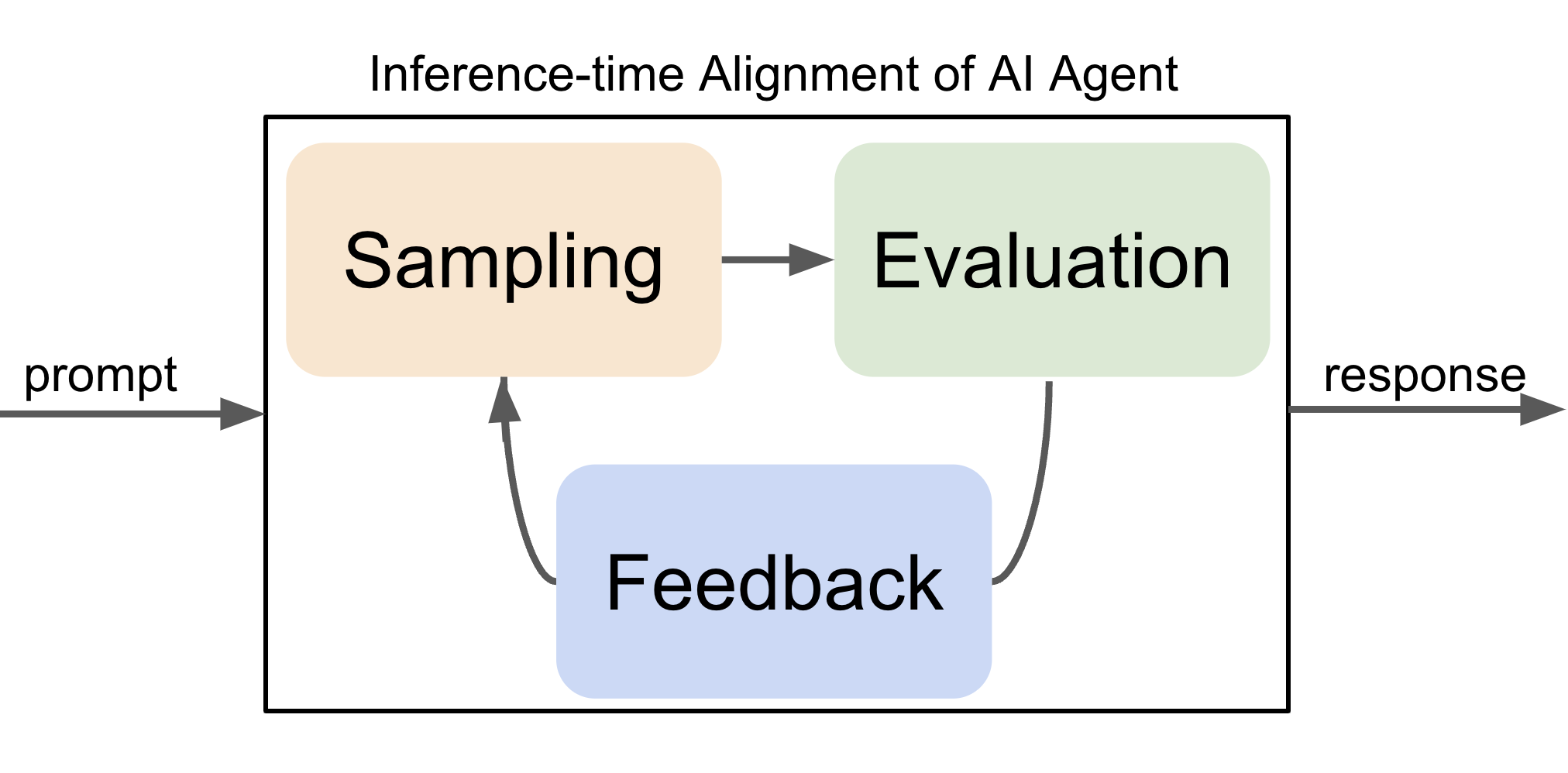}
    \vspace{-6pt}                       %
    \caption{General pipeline for inference-alignment approaches for AI agents.}
    \label{fig:inf_bb}
    \vspace{-6pt}                       
\end{wrapfigure}
AI agents demonstrate remarkable potential across a wide range of applications, yet they still face challenges in complex tasks requiring strategic planning and multimodal reasoning. For example, on challenging benchmarks such as {Sketch2Code} \citep{li2024sketch2codeevaluatingvisionlanguagemodels}, {Text2SQL} \citep{li2024can}, {Intercode} \citep{intercode}, and {Webshop} \citep{yao2023webshopscalablerealworldweb}, even state-of-the-art agents can only achieve 20–30\% accuracy. While post-training via SFT and RL \citep{rlhf1, rlhf2, rlhf3} has been an effective tool to enhance the capabilities of generative models; however, it is not directly applicable to AI agents. This is because AI agents frequently operate as black-box systems with no access to the internals~\citep{agent2, agent1}.  Hence, we focus on inference-time approaches which are (i) API-friendly (applicable to closed-source models accessible via commercial APIs) and (ii) compatible with various underlying models.

\noindent At inference time, alignment strategies broadly iterate over three components: \textit{sampling}, \textit{evaluation}, and \textit{feedback} as illustrated in Figure \ref{fig:inf_bb}. A typical inference-time iteration involves sampling one or more outcomes from the generative model, evaluating them using a judge/reward, and then generating feedback to improve subsequent outcomes. The role of \textit{sampling} and \textit{evaluation} is well-studied in literature with efficient and scalable algorithms based on best-of-N (BoN) approaches \citep{nakano2021webgpt,bon3}. BoN-based approaches operate by sampling multiple responses and selecting the best according to a verifier, effectively leveraging the sampling and evaluation steps.  While BoN and variants \citep{nakano2021webgpt, bon3, verdun2025soft} are highly parallelizable, allowing responses to be sampled and evaluated concurrently to reduce wall-clock latency, it may not be compute-optimal. That is, to achieve a desirable outcome, BoN may require generating a large number of samples and evaluating each independently. This parallelism trades off compute efficiency for lower latency. Therefore, when compute is constrained, simply generating more samples is not a viable path to better performance, highlighting the need for a more adaptive approach. More importantly, a key limitation of BoN approaches is their inability to incorporate feedback mechanisms for iterative refinement.  

\noindent To address the above-mentioned limitations, recently, several sequential approaches have emerged \citep{refine1, tryagain, selfdebug, robeyns2025selfimprovingcodingagent, jiang2024selfplanningcodegenerationlarge} that aim to explicitly integrate feedback for iterative refinement. This trend towards self-correction is evident in techniques such as self-refine \citep{refine1}, where models iteratively improve their outputs based on self-critique, and try-again strategies \citep{tryagain} that prompt the LLM to generate alternative responses when initial attempts fail. Other approaches include \citep{selfdebug}, where LLMs identify and correct errors in their own generated code. Additionally, recent advancements in agentic systems have showcased their ability to self-improve \citep{robeyns2025selfimprovingcodingagent, jiang2024selfplanningcodegenerationlarge}.

\begin{figure*}[t]
    \centering
    \includegraphics[width=0.9\textwidth]{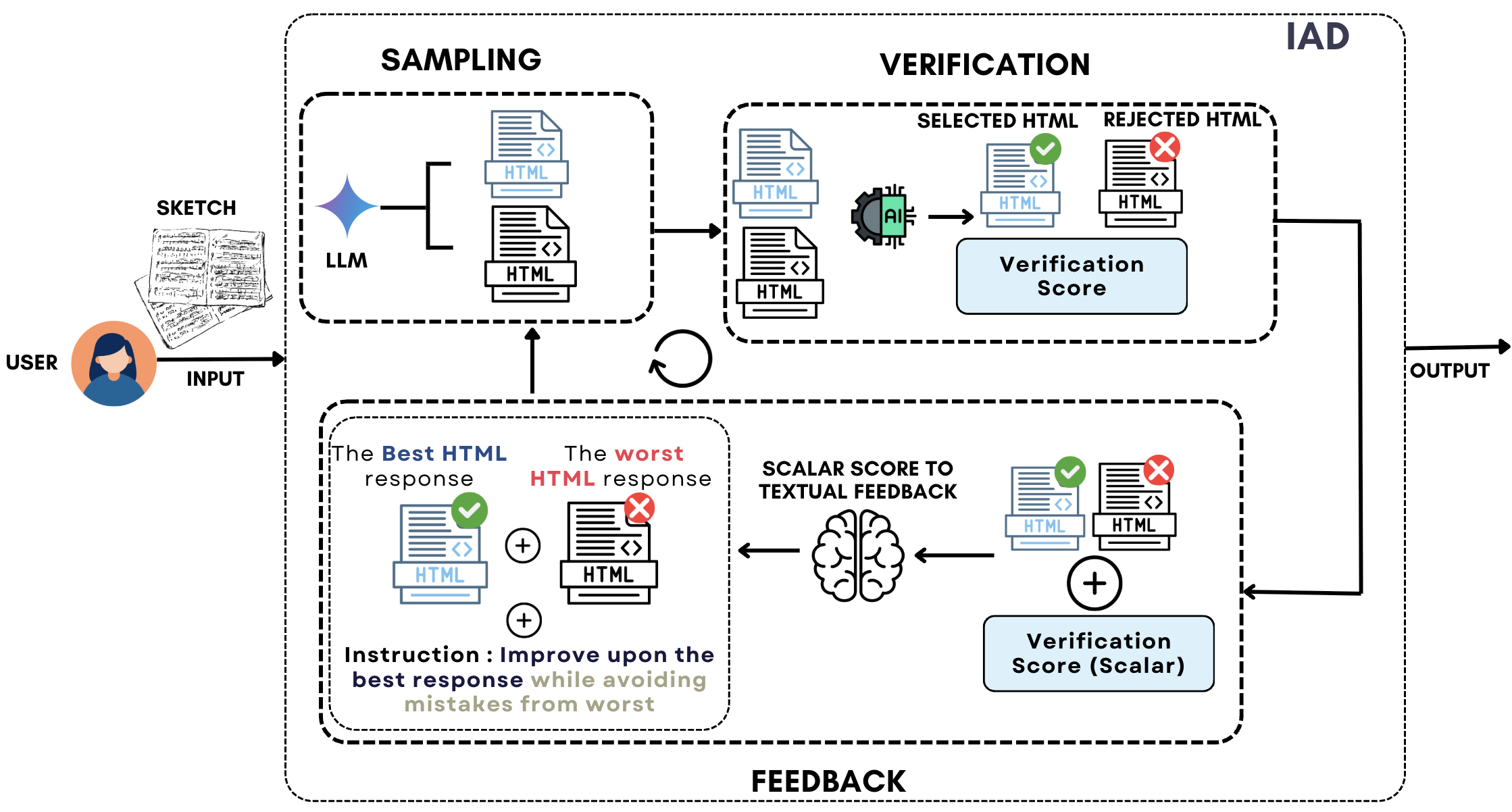}
     \caption{Overview of Iterative Agent Decoding (IAD), highlighting the sampling, evaluation, and feedback components of inference-time alignment. The figure illustrates how feedback is constructed from verification signals and used to iteratively refine model outputs.} 
    \label{fig:overall}
\end{figure*}

\noindent While existing methods creatively use feedback, they do not systematically study it. They often lack a focused analysis of how feedback should be designed, integrated, or optimized. While prior approaches \citep{selfdebug, tryagain, refine1} demonstrate the utility of adaptive learning with feedback, a clear understanding of the mechanisms by which LLMs learn from this feedback, especially with scalar feedback or score, is still developing. As a result, the role of feedback in inference-time alignment, especially in black-box agentic environments, remains under-examined. We aim to address this gap by focusing on feedback and systematically understanding its impact on test-time scaling of agentic workflows. Specifically, we focus on inference-time alignment strategies designed to maximize performance under a fixed compute budget, where compute is defined concretely as the total number of API calls or forward passes through the model. In such compute-constrained settings, incorporating feedback becomes critical for effective test-time scaling. Thus, while BoN strikes a balance between latency and compute, approaches that leverage feedback are better suited when compute is the principal focus.

\noindent \textbf{Our contributions (Understanding the role of feedback).}  To understand the role and impact of feedback on inference-time alignment, we require a unified framework that flexibly integrates diverse forms of feedback, guided by a verifier, without relying heavily on prompt engineering, which can be costly and fragile. To this end, we introduce Iterative Agent Decoding (IAD), a sequential framework designed to incorporate various forms of feedback to understand the role of feedback in inference-time alignment of agents. Through IAD we uncover several intriguing results, detailed below.

\begin{figure*}[t]
    \centering
    \begin{subfigure}[b]{0.329\textwidth} 
        \includegraphics[width=\textwidth]{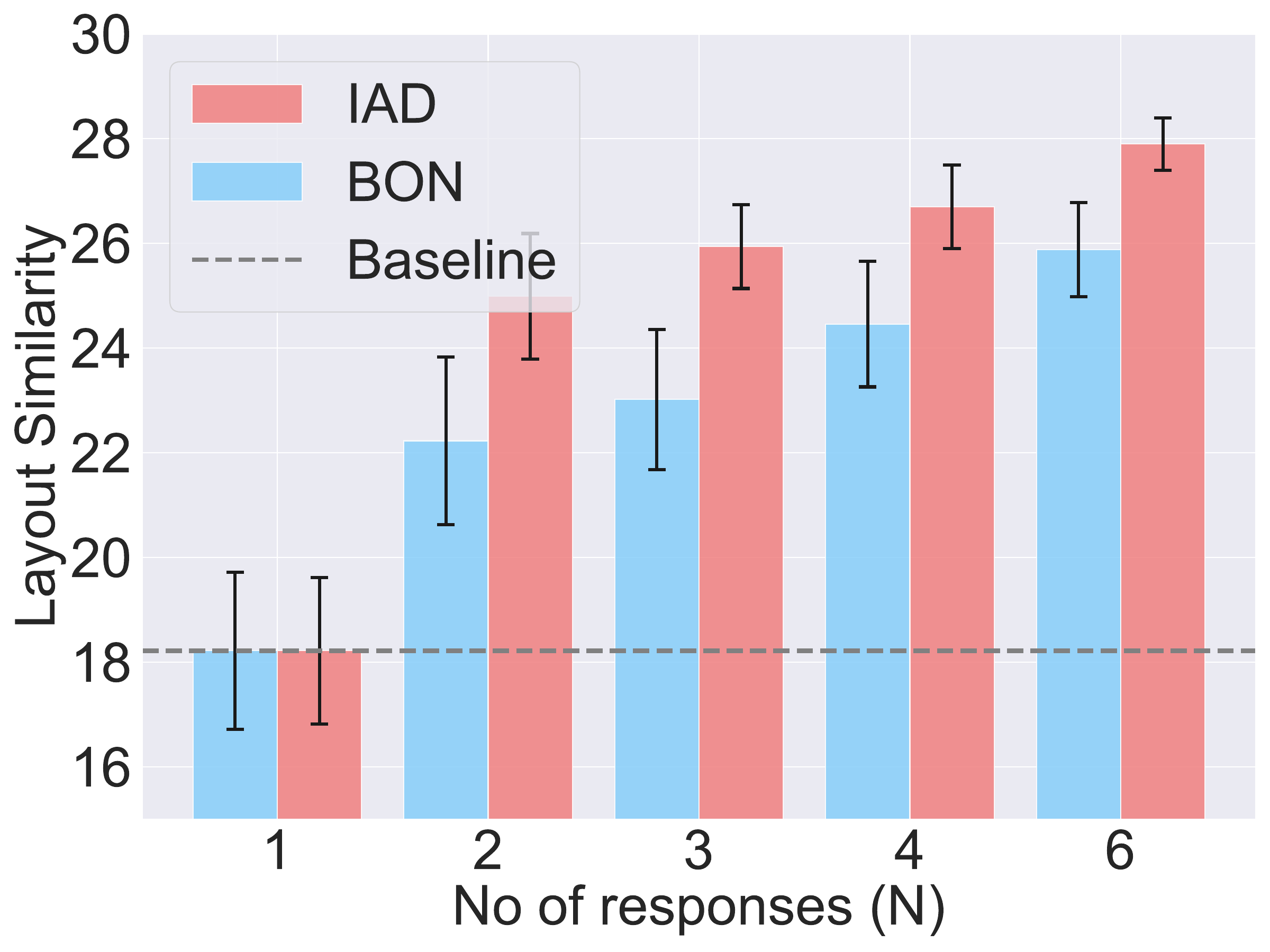} 
        \caption{\centering}
    \end{subfigure}
    \hfill 
    \begin{subfigure}[b]{0.329\textwidth} 
        \includegraphics[width=\textwidth]{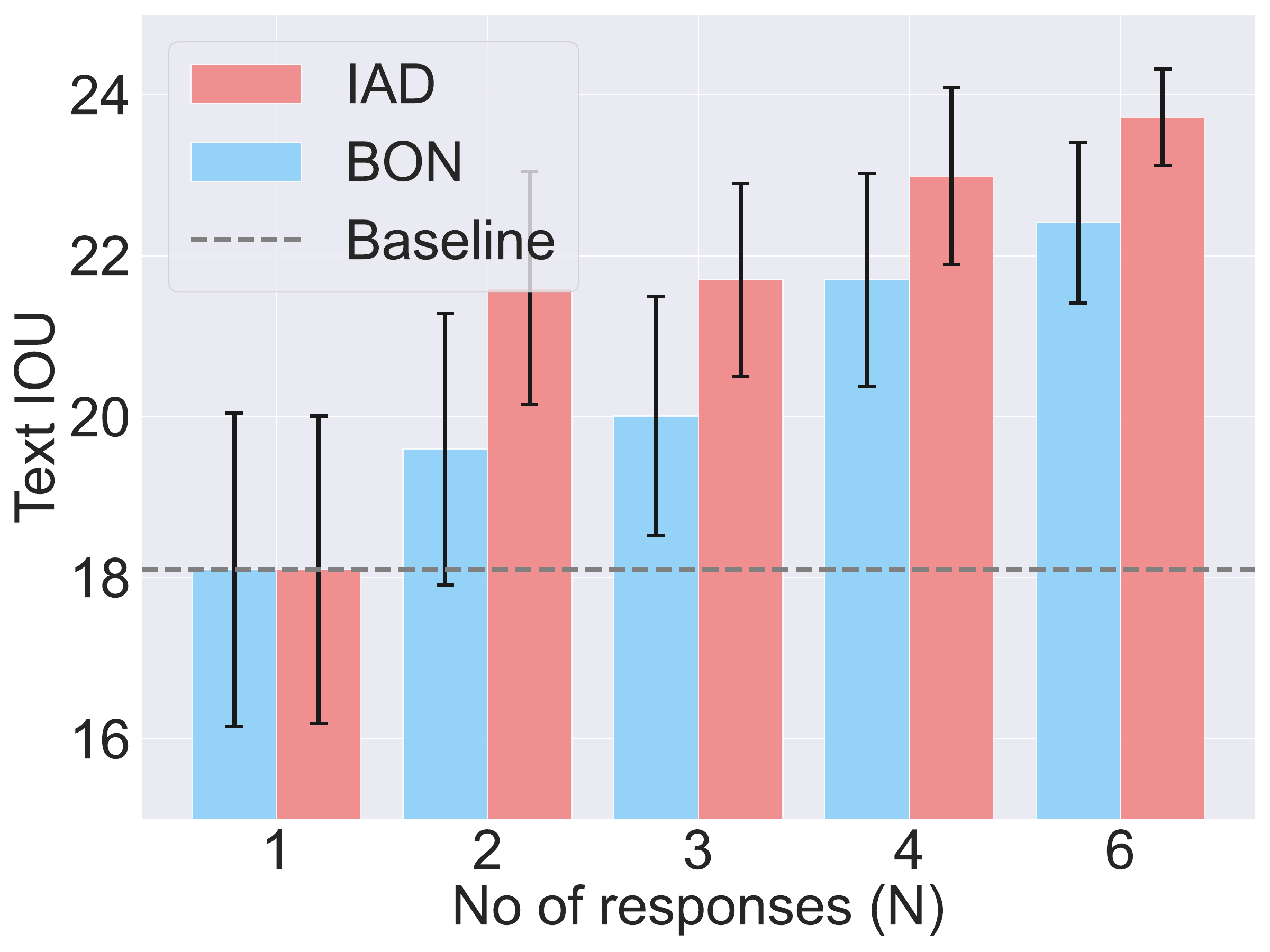} 
        \caption{\centering}
    \end{subfigure}
    \begin{subfigure}[b]{0.329\textwidth} 
        \includegraphics[width=\textwidth]{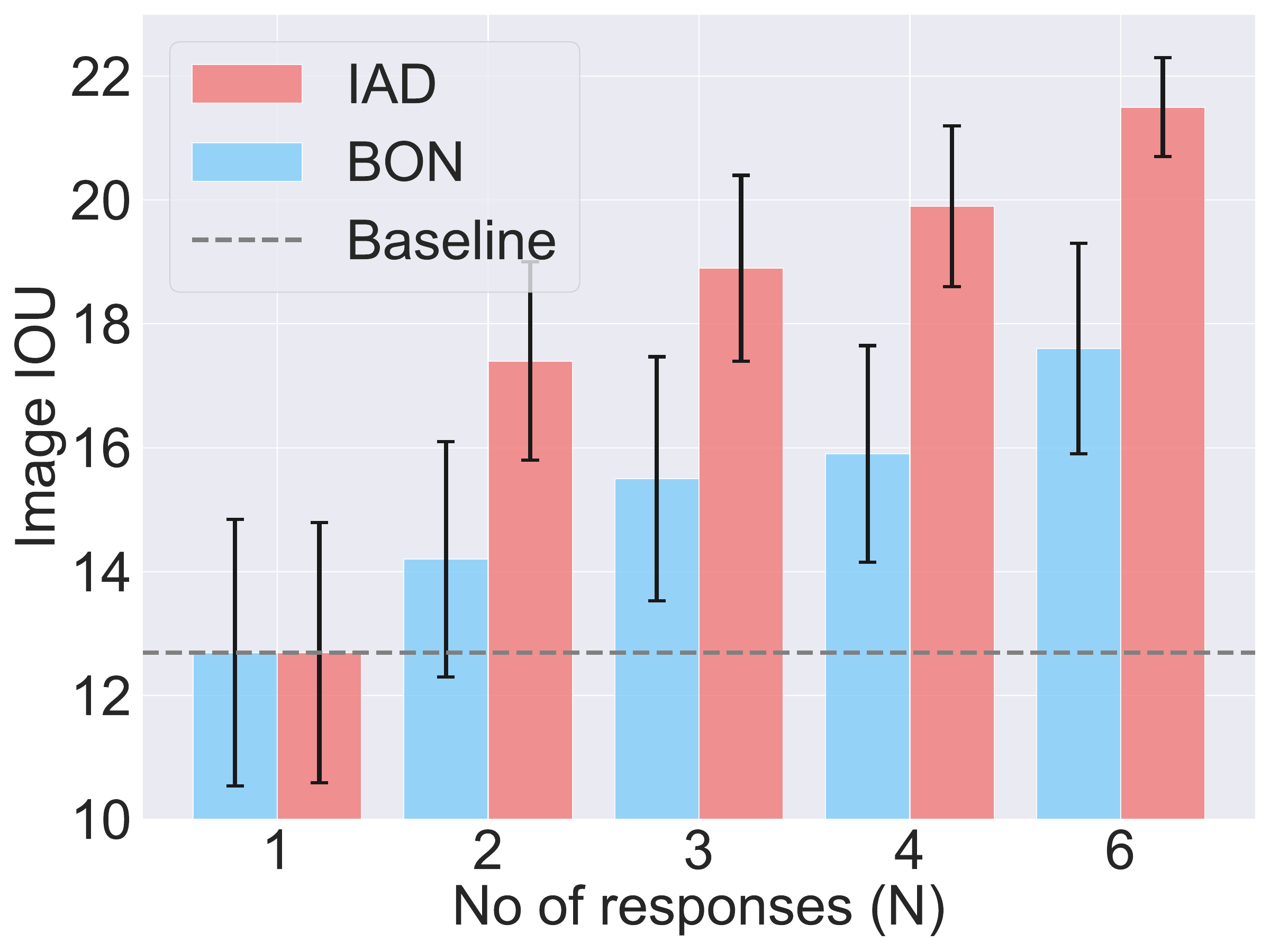} 
        \caption{\centering}
    \end{subfigure}
    \begin{subfigure}[b]{0.329\textwidth} 
        \includegraphics[width=\textwidth]{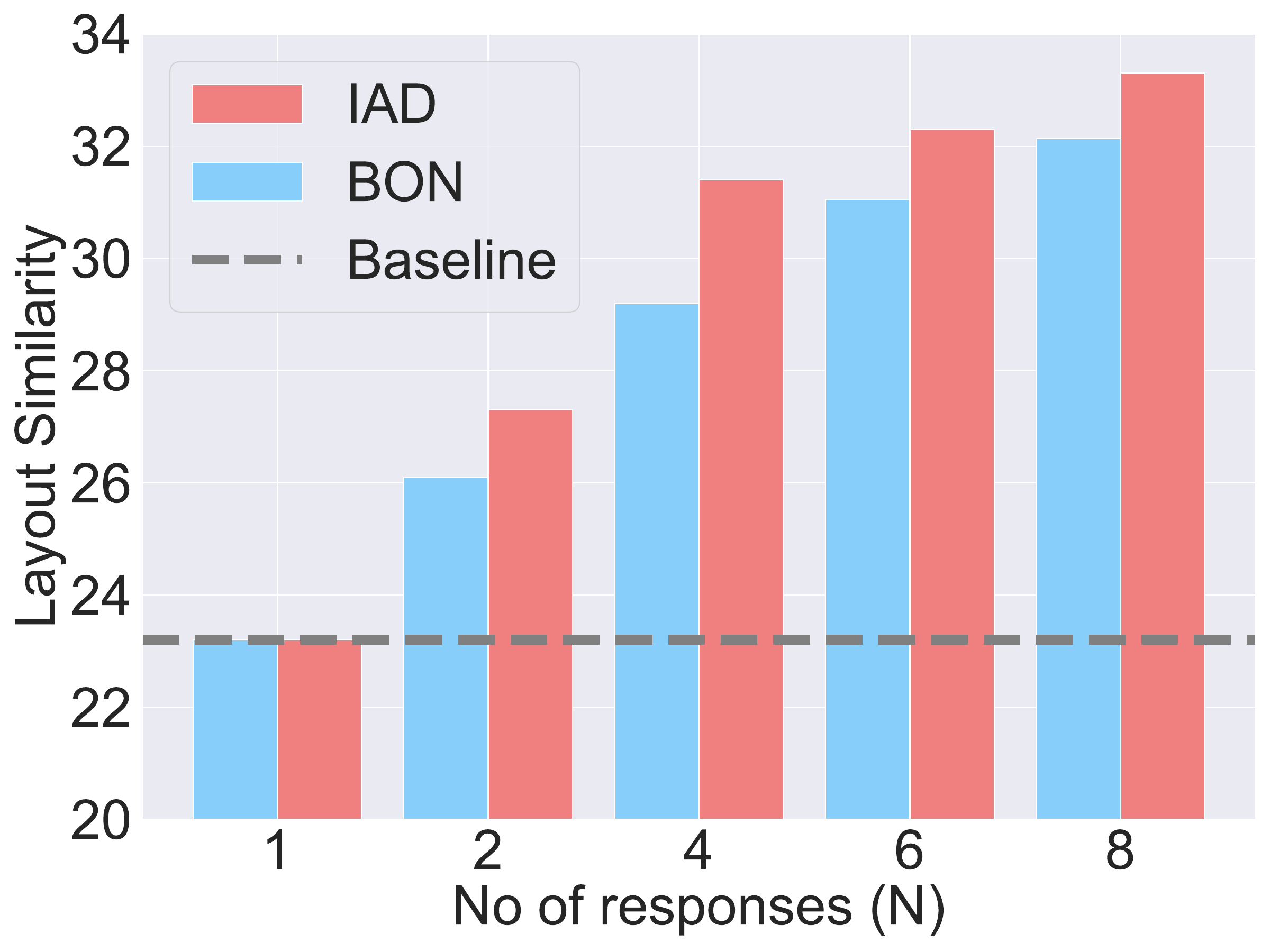} 
        \caption{\centering}
    \end{subfigure}
    \hfill 
    \begin{subfigure}[b]{0.329\textwidth} 
        \includegraphics[width=\textwidth]{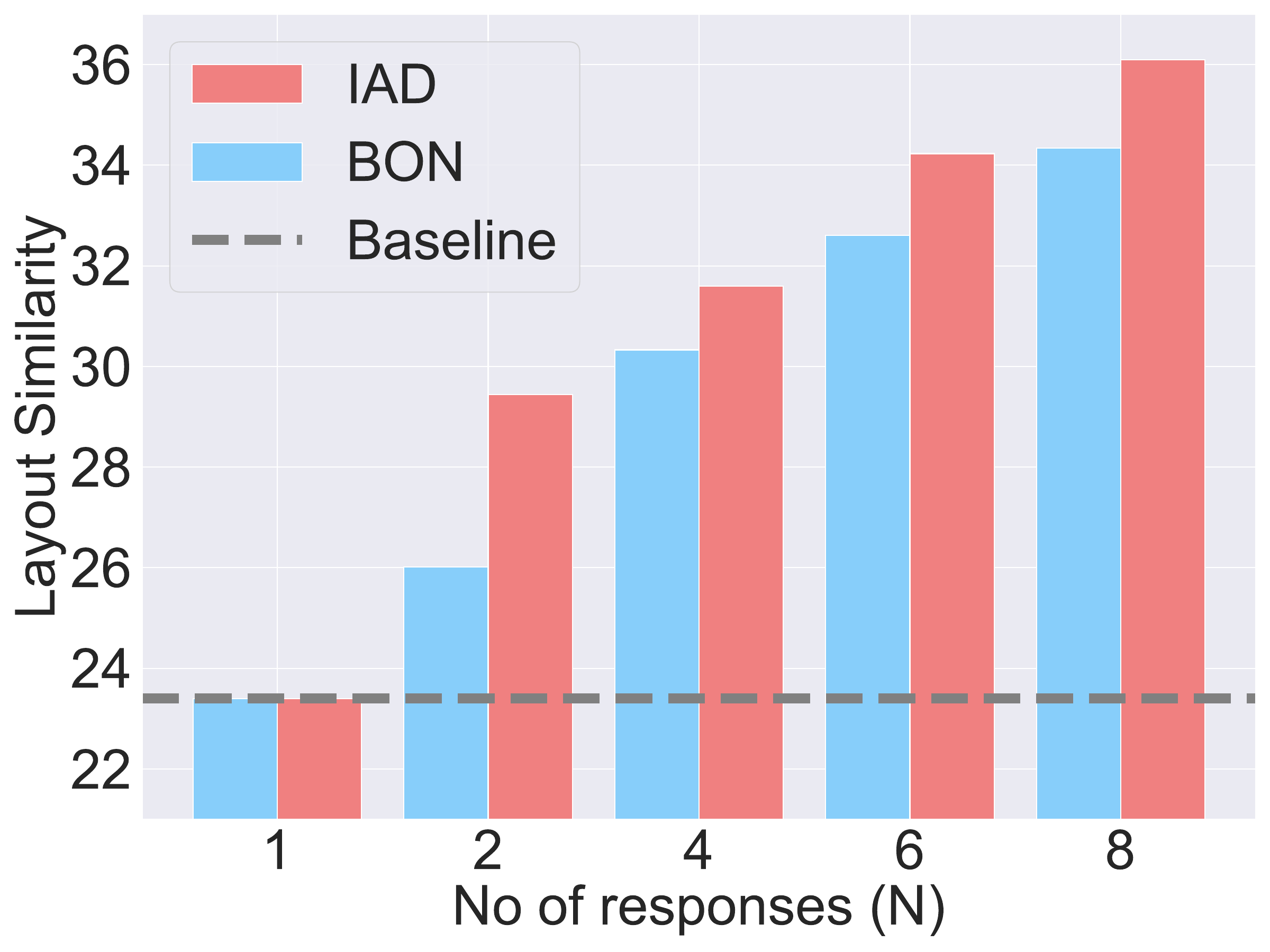} 
        \caption{\centering}
    \end{subfigure}
    \hfill 
    \begin{subfigure}[b]{0.329\textwidth} 
        \includegraphics[width=\textwidth]{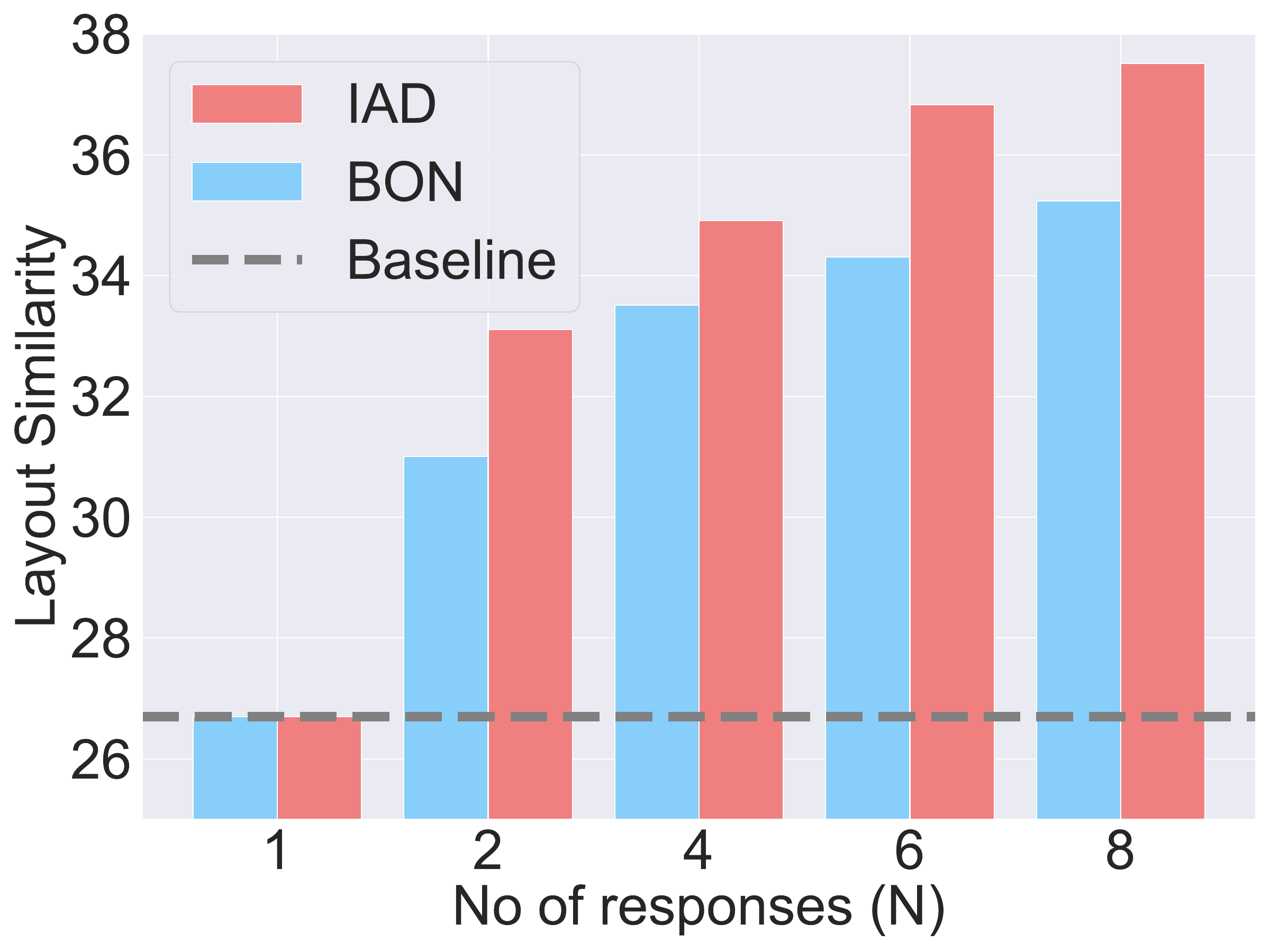} 
        \caption{\centering}
        \label{fig:subfig1}
    \end{subfigure}
    \hfill 
\vspace{-.25in}    \caption{\textbf{Test-time Scaling with Feedback in Sketch2code}: This figure provides an accuracy vs compute performance comparison of IAD (ours)  against Best-of-N sampling (SoTA) and single-turn generation with Gemini-1.5-Pro w.r.t metrics - (a) Layout Similarity (b) TextIoU (c) ImageIoU across varying the number of generations (N). The figure demonstrates IAD outperforms BoN consistently across N, but as N increases, the gap reduces. Fig (d, e, f) provides a comparison with improved model capabilities from \textbf{Gemini-2.0-Flash, Gemini-2.5-Flash, Gemini-2.5-Pro} respectively, shows that the benefit of feedback-based refinement diminishes with more capable models.}
    \label{fig:mainfig_sk2code}
    \vspace{-.15in}
\end{figure*}

\noindent \textbf{1. Accuracy vs compute: budget-constrained scaling with feedback:} We demonstrate that feedback plays a critical role in budget-constrained settings and helps achieve up to a 10\% accuracy gain over strong feedback-free baselines with the same budget. However, this margin narrows with increased budget, which allows for the generation of more candidates (e.g., larger $N$ in BoN) or using more capable models like Gemini-1.5 to Gemini-2.0 and Gemini-2.5 (cf. Figure \ref{fig:mainfig_sk2code}).

\noindent \textbf{2. Impact of adaptive feedback beyond sampling gain:} We perform controlled experiments to capture the gains from feedback beyond the diversity gains from sampling, and demonstrate that the gains in IAD arise from adaptive, feedback-guided refinement, yielding performance improvements of 4–8\% in both Sketch2Code and Text2SQL. This also highlights that feedback plays a crucial role in agentic tasks and scenarios where sampling diversity is low.

\noindent \textbf{3. Design and role of feedback form:} We investigate the impact of different forms of feedback, primarily scalar and textual on inference-time alignment. Textual feedback is relatively straightforward to integrate, as it can be directly incorporated into the model prompt. However, designing effective feedback from scalar rewards and preferences remains underexplored. In IAD, we focus on methods to extract useful signals from scalar rewards and demonstrate that the performance gains crucially depend on the design of proper feedback. Our experiments show that transforming scalar feedback into directional prompts yields gains of up to 6–7\% over baselines in Sketch2Code and Text2SQL.

\noindent \textbf{4. Sensitivity to feedback quality:} We study the sensitivity and robustness of inference alignment to feedback quality via controlled sparsity and noise in feedback signals. We see that IAD remains effective under moderate degradation, but its performance declines with increasing noise. Under high sparsity or heavy noise, the performance gains over sampling-based methods drop by approximately 4–5\%, highlighting the importance of feedback fidelity for reliable test-time scaling.

\section{Inference-time Agent Alignment}

 In the black box agentic settings, we only have access to a {reference policy} \(\pi_0(\cdot|x)\), whose internals are inaccessible. The core challenge is to optimize inference-time decisions such that the generated response from \(\pi_0(\cdot|x)\) is better aligned with the outputs of optimal policy \(\pi^*(\cdot|x)\). We assume access to a {verifier function} \(R(x, y)\), which evaluates the quality of a response \(y\) (further details in Appendix).

\subsection{Iterative Agent Decoding} 

\noindent
To study the role of feedback, we introduce Iterative Agent Decoding (IAD), a general sequential framework designed to incorporate various feedback forms for understanding the role of feedback in inference-time alignment of agents.

\begin{figure*}[t]
    \centering
    \begin{subfigure}[b]{0.329\textwidth}  
        \centering
        \scalebox{1.0}{\includegraphics[width=\textwidth]{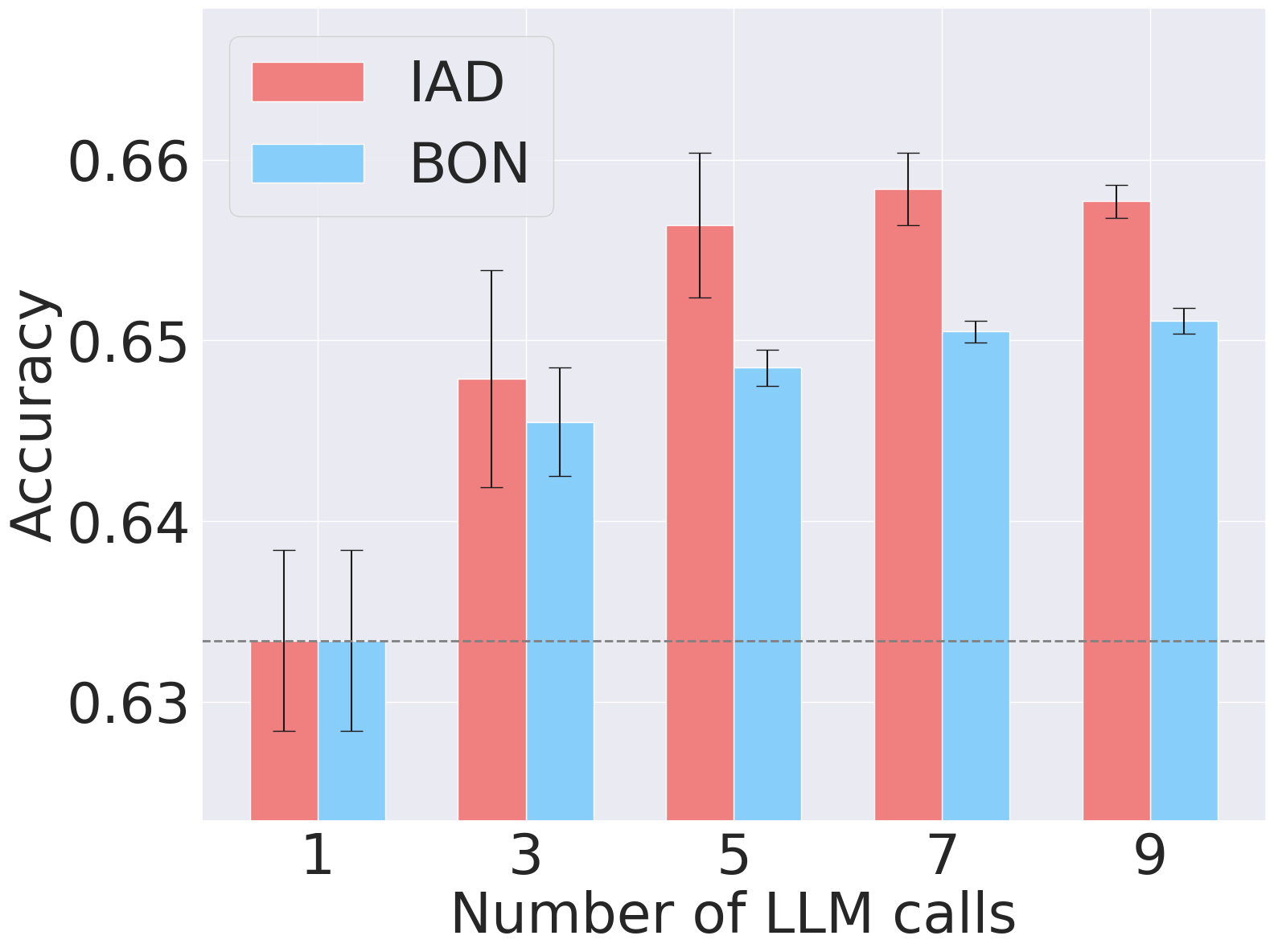}}
        \caption{\centering}
    \end{subfigure}
    \begin{subfigure}[b]{0.329\textwidth}  
        \centering
        \scalebox{1.0}{\includegraphics[width=\textwidth]{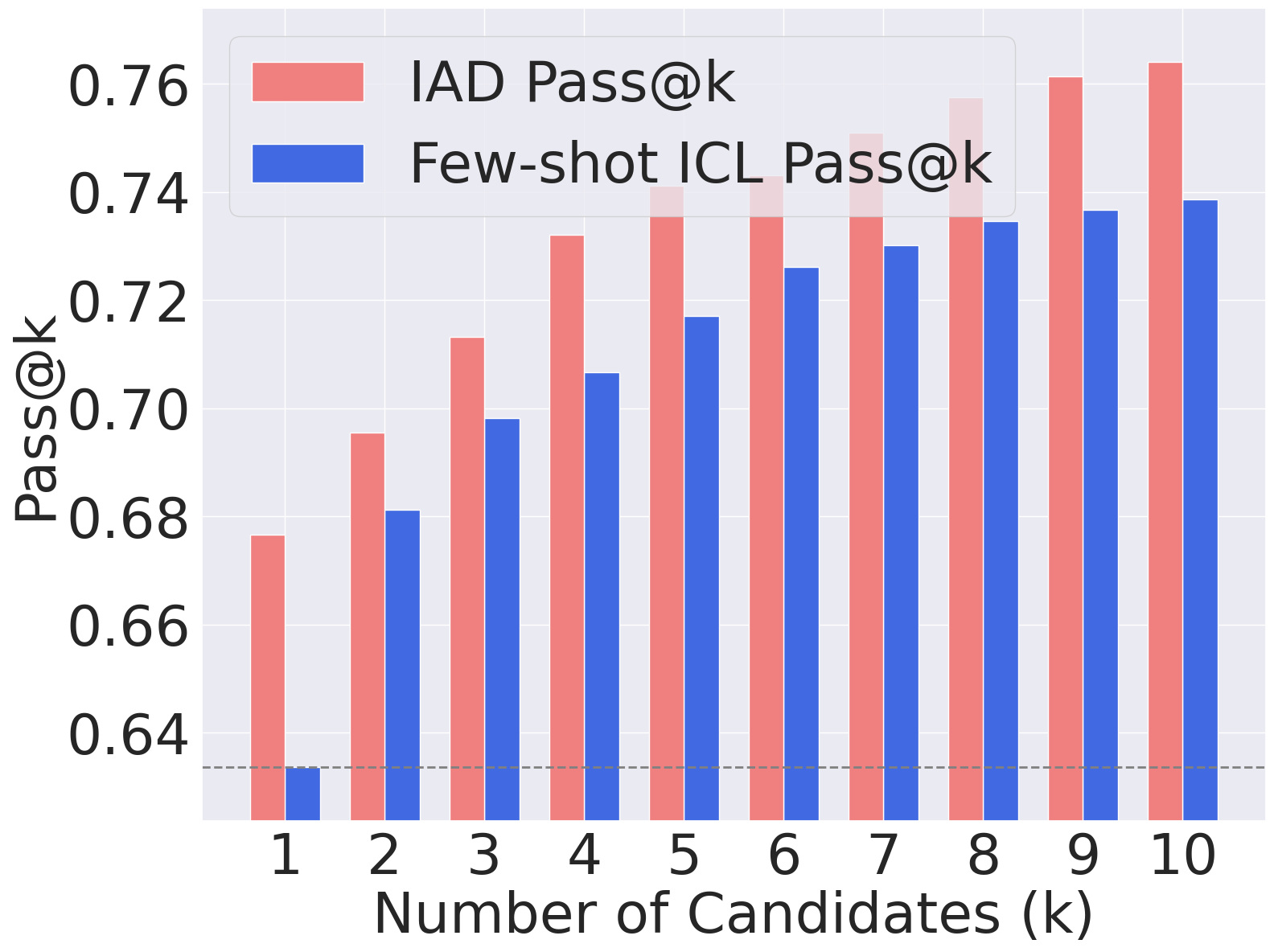}}
        \caption{\centering}
    \end{subfigure}
    \begin{subfigure}[b]{0.329\textwidth}  
        \centering
        \scalebox{1.0}{\includegraphics[width=\textwidth]{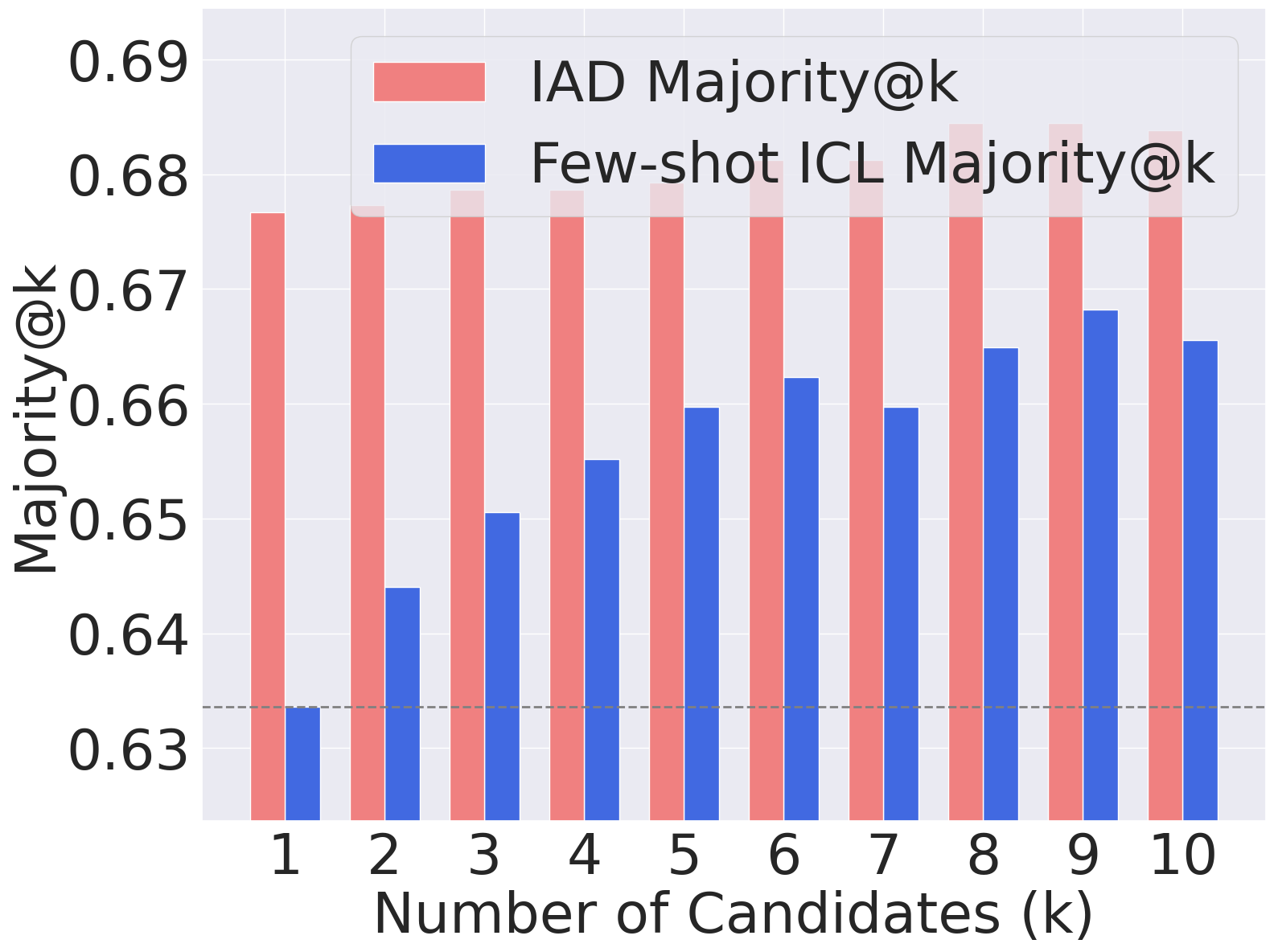}}
        \caption{\centering}
    \end{subfigure}
    
     \begin{subfigure}[b]{0.329\textwidth}  
        \centering
        \scalebox{1.0}{\includegraphics[width=\textwidth]{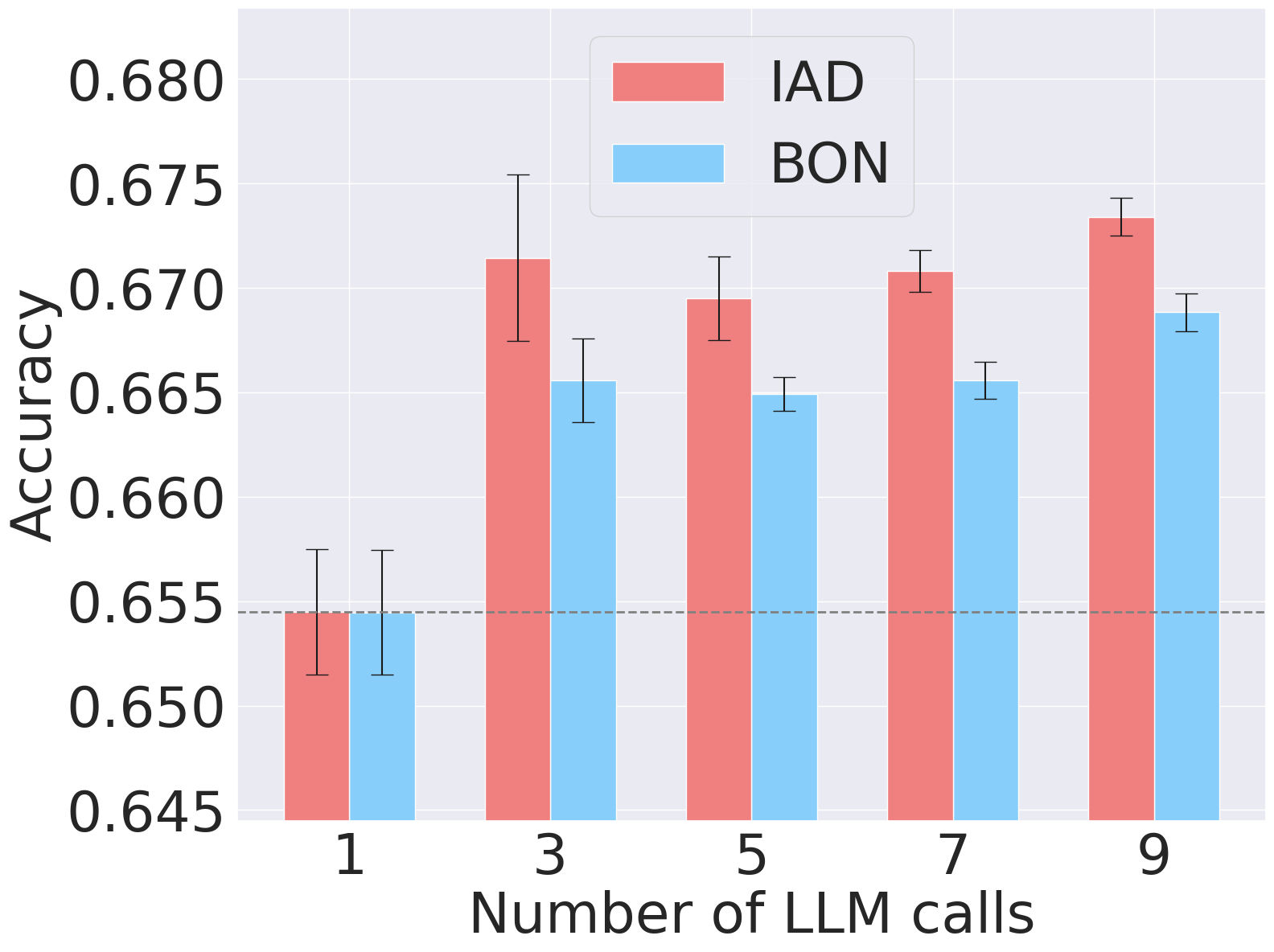}}
        \caption{\centering}
    \end{subfigure}
    \begin{subfigure}[b]{0.329\textwidth}  
        \centering
        \scalebox{1.0}{\includegraphics[width=\textwidth]{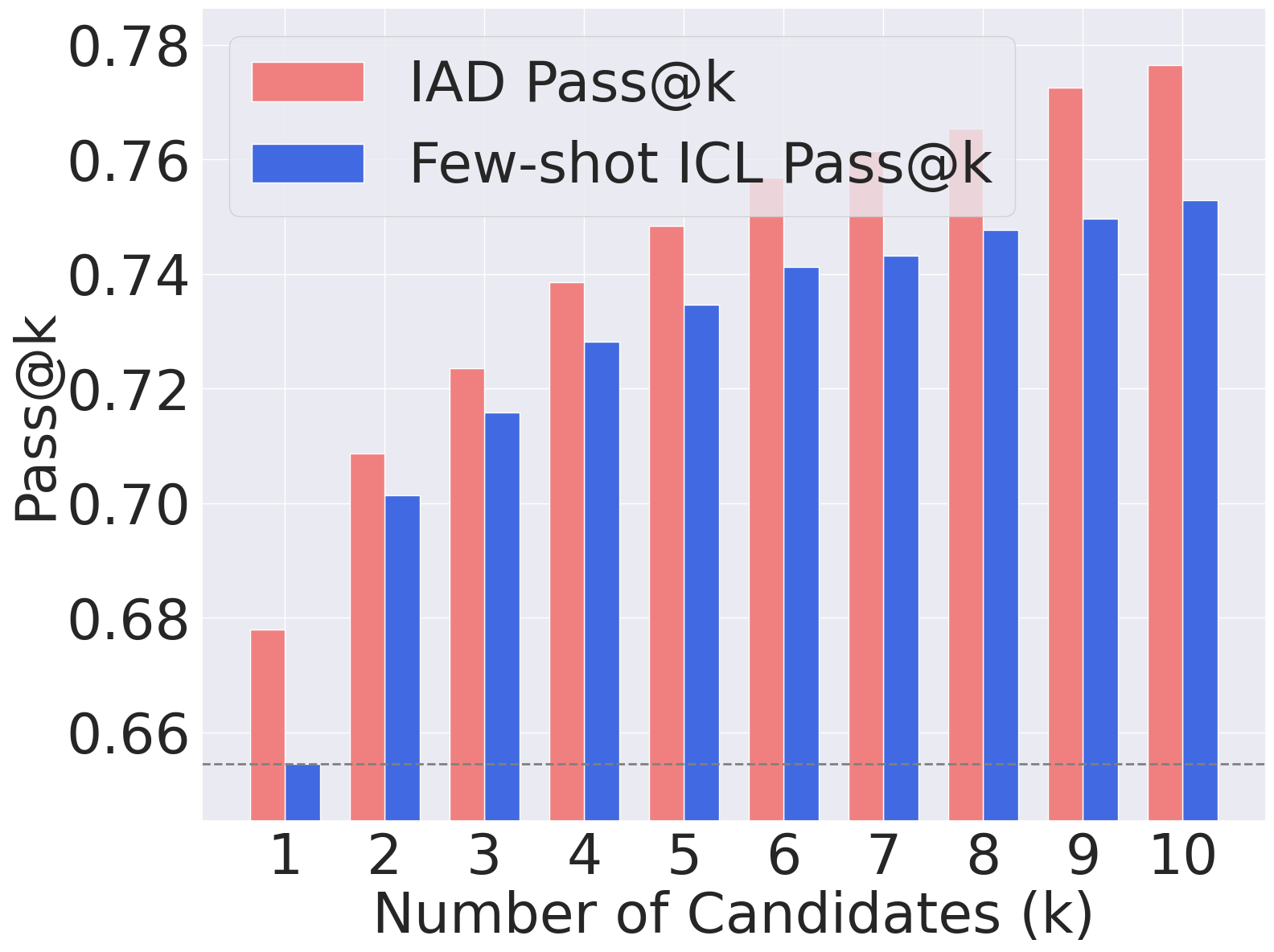}}
        \caption{\centering}
    \end{subfigure}
    \begin{subfigure}[b]{0.329\textwidth}  
        \centering
        \scalebox{1.0}{\includegraphics[width=\textwidth]{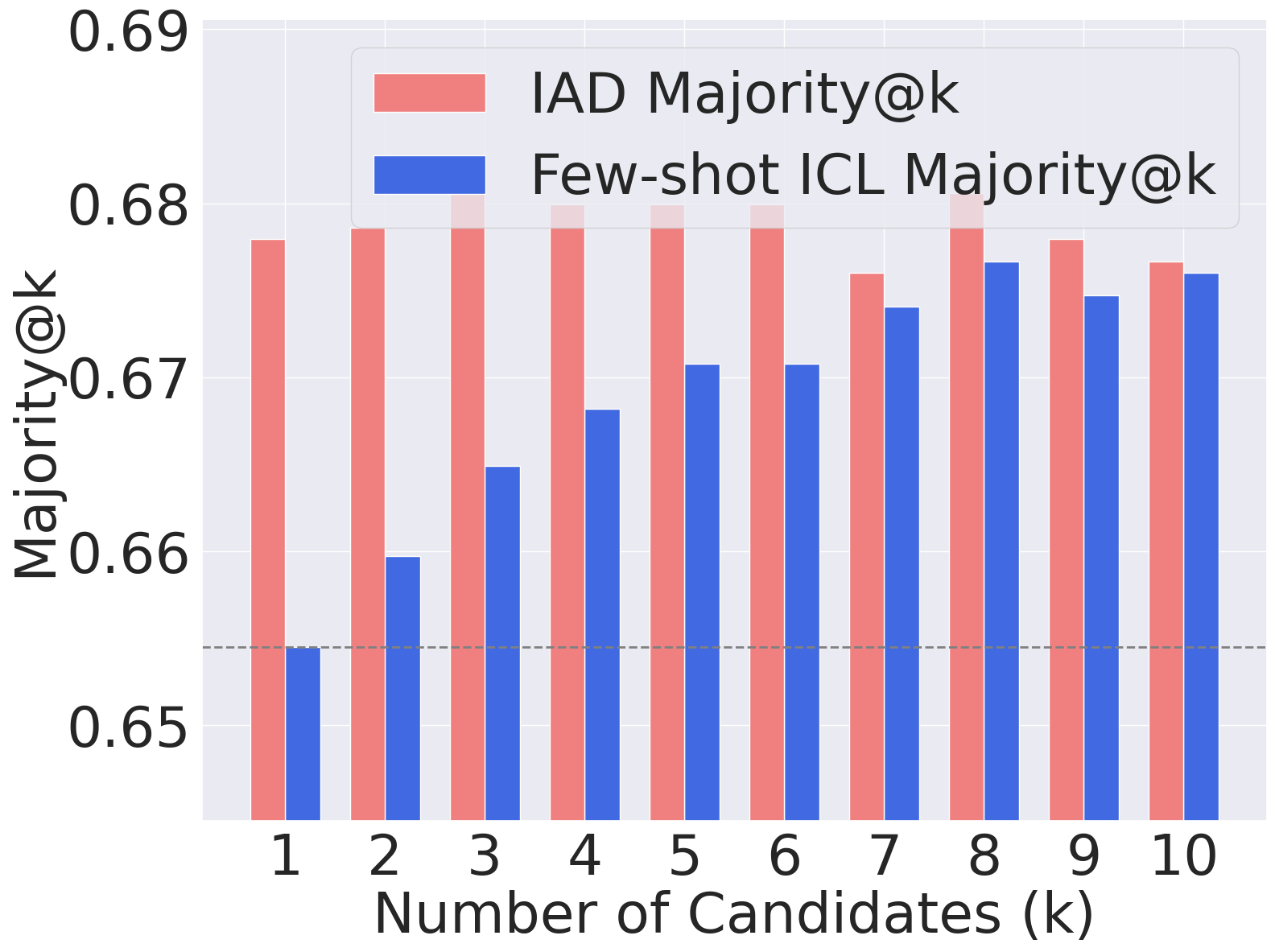}}
        \caption{\centering}
    \end{subfigure}
    %
   
    \caption{\textbf{\textbf{Test-time Scaling with Feedback in Text2SQL}}: (a)Accuracy vs compute comparison between the best-of-N method and our approach for Gemini-1.5-flash.  (b) Pass@K performance comparison between queries generated using our approach and Few-shot ICL with Gemini-1.5-flash. (c) Majority@K performance comparison between queries generated using our approach and Few-shot ICL with Gemini-1.5-flash.  (d) Accuracy comparison between the best-of-N method and our approach for Gemini-1.5-pro. IAD consistently outperforms all the baselines across the settings. (e) Pass@K performance comparison between queries generated using our approach and Few-shot ICL with Gemini-1.5-pro.  (f) Majority@K performance comparison between queries generated using our approach and Few-shot ICL with Gemini-1.5-pro. }
    \label{fig:candidate_generator_comparison}
\end{figure*}

\noindent The IAD process begins with an initial input $x$ and iteratively improves a candidate solution $\hat{y}$ over a series of iterations. Each iteration $t$ is composed of the following interconnected stages:

\noindent\textbf{Step 1: Sampling and Generation.} At each step $t$, a new candidate solution $y_{t+1}$ is sampled from the reference policy $\pi_0$. This policy is conditioned on the input $x$ along with the best solution found in previous iterations $\hat{y}_t$ and the feedback $fb_t$
\begin{align}
    y_{t+1} \sim \pi_0(\cdot|x ,\hat{y}_{t}, s_t, p_t),
\end{align}
Feedback $fb_t$ encodes guiding instructions  incorporated into the model’s input prompt for ex: \textit{Surpass the best response, avoiding previous mistakes}  or \textit{Improve upon the previous best solution by addressing the following errors...}

\noindent \textbf{Step 2: Evaluation and Selection} : The newly generated candidate $y_{t+1}$ is then evaluated against the current best $\hat{y}_t$ using the reward function or verifier $R(x,y)$. The better solution is selected and carried forward to the next iteration as the new best-so-far
\begin{align}
\hat{y}_{t+1} = \arg\max_{y \in (y_t, \hat{y}_{t})} R(x, y).
\end{align}
IAD has the flexibility to incorporate critique-based feedback, such as LLM-judge, which identifies specific areas needing improvement.

\noindent \textbf{Step 3: Feedback Integration.}  One of the most crucial components in iterative decoding approaches is its ability to incorporate rich, structured feedback to guide the next generation step. The evaluation from Step 2 can be either: 

\noindent (i) textual feedback, which provides natural language instructions or critiques; 

\noindent (ii) scalar feedback, such as numerical scores or preference comparisons.

\noindent Textual feedback is relatively easy to integrate, as it can be directly injected into the model’s context or prompt. However, generating high-quality textual feedback is often expensive, requiring either powerful LLM judges or human annotation, and may vary in precision and consistency.

\noindent \textbf{Challenge in Feedback (Score to Text) Extraction and Our Key Approach}: \noindent On the other hand, scalar feedback is significantly easier to obtain; however, translating these signals into actionable guidance remains a key challenge. Directly appending the scalar value to the prompt, as done in several prior works \citep{llmopt}, does not leverage the full informational content of the signal and often fails to yield meaningful improvements. Hence, we design a feedback integration method in IAD that translates scalar scores into structured directional feedback. At each iteration, IAD transforms score-based or comparative feedback into structured guidance by identifying the best and worst responses at each iteration:
\begin{align}
y^t_{\text{best}} &= \arg\max_{y \in {y_{t+1}, \hat{y}_t}} R(x, y), \\
y^t_{\text{worst}} &= \arg\min_{y \in {y_{t+1}, \hat{y}_t}} R(x, y)
\end{align}
These are explicitly fed back into the model through prompt conditioning. The feedback is constructed by concatenating the instruction with the best and worst responses:
\begin{align}
fb_t = \texttt{[Instruction]} \oplus \texttt{Best:} y^t_{\text{best}} \oplus \texttt{Worst:} y^t_{\text{worst}}
\end{align}
where the Instruction states generic guidance like \textit{Improve upon the best response while avoiding mistakes from the worst}. This feedback thus provides a clear directional signal for improvement.

\noindent The accepted response $\hat{y}_t$ updates the context for subsequent generations, progressively re-weighting the proposal distribution toward higher-quality outputs. By conditioning on $\hat{y}_t$, the sampling process is guided towards responses with higher rewards, reducing the gap between the reference distribution $\pi_0(\cdot|x, \hat{y}_t)$ and the optimal distribution $\pi^*(\cdot|x)$, as observed in experimental results.

\noindent \textbf{Example}: For instance, in an SQL code generation task, the model might initially produce a non-functional or erroneous SQL statement. However, in the next iteration, it generates a different SQL code, and through verifier comparisons, we determine which version is better—e.g., if the later version passes more test cases, we infer that the model should move in that direction. Using an LLM as a judge makes this refinement process more targeted, as it can provide explicit feedback on errors like \textit{"This condition is incorrect", "Fix the syntax here", or "This table join is unnecessary"} and suggest improvements at each iteration.

\begin{figure*}[t]
    \centering
    \begin{subfigure}[b]{0.329\textwidth}
        \centering
        \includegraphics[width=\textwidth]{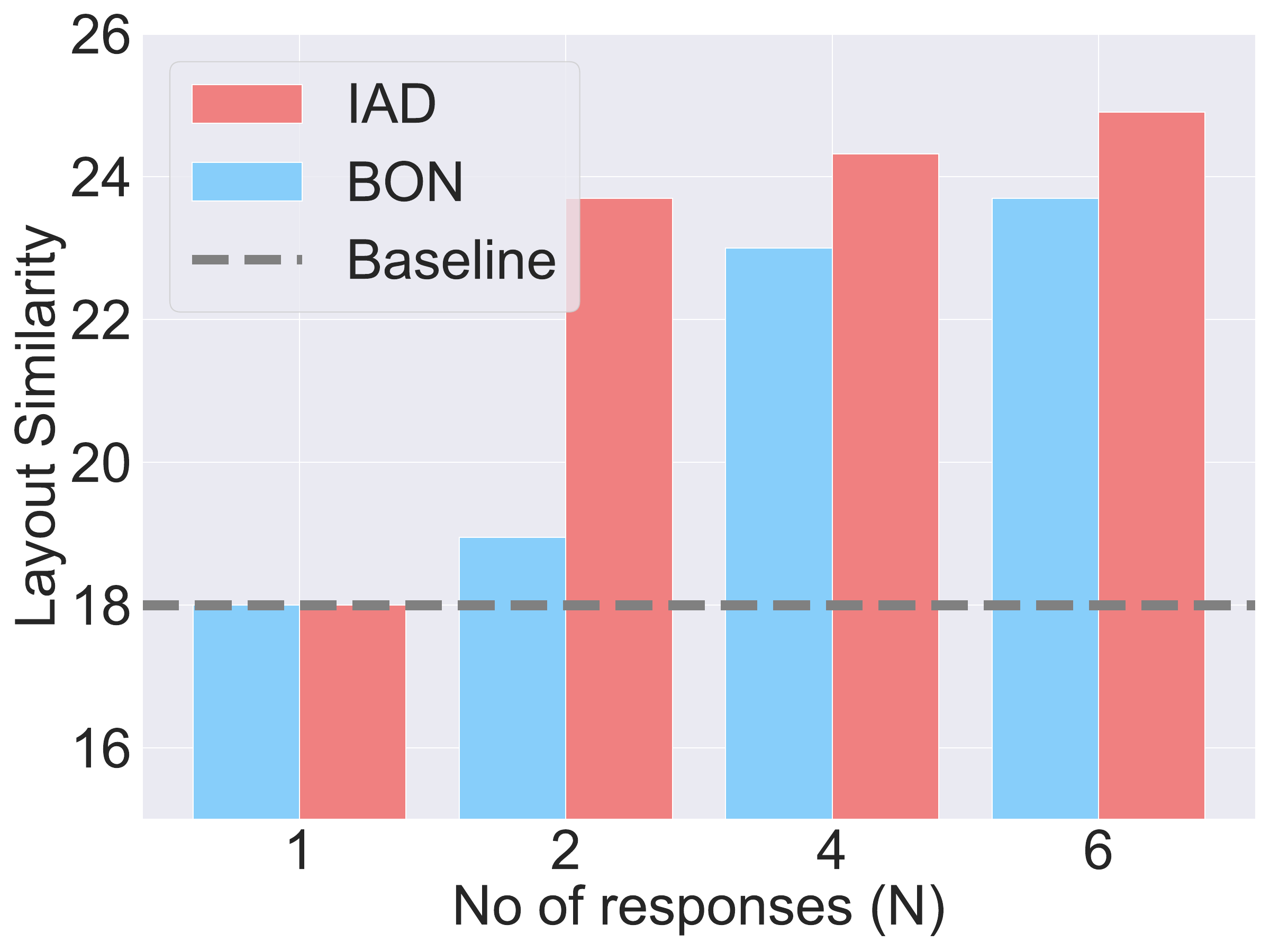} 
        \caption{}
    \end{subfigure}
    \hfill 
    \begin{subfigure}[b]{0.329\textwidth}
        \centering
        \includegraphics[width=\textwidth]{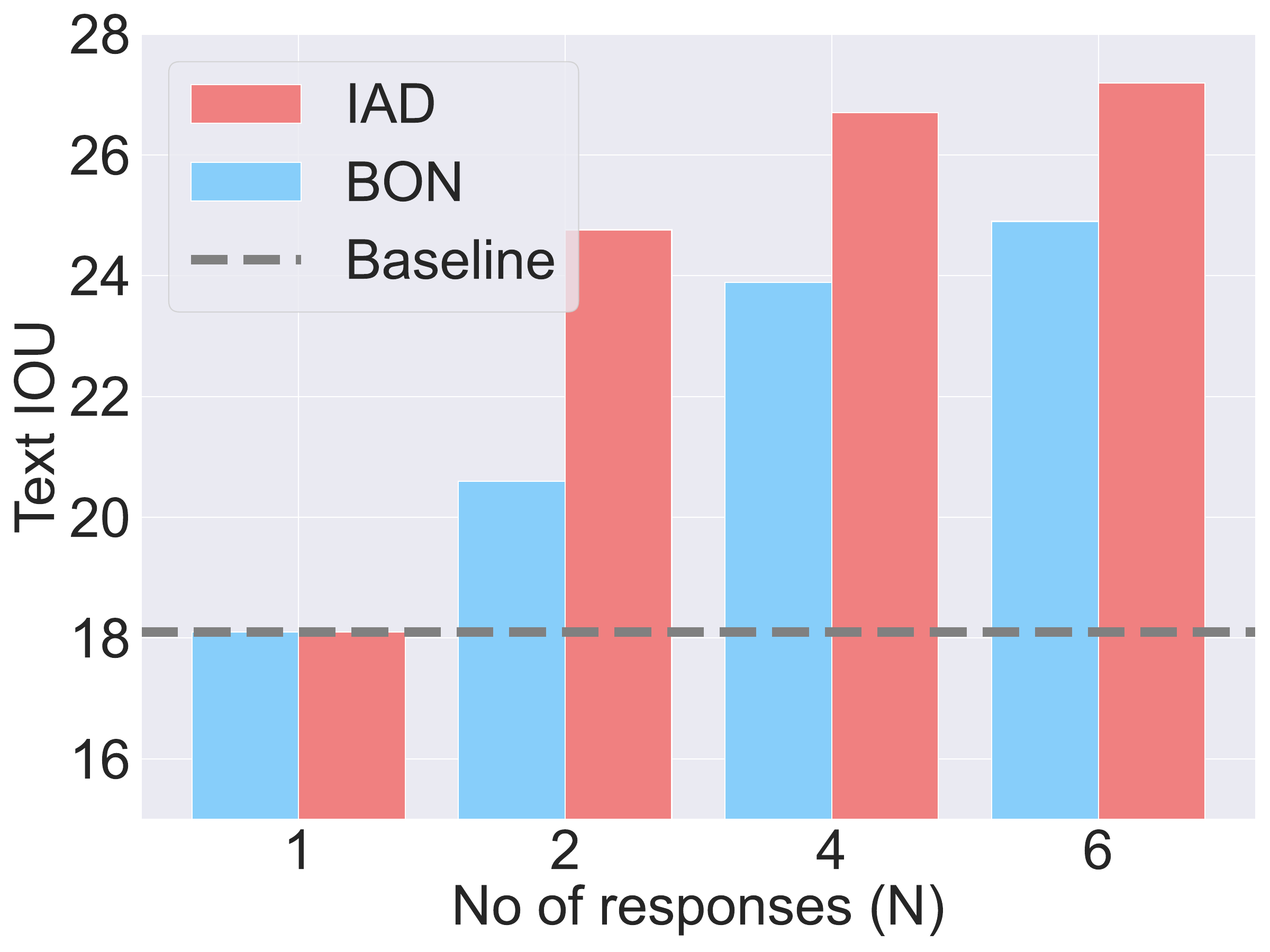} 
        \caption{}
    \end{subfigure}
    \begin{subfigure}[b]{0.329\textwidth}
        \centering
        \includegraphics[width=\textwidth]{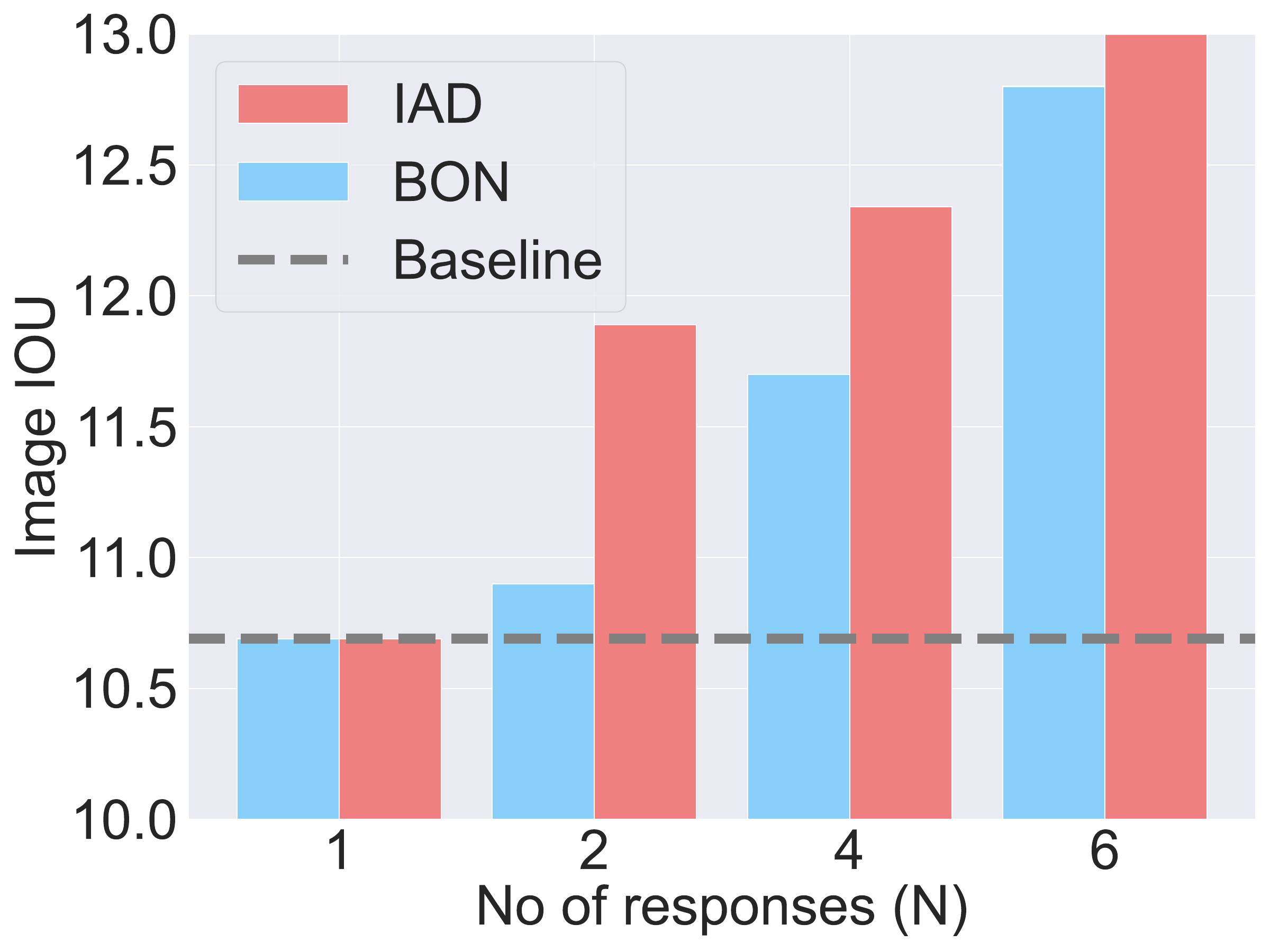} 
        \caption{}
    \end{subfigure}
    \begin{subfigure}[b]{0.32\textwidth}
        \centering
        \includegraphics[width=\textwidth]{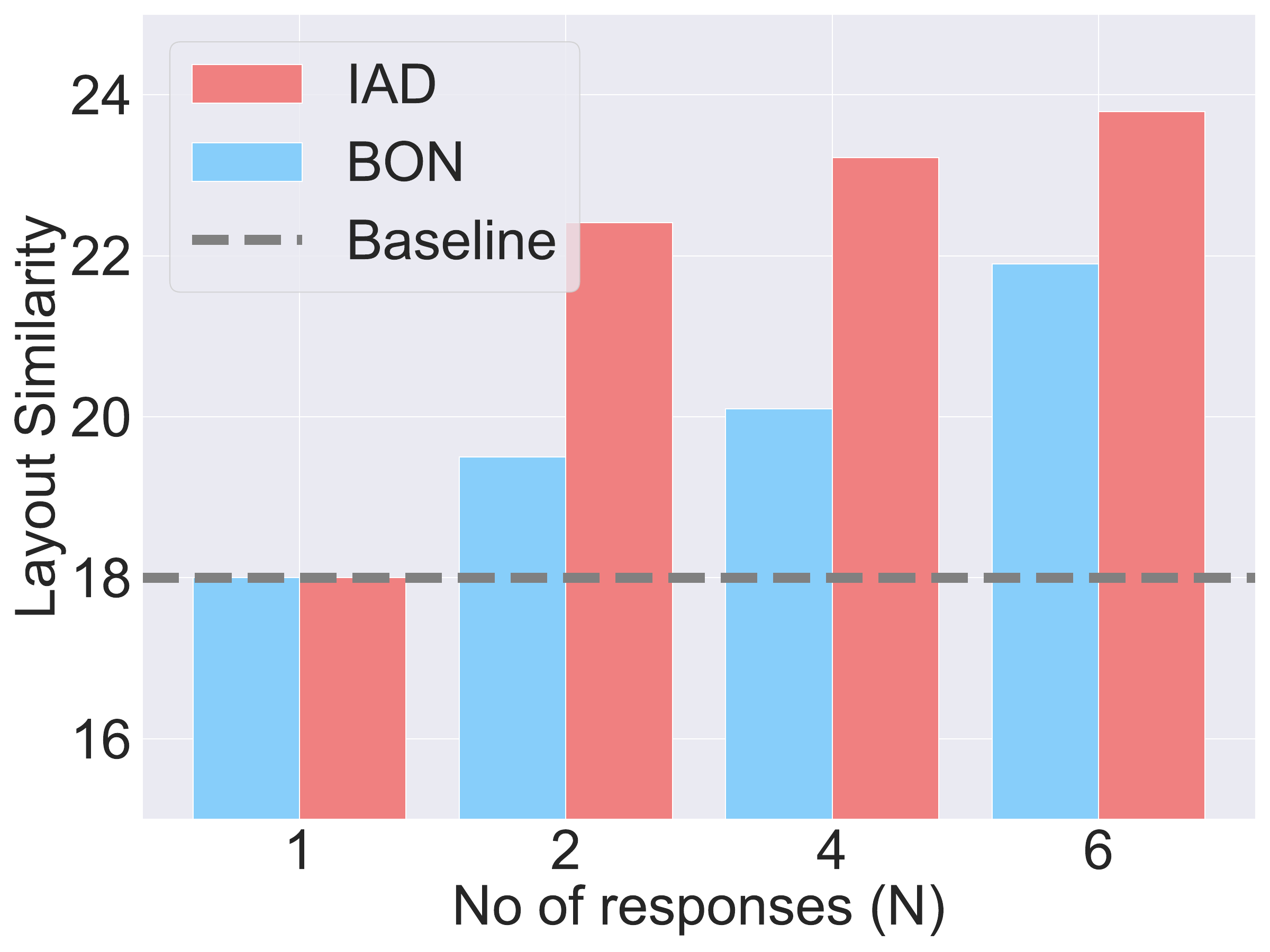} 
        \caption{}
    \end{subfigure}
    \hfill 
    \begin{subfigure}[b]{0.32\textwidth}
        \centering
        \includegraphics[width=\textwidth]{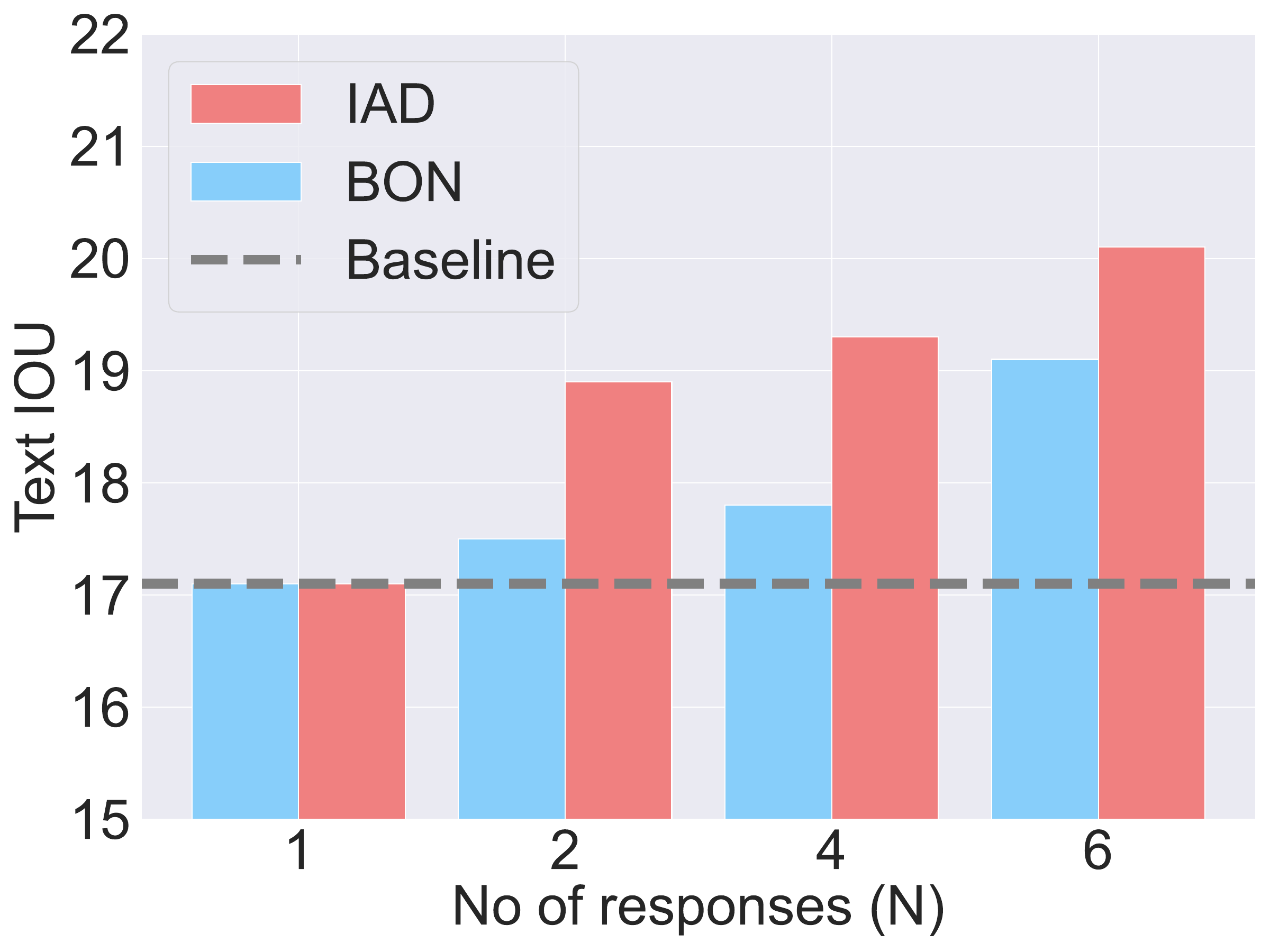} 
        \caption{}
    \end{subfigure}
    \begin{subfigure}[b]{0.32\textwidth}
        \centering
        \includegraphics[width=\textwidth]{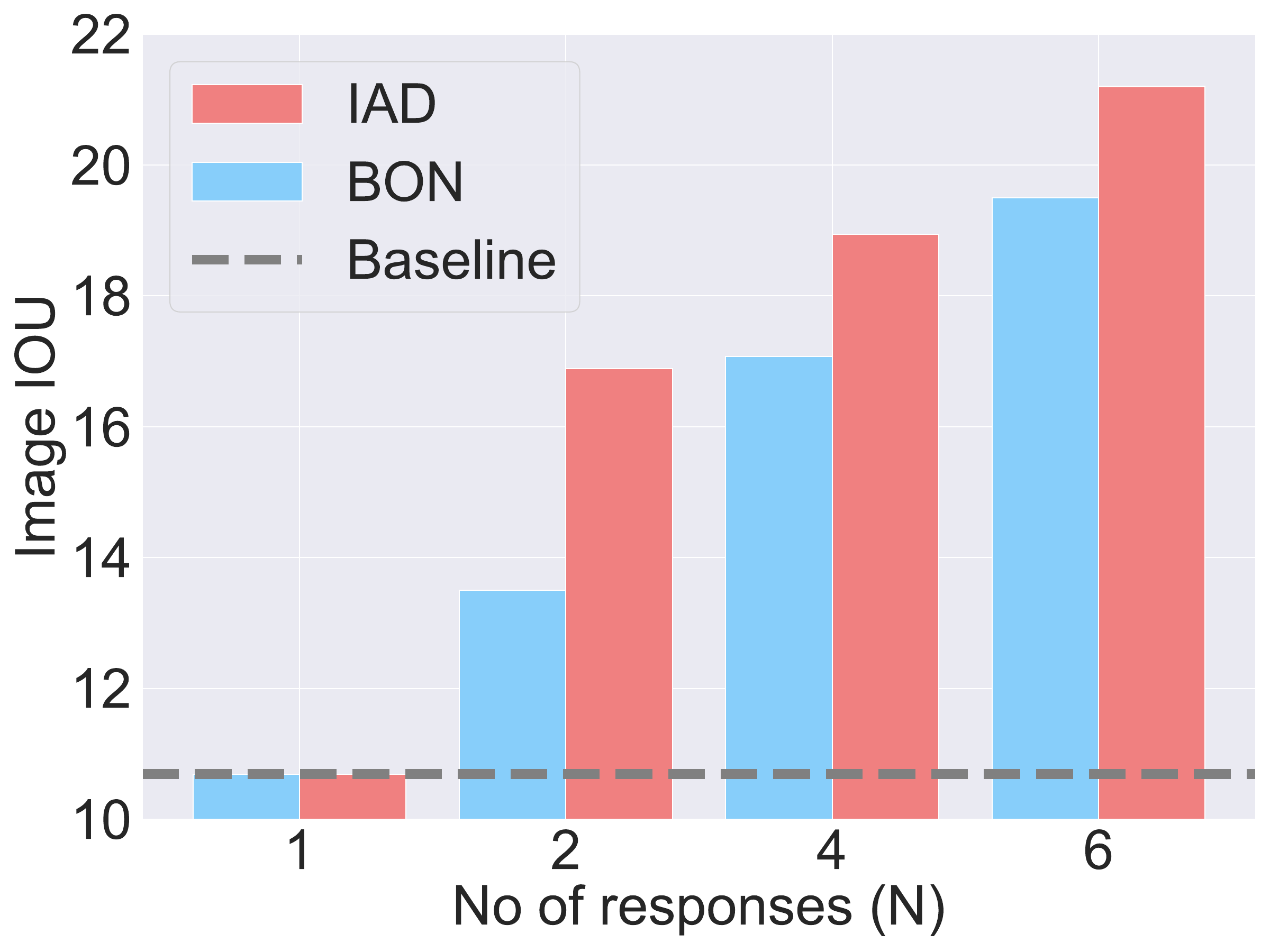} 
        \caption{}
    \end{subfigure}
    \caption{\textbf{Sketch2code}: This figure provides a comparison of IAD (ours) with Best-of-N sampling (SoTA) and single-turn generation with Gemini-1.5-Flash for the metrics - (a, d) Layout Similarity (b, e) TextIoU (c, f) ImageIoU across varying the number of generations (N). Top 3 rows, the optimization is done taking Text IOU as the verifier and the bottom 3 rows with Image IOU as the verifier. So, this also shows both the generalisability and performance improvement of IAD over baselines.}
    \label{fig:mainfig}
\end{figure*}

\begin{figure*}[!t]
    \centering
    \begin{subfigure}[b]{0.43\textwidth}  
        \centering
        \scalebox{1.0}{\includegraphics[width=\textwidth]{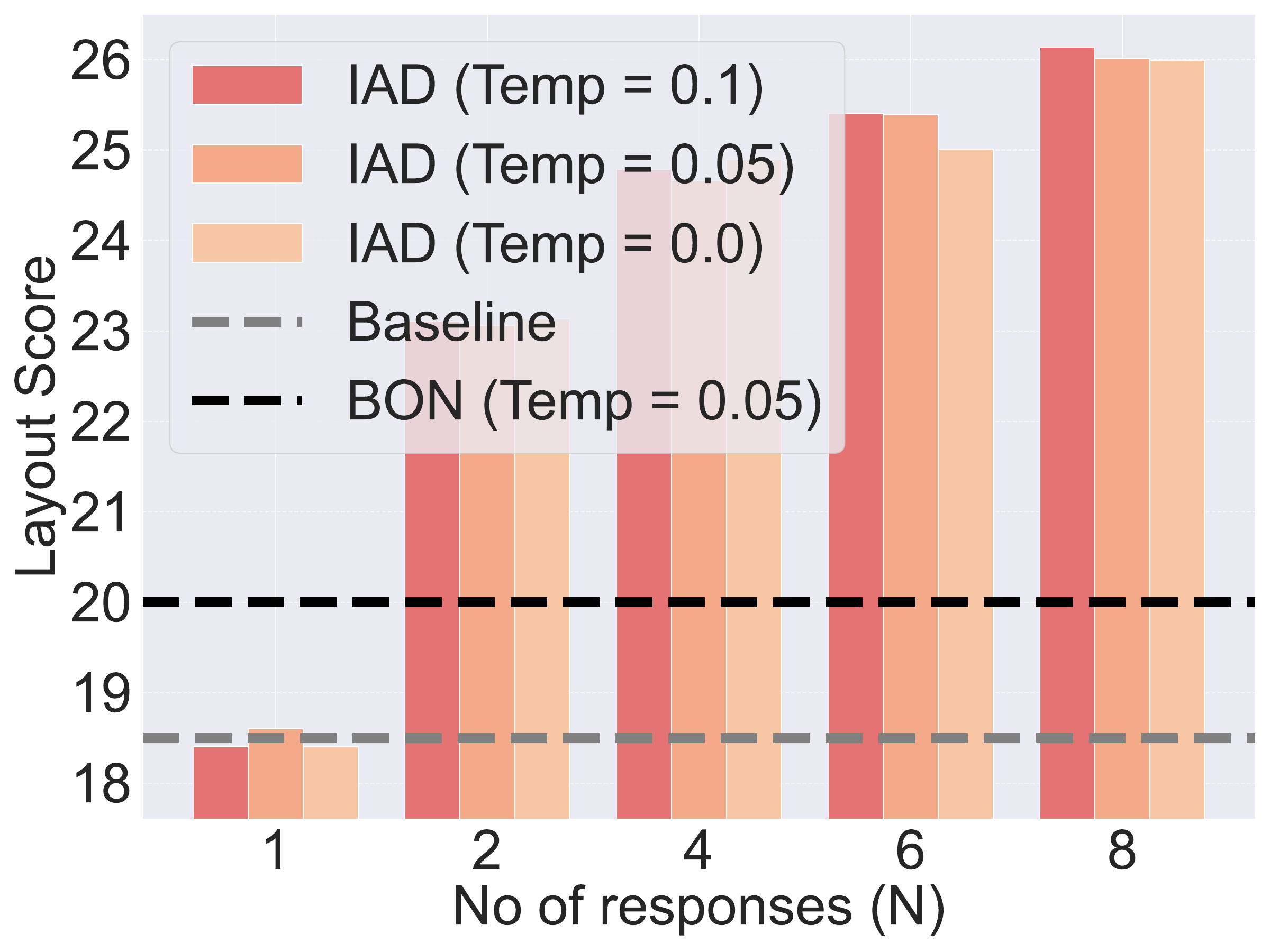}}
        \caption{\centering}
    \end{subfigure}
    \begin{subfigure}[b]{0.44\textwidth}  
        \centering
        \scalebox{1.0}{\includegraphics[width=\textwidth]{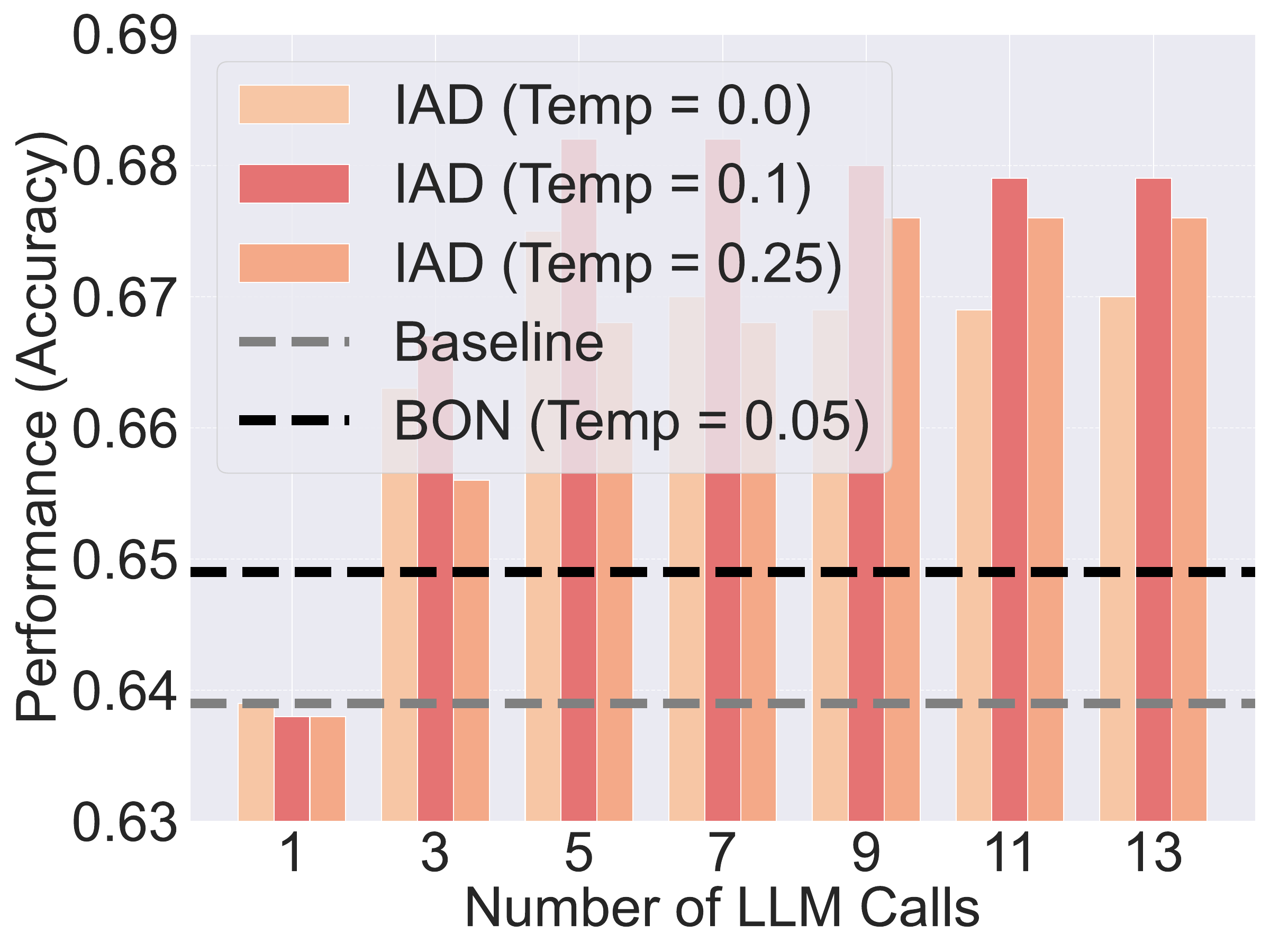}}
        \caption{\centering}
        \label{fig:example_plot_iter}
    \end{subfigure}
  \vspace{-.1in}  \caption{\textbf{Beyond Sampling in Sketch2code}: (a) Compares the performance of IAD at low temperature across varying number of generations to disentangle the effect of stochasticity and the improvement from iterative feedback based on IAD. (b) \textbf{Text2SQL} :Compares the accuracy of IAD on BIRD development set at low temperature across varying number of generations to highlight the improvement from iterative feedback based on IAD. Baseline is Gemini-1.5-pro. BON remains similar as stochasticity improves across the number of responses.}
    \label{fig:low_temp_abl1}
\vspace{-.1in}
\end{figure*}

\begin{figure*}[t]
    \centering
    \begin{subfigure}[b]{0.4\textwidth}  
        \centering
        \scalebox{1.0}{\includegraphics[width=\textwidth]{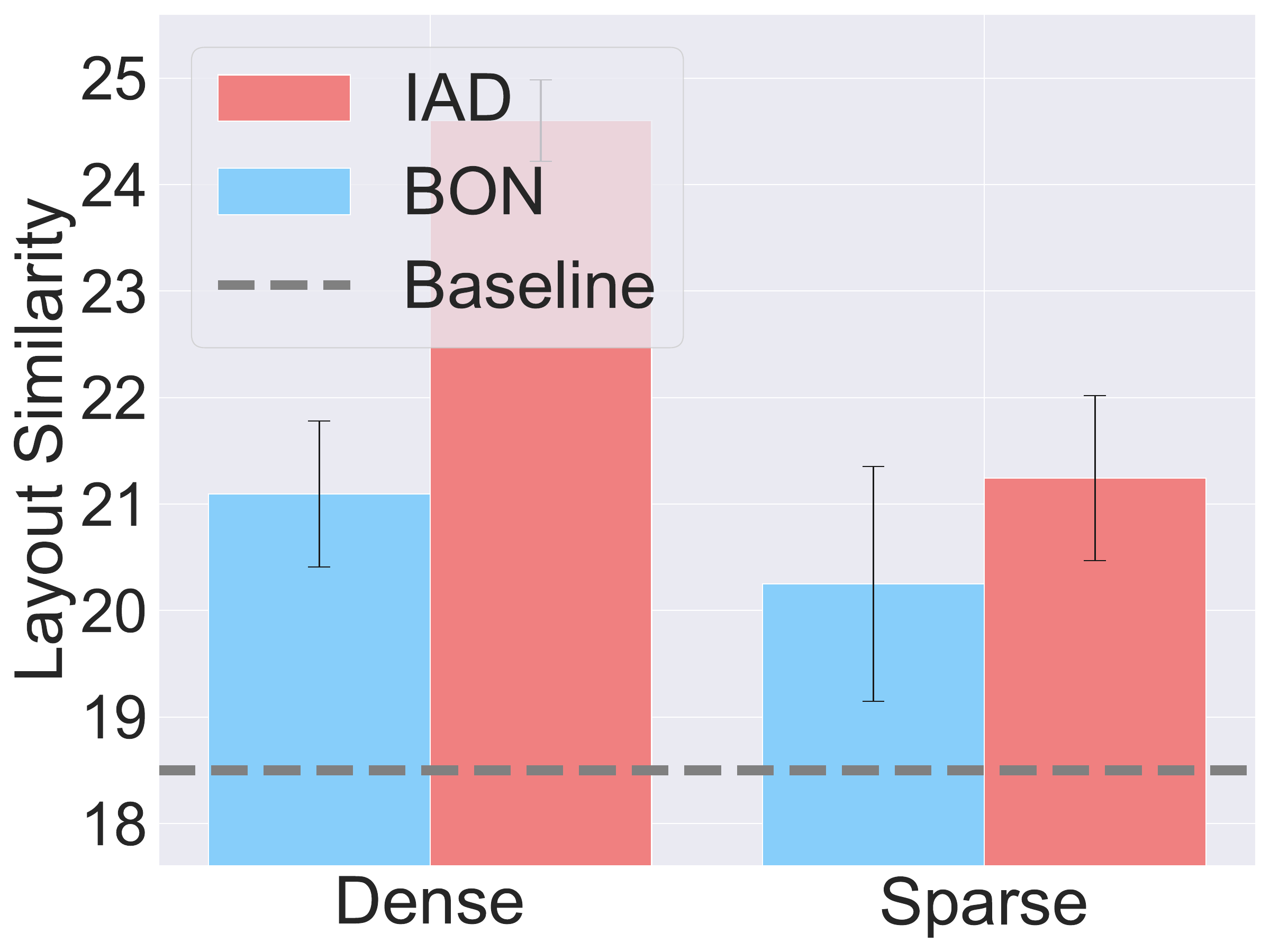}}
        \caption{\centering}
    \end{subfigure}
    \begin{subfigure}[b]{0.4\textwidth}  
        \centering
        \scalebox{1.0}{\includegraphics[width=\textwidth]{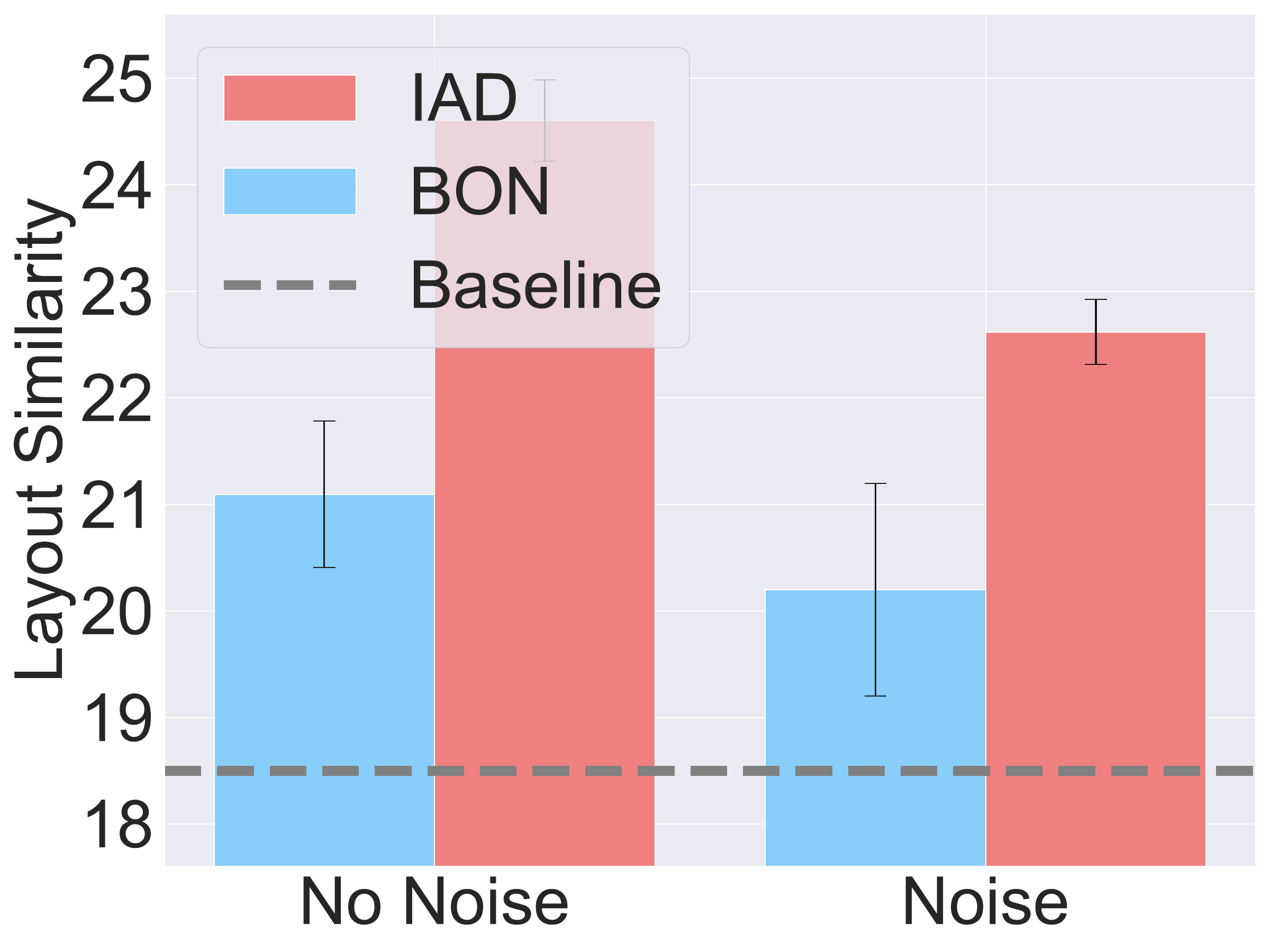}}
        \caption{\centering}
    \end{subfigure}
    \begin{subfigure}[b]{0.4\textwidth}  
        \centering
        \scalebox{1.0}{\includegraphics[width=\textwidth]{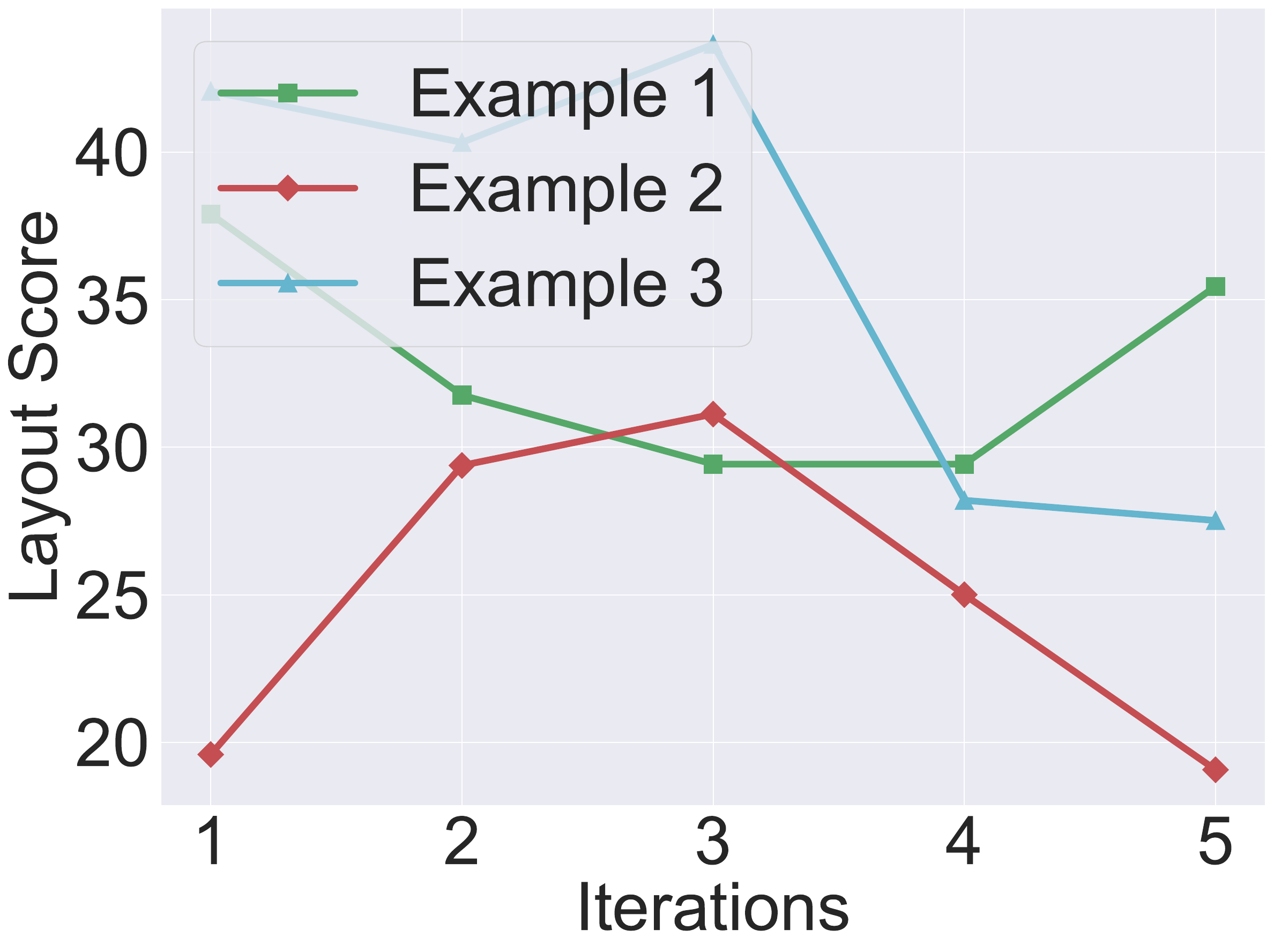}}
        \caption{\centering}
    \end{subfigure}
     \begin{subfigure}[b]{0.4\textwidth}  
        \centering
        \scalebox{1.0}{\includegraphics[width=\textwidth]{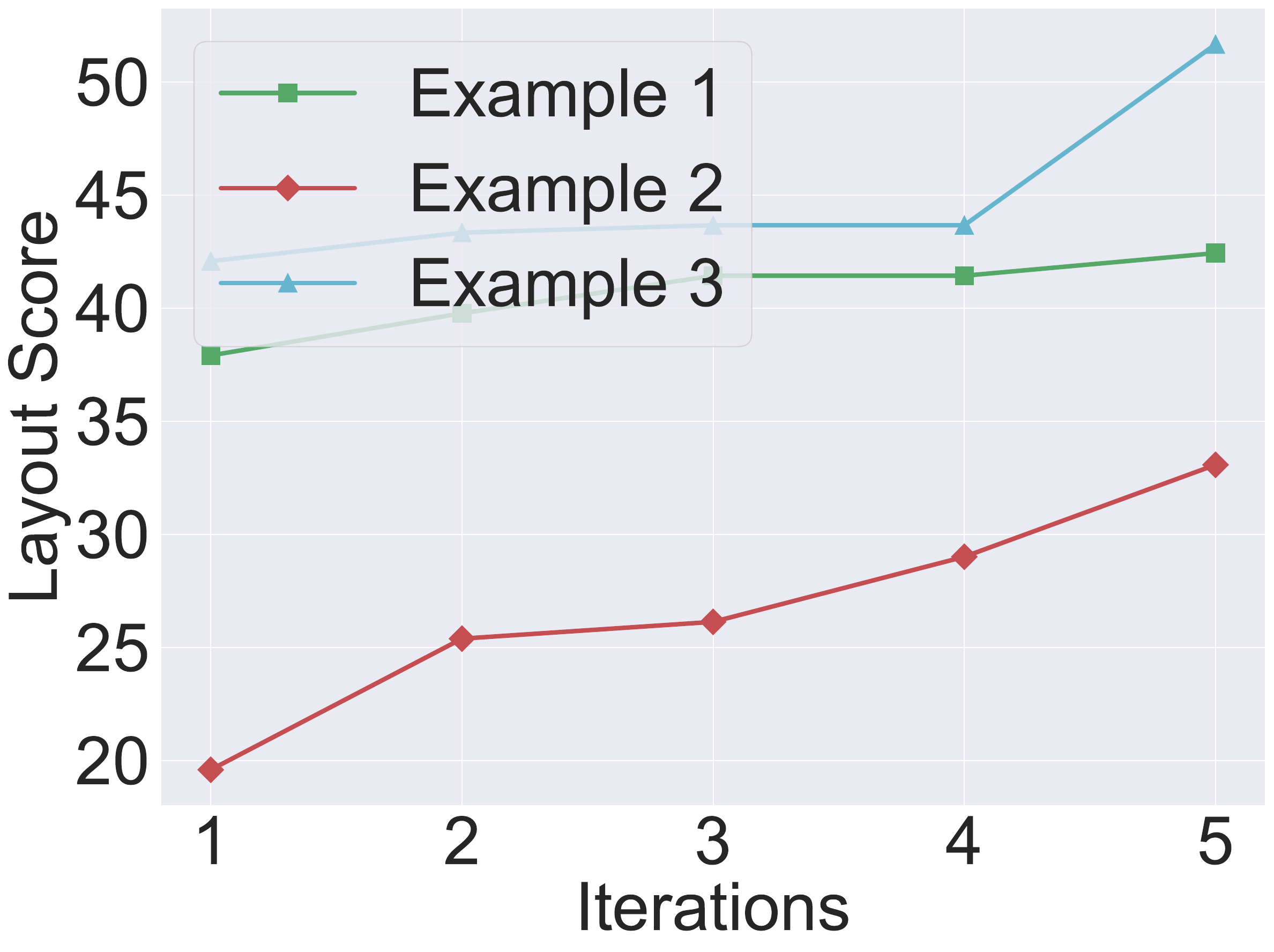}}
        \caption{\centering}
    \end{subfigure}
    %
    \caption{\textbf{Sketch2code}: (a) Represents the performance of IAD with N = 2 w.r.t Layout score with varied sparsity of feedback which shows gains reduces over BON. (b) Performance comparison of IAD with BON with N = 2 at varied noise levels of feedback, which shows gains reduced over BON. However, IAD still improves over baselines under Sparse and Noisy feedback. Figure (c) Represents the performance of Self-Refine \citep{refine1} based approaches across iterations using LLM as a judge. (d) Represents the performance of IAD with a scalar score, which shows monotonic improvements, highlighting the importance of the optimality of feedback and verification.}
    \label{fig:exm_plot}
\end{figure*}
\noindent

\section{Experiment Setup}
In this section, we provide a detailed discussion on the experimental analysis with respect to the agentic tasks below. For all our experiments,  we have utilized an NVIDIA A100-SXM4-40GB GPU with 40GB of VRAM, running on CUDA 12.4 and driver version 550.90.07.  To enable a comprehensive evaluation, we organize the experimental setup into three components: (1) the benchmarks used to assess agentic capabilities across diverse modalities, (2) the base models serving as generative backbones, and (3) the set of baselines for comparative analysis.

\subsection{Benchmarks} \noindent We perform empirical analysis on 4 challenging agentic environment benchmarks detailed as follows. 

\noindent \textbf{Text2SQL} tasks map natural language questions to executable SQL queries over structured databases. These tasks require deep reasoning to interpret user intent and generate syntactically and semantically correct queries. We use the BIRD benchmark \cite{li2024can}, which includes 12,751 question-SQL pairs spanning 95 databases and 37 professional domains. Performance is evaluated using execution accuracy (EX), where a prediction is correct if it yields the same results as the ground truth query when executed. 

\noindent \textbf{Sketch2code} \citep{li2024sketch2codeevaluatingvisionlanguagemodels} tests the multi-modal abilities of agents by converting wireframe sketches into functional HTML prototypes. The task requires aligning visual cues with structured code, handling layout ambiguities, and generating precise UI structures. Evaluation is based on three metrics: Layout Similarity (IoU over UI components), Text IoU (text alignment accuracy) and Image IoU (CLIP-based visual similarity). These metrics strongly correlate with human judgment. 

\noindent \textbf{Intercode} \citep{intercode} evaluates agents in structured programming tasks across four domains: Bash, Python, SQL, and CTF. Each task involves a series of agent decisions, including command execution and error handling, requiring both syntactic precision and logical planning. Performance is measured using reward signals derived from task execution correctness. We focus on the Bash domain for our experiments, where agents must generate correct shell commands based on natural language instructions. 

\noindent \textbf{Webshop} \citep{yao2023webshopscalablerealworldweb} simulates a real-world online shopping environment, requiring agents to perform long-horizon, language-guided decision-making. Given product queries, agents interact with webpages (search, click, select) to find matching items. Key evaluation metrics include: Success Rate (SR): whether the selected item satisfies all constraints. Progress Rate (PR): how closely the agent's actions align with task goals.

\subsection{Base Models} We experiment with a wide range of large language and vision-language foundation models across all four benchmark tasks to study the generality of IAD across different agentic settings. Here we outline the base models used for each benchmark. For \textbf{Sketch2Code}, we use Gemini-2.5-Pro \citep{deepmind2024gemini2}, Gemini-2.0-Flash, Gemini-1.5-Pro, and Gemini-1.5-Flash \citep{geminiteam2024gemini15unlockingmultimodal} as base models to compare test-time scaling approaches with and without feedback. Furthermore, we include InternVL2-8b \citep{chen2025expandingperformanceboundariesopensource}, Llava-1.6-8b \citep{liu2023improvedllava}, Claude-3-Sonnet , GPT-4o-Mini, Claude-3-Opus, Claude-3-Haiku, Claude-3.5-Sonnet and GPT-4o \citep{openai2024gpt4technicalreport} as baseline models for broader comparison. Vision-language models are used to handle the multi-modal input (user sketches) and generate HTML output. For \textbf{Text2SQL}, we use Gemini-1.5-Pro, Gemini-1.5-Flash, GPT-4o. These are used both as generative models and occasionally as judges to provide feedback. For \textbf{Webshop}, we use Gemini-1.5-Pro, Gemini-1.5-Flash, and GPT-4o to evaluate the effectiveness of IAD in long-horizon decision-making within a simulated online shopping environment. Additionally, we also use Lemur-70b \citep{xu2024lemurharmonizingnaturallanguage}, Mistral-7b \citep{jiang2023mistral7b}, Vicuna-13b-16k \citep{zheng2023judgingllmasajudgemtbenchchatbot}, GPT-3.5-Turbo-16k, Text-Davinci-003 \citep{ye2023comprehensivecapabilityanalysisgpt3}, DeepSeek-67b \citep{deepseekai2024deepseekllmscalingopensource} for additional baseline comparisons. Finally, for Intercode \citep{intercode}, GPT-4 is used as the base model for generating Bash commands and interacting in structured programming tasks.

\subsection{Baselines} We compare and contrast IAD against a range of black-box inference-time approaches, which include: {(1) Single-Turn:}  Zero-shot generation from the base model using only the input prompt. {(2) Best-of-N} (BoN) and its variants \citep{bon1, bon2, bon3, gui2025bonbon}: Sampling-based methods where multiple outputs are generated from the base model and the best is selected via a verifier. (3) Self-Refine \citep{refine1} and related methods such as Self-Debug \citep{selfdebug} and Try-Again \citep{tryagain}: Iterative refinement approaches that use self-critique or LLM-based retry mechanisms to improve outputs. (4) We also compare IAD with few-shot prompting-based methods such as DIN-SQL \citep{pourreza2024din} and DAIL-SQL \citep{gao2023text}, as well as multi-agent test-time approaches like MAC-SQL \citep{wang2023mac}, to compare the performance of IAD over test-time baselines.

\noindent We study the effect of feedback in IAD along four key dimensions: (1) Accuracy vs compute - Budget controlled scaling (2) Impact of adaptive feedback beyond sampling diversity (3) Impact of feedback modalities, (4) sensitivity to  feedback quality under varying levels of noise and sparsity

\begin{figure*}[t]
    \centering
    \begin{subfigure}[b]{0.43\textwidth}  
        \centering
        \scalebox{1.0}{\includegraphics[width=\textwidth]{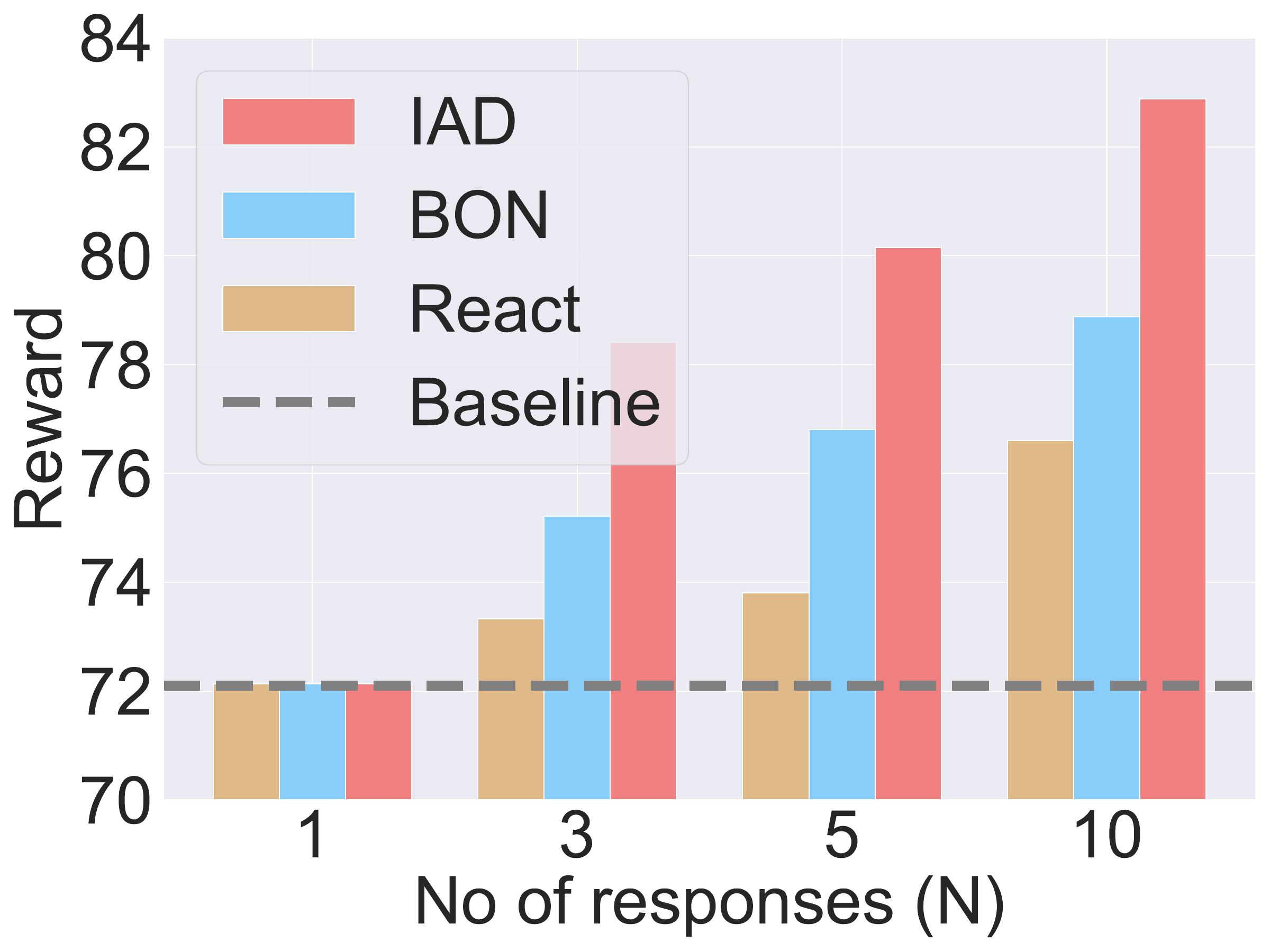}}
        \caption{\centering}
        \label{fig:intercode}
    \end{subfigure}
    \begin{subfigure}[b]{0.43\textwidth}  
        \centering
        \scalebox{1.0}{\includegraphics[width=\textwidth]{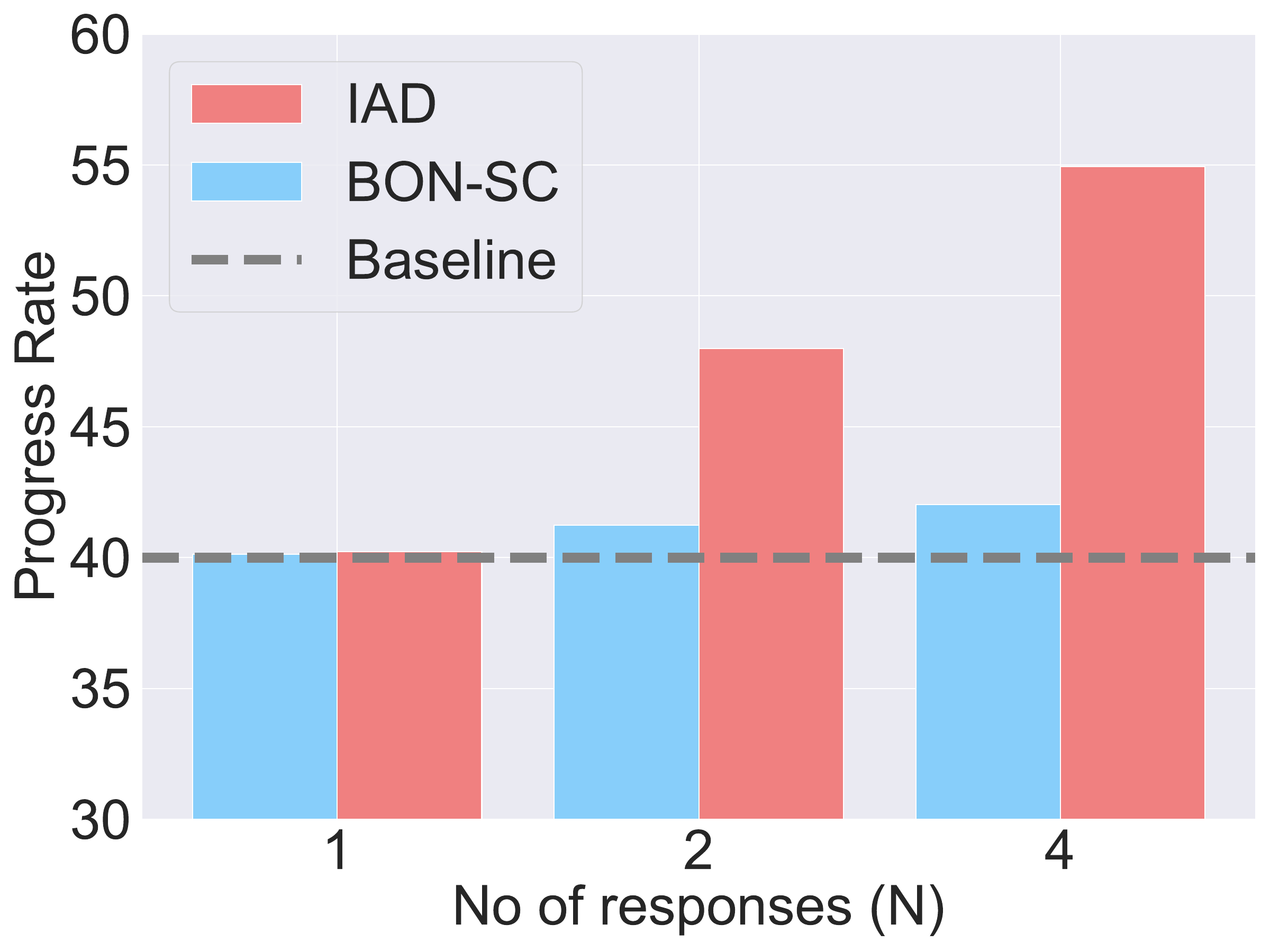}}
        \caption{\centering}
    \end{subfigure}
    \caption{(a) \textbf{Test-time scaling with feedback in Intercode} : Comparison of IAD with baselines (GPT-4) on the accuracy–compute trade-off shows that IAD, with appropriate feedback, enables more effective test-time scaling. \textbf{Intercode} : Comparison of IAD with baselines (GPT-4) on the accuracy–compute trade-off shows that IAD, with appropriate feedback, enables more effective test-time scaling. (b) \textbf{Test-time scaling with feedback in Webshop} : Progress-rate comparison of IAD over Best of N with Self-consistency (BON-SC) and baseline on a random selection of 20\% of the total evaluation set using weaker Gemini-1.5-Flash-01 (weaker model). Figure highlights the role of feedback in IAD with a weaker model.}
    \label{fig:new_envs}
\end{figure*}

\section{Experiment Results and Analysis}

\subsection{Accuracy vs compute tradeoff} 

We investigate how IAD scales with compute and model capacity across Sketch2Code, Text2SQL, Intercode, and Webshop. Our results show that with accurate feedback, IAD achieves up to 10\% accuracy gains over feedback-free baselines like BoN under constrained budgets as shown in  Table~\ref{tab:sketch2code_comparison}, Table~\ref{tab:text-to-SQL}. However, as compute increases, via more samples or stronger models, the performance gap narrows, indicating diminishing returns from feedback alone.

\begin{table}[htbp]
\centering
\resizebox{0.35\textwidth}{!}{\begin{tabular}{lc}
\hline
\textbf{Method} & \textbf{Exe Acc} \\
\hline
DIN-SQL + GPT-4 & 50.72 \\
DAIL-SQL + GPT-4 & 54.76 \\
MAC-SQL + GPT-4 & 57.56 \\
MCS-SQL + GPT-4 & 63.36 \\
E-SQL + GPT-4o & 65.58 \\
\textbf{IAD + GPT-4o} & 65.97 \\
\textbf{IAD + Gemini-1.5-pro} & \textbf{68.05}  \\
\hline
\end{tabular}}
\caption{ \textbf{Text2SQL} - Execution accuracy comparison of previous works with our proposed approach}
\label{tab:text-to-SQL}
\end{table} 

\noindent \textit{Sketch2Code:} Layout Similarity, Text IoU, and Image IoU steadily improve with more IAD iterations. Even with just N=2 and a weaker model (Gemini-1.5-Pro), IAD surpasses BoN and single-turn baselines by 3–4\% (Figure~\ref{fig:mainfig_sk2code}, Table~\ref{tab:sketch2code_comparison}). But with more capable models (e.g., Gemini-2.5-Pro), BoN narrows the gap, suggesting parallel sampling benefits from model strength. 

\noindent \textit{Text2SQL:} IAD outperforms few-shot CoT and BoN using the same models (Gemini-1.5-Pro/Flash), achieving higher execution accuracy with fewer LLM calls (Figure~\ref{fig:candidate_generator_comparison}). Through just three rounds of feedback, IAD effectively corrects syntactic and semantic SQL errors, demonstrating feedback-driven compute efficiency. Compared against diverse baselines (DIN/DAIL/MCS/MAC-SQL) that avoid fine-tuning, IAD still leads (Table~\ref{tab:text-to-SQL}), reinforcing feedback’s central role. A similar trend is observed in Intercode Figure \ref{fig:intercode}, which shows sequential refinement with appropriate feedback outperforms baselines with low budget. 

\noindent\textit{Webshop:} IAD outperforms BoN-SC and strong baselines like GPT-4o and Gemini. For example, SR improves from 29.3\% (Gemini-1.5-Pro) and 41.09\% (BoN-SC + GPT-4o) to \textbf{44.68\%} (IAD + GPT-4o), a 3--4\% absolute gain. In weaker models like Gemini-1.5-Flash, iterative refinement significantly boosts performance where BoN plateaus.

\noindent \textbf{Key Insights :} We observe that feedback plays a crucial role in budget-controlled regimes—IAD consistently outperforms non feedback based baselines like BoN when compute (e.g., number of samples or model strength) is limited. However, as computational budget or model capability increases, the advantage of feedback-based iterative refinement diminishes.

\begin{figure*}[t]
    \centering
    \includegraphics[width=0.9\textwidth]{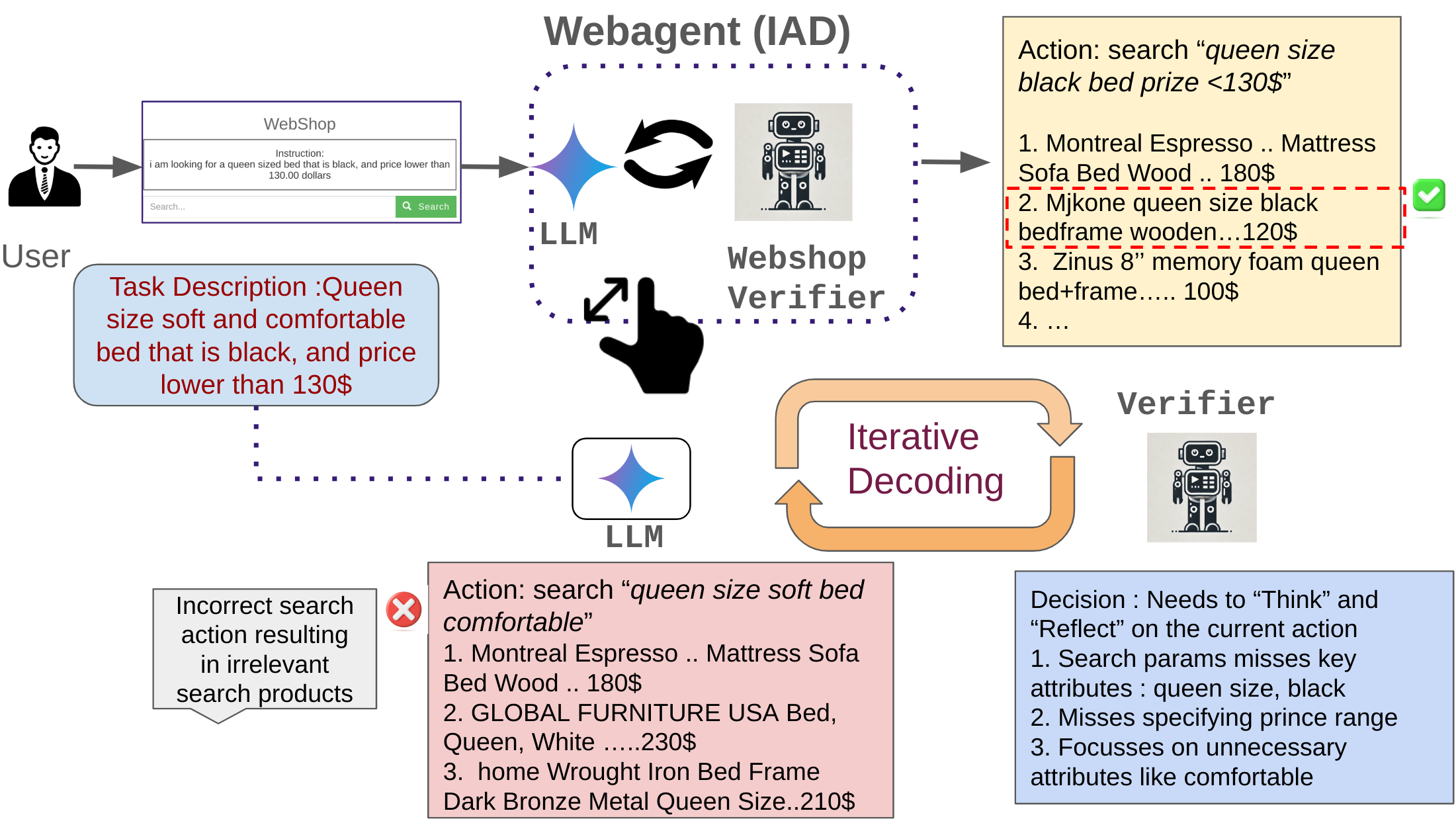}
     \caption{Qualitative illustration of IAD in improving the action taking capabilities of Webagent for Webshop environment \citep{yao2023webshopscalablerealworldweb} over direct generation, where IAD improves the search action by iteratively reviewing and refining the actions.} 
    \label{fig:webshop_example}
\end{figure*}

\subsection{Impact of adaptive feedback } 
Prior works lack a clear experimental setup to disentangle the true source of improvement in their approaches \citep{pair, flirt, wang2022self}. A key concern thus exists is whether the gains arise from sampling diversity or from the actual effectiveness of the feedback. However, most prior methods do not include comparisons against stochastic sampling baselines like BON, leaving it unclear whether their feedback mechanisms meaningfully contribute to performance improvements. To answer this, we conduct a controlled experiment to isolate the effect of adaptive feedback in IAD from sampling diversity. Specifically, we reduce the generation temperature (to 0.1, 0.05, and 0.0), thereby minimizing randomness in the generation process. We then evaluate the performance of both IAD and BON over multiple iterations under these low-stochasticity settings. \textit{Sketch2Code:} BoN saturates at layout score \textasciitilde 21.9 even with N=6, while IAD surpasses 26 with 6 iterations. Figure~\ref{fig:low_temp_abl1} confirms that adaptive refinement—not randomness—drives gains. \textit{Text2SQL:} Accuracy increases steadily with IAD even at low temperature settings (Figure~\ref{fig:low_temp_abl1}). This validates the role of feedback in semantic correction. 

\noindent \textbf{Key Insight}: The results demonstrate that the gains from IAD are not merely due to diversity in generation/sampling, but rather from the verifier-guided adaptive feedback-driven refinement. This also highlights a critical insight that when the diversity in the policy is low, feedback based sequential approaches can perform significantly better than sampling based BON approaches.

\subsection{Design and role of feedback form}
We investigate the impact of different forms of feedback on inference alignment. When feedback is textual, it is relatively easy to integrate as is. However, designing effective feedback from scalar rewards and preferences remains underexplored. In IAD, we focus on various methods to extract useful signals from scalar rewards and demonstrate that performance gains crucially depend on the design of proper feedback. We analyze both the scenarios: \textit{1. Textual feedback from LLM as a judge}  \textit{2. Scalar rewards. Majority of the prior sequential approaches including} \citep{refine1} have been designed with a focus on approach 1 i.e LLM as a judge and taking feedback from the same. However, we demonstrate that directly using off-the-shelf LLMs as judges fails to deliver consistent or monotonic improvements in complex scenarios (Figure \ref{fig:exm_plot}). Our experiments reveal that performance often plateaus, fluctuates, or even degrades across iterations. To highlight this, we include both quantitative and qualitative examples (Appendix) where self-LLM feedback is repetitive and uninformative, failing to identify or correct meaningful issues. This supports our central claim: for effective inference-time optimization, near-optimal verifiers are essential, especially in complex agentic scenarios like Sketch2code, text2sql. 

\paragraph{How to design feedback from a scalar reward?}  A crucial challenge lies in designing meaningful and informative feedback from scalar reward or preference, since the eventual feedback needs to be in textual form. Thus IAD is designed to extract as much signal as possible from the available feedback. Rather than relying solely on absolute reward scores, IAD transforms score-based or comparative feedback into structured guidance by identifying the best and worst responses at each iteration. These are explicitly fed back into the model through prompt conditioning, providing a clear directional signal for improvement. This process not only reinforces the distinction between good and bad outputs but also enables the use of dense feedback—even from weak or indirect supervision sources—turning minimal signals into effective updates. Intuitively, this approach is analogous to zeroth-order optimization, where two sampled values from the objective function are sufficient to guide the optimization process toward the maximum. Similarly, feedback on the best and worst responses helps steer the model. We compare IAD with the above feedback for Sketch2code and show that it performs almost comparable with LLM judge with the reference.

\begin{table}[ht]
\centering
\resizebox{0.45\textwidth}{!}{\begin{tabular}{lcc}
\hline
\textbf{Models} & \textbf{(PR)} & \textbf{(SR)} \\
\hline
Lemur-70b              & 71  & 11   \\
Mistral-7b             & 68.2  & 13.9 \\
Vicuna-13b-16k         & 73  & 21   \\
Gemini-1.5-Flash       & 71.3  & 26.5 \\
GPT-3.5-Turbo-16k      & 73  & 27   \\
Text-Davinci-003       & 72  & 29   \\
Gemini-1.5-Pro         & 73.5  & 29.3 \\
BON-SC + Gemini-1.5-Pro   & 74.12 & 30.31  \\
DeepSeek-67b           & 72  & 31   \\
GPT-3.5-Turbo          & 76  & 35   \\
\textbf{IAD + Gemini-1.5-Pro}   & {71}  & \textbf{38.3} \\
GPT-4                  & 75.8  & 38.5   \\
GPT-4o                  & 73.1  & 40.3   \\
BON-SC + GPT-4o   &  74.21 & 41.09\\
\textbf{IAD + GPT-4o}   & {74.6}  & \textbf{44.68} \\
\hline
\end{tabular}}
\caption{\label{tab:model_rates} \textbf{Webshop}- Progress Rate (PR) and Success Rate (SR) for Models in the Webshop Environment~\citep{yao2023webshopscalablerealworldweb}. Perform a comparison of IAD (with two models GPT-4o and Gemini-1.5-Pro) against SoTA baselines for the Webshop Leaderboard with the evaluation similar to followed in Agentboard \citep{agentboard}. For BON-SC, we generated N=3 responses, similarly we used three feedback steps for IAD.}
\end{table}

\subsection{Sensitivity to feedback quality} We study the sensitivity and robustness of inference alignment to feedback quality via controlled sparsity and noise in feedback signals. 

\paragraph{Sparse Rewards.} In many agentic tasks, dense reward signals may not be available. We study this setting in the context of Sketch2code \citep{li2024sketch2codeevaluatingvisionlanguagemodels} where we systematically sparsify our verification signal by varying the level of feedback sparsity and comparing the performance of IAD against the BON baseline with and without sparsity. Sparsification was achieved by providing feedback only when the verifier score exceeded a threshold; the higher the threshold, the sparser the reward signal. We define the sparsity levels and evaluate the performance of IAD versus BON with N=2 responses per prompt.

\noindent We observe that as the level of sparsity increases, the performance gap between IAD and BON narrows. Both inference-time approaches—BON and IAD—show a decline in performance, approaching the baseline under extreme sparsity conditions. The primary hypothesis behind the drop in IAD performance is that IAD relies on adaptive feedback: the LLM is conditioned on the best and worst responses from previous iterations, akin to a zeroth-order optimization method. Feedback on both ends (positive and negative) helps steer the model toward reward-maximizing generations through updates to the system prompt. However, under extreme sparsity, most responses—regardless of quality—receive zero reward. This results in random selection and noisy update directions, limiting IAD's ability to effectively adapt and improve.

\paragraph{Noisy Verification.} To further investigate the effect of noisy verification and how it impacts the performance of IAD compared to BON, we introduce varying levels of noise into the reward signal. Specifically, we add Gaussian noise with different variances to the reward scores, simulating imperfect or noisy verifier conditions. We then run both IAD and BON with N=2 responses per prompt and evaluate their performance under these noisy settings. This setup allows us to assess the robustness of IAD to reward noise and compare its stability and effectiveness relative to BON when the verification signal is noisy.

\noindent \textbf{Key Insights :} We observe trends similar to those in the sparse reward setting. Notably, IAD remains reasonably robust to mild noise in the reward signal. This is because IAD relies on adaptive feedback, where the LLM is conditioned on the best and worst responses from previous iterations, effectively leveraging pairwise comparisons rather than absolute scores. As long as the noise is limited, the relative preference between responses is preserved. For instance, if one layout score is 0.5 and another is 0.3, mild noise might shift them to 0.55 and 0.24, respectively, maintaining the same ordering. Therefore, IAD continues to improve under reasonable noise levels. However, as noise increases significantly, it can flip these preferences, leading to unstable updates and a decline in performance.

\section{Conclusion} 
In this work, we explore the underexplored role of feedback in inference-time alignment for black box AI agents. To understand the effect of feedback in inference alignment, we introduce Iterative Agent Decoding (IAD), a general sequential framework. Our study analyzes feedback through four lenses: (1) accuracy vs. compute trade-offs, (2) gains beyond sampling diversity, (3) feedback modality integration, and (4) sensitivity to feedback quality.
Empirically, we find that feedback is especially valuable under constrained budgets, achieving up to 10\% gains over feedback-free baselines. We also highlight the challenge of integrating diverse feedback modalities into sequential designs, which has not been critically explored in the literature. While textual feedback integrates naturally, representing scalar or preference signals remains an open challenge. We also observe that IAD’s benefits diminish under highly sparse or noisy feedback, underscoring the importance of feedback fidelity for effective alignment.

\section*{Limitations}
While IAD, our iterative decoding approach, improves upon prior baselines by better leveraging verifier feedback, it is inherently sequential, leading to increased user-facing latency compared to easily parallelizable BoN approaches. Addressing this tradeoff between quality improvement, computational cost, and user facing latency remains an important area for future research, which may require properly combining these techniques with adaptive stopping, controlled decoding~\citep{decoding1}, speculative decoding~\citep{leviathan2023fast}. Additionally, more efficient verifier-guided selection could improve the efficiency in iterative decoding for agentic tasks. As we learnt, the verifier (or judge) plays a crucial role in our approach. Thus a more concrete investigation and selection of a judge for these challenging tasks is a valid and crucial next step of our work.
\noindent
We highlight that this work is of an academic nature and has no direct or immediate harmful impacts on society. However, since this work deals with improving AI agents, it should be done under safety protocols and guidelines. We want to highlight that this study is limited to English language text primarily due to the nature of the open-source datasets used.

\bibliographystyle{abbrvnat}
\nobibliography*
\bibliography{custom}

\begin{thebibliography}{46}
\providecommand{\natexlab}[1]{#1}
\providecommand{\url}[1]{\texttt{#1}}
\expandafter\ifx\csname urlstyle\endcsname\relax
  \providecommand{\doi}[1]{doi: #1}\else
  \providecommand{\doi}{doi: \begingroup \urlstyle{rm}\Url}\fi

\bibitem[Amini et~al.(2024)Amini, Vieira, and Cotterell]{bon2}
A.~Amini, T.~Vieira, and R.~Cotterell.
\newblock Variational best-of-n alignment, 2024.
\newblock URL \url{https://arxiv.org/abs/2407.06057}.

\bibitem[Bai et~al.(2022)Bai, Jones, Ndousse, Askell, Chen, DasSarma, Drain, Fort, Ganguli, Henighan, et~al.]{rlhf3}
Y.~Bai, A.~Jones, K.~Ndousse, A.~Askell, A.~Chen, N.~DasSarma, D.~Drain, S.~Fort, D.~Ganguli, T.~Henighan, et~al.
\newblock Training a helpful and harmless assistant with reinforcement learning from human feedback.
\newblock \emph{arXiv preprint arXiv:2204.05862}, 2022.

\bibitem[Beirami et~al.(2024)Beirami, Agarwal, Berant, D'Amour, Eisenstein, Nagpal, and Suresh]{bon3}
A.~Beirami, A.~Agarwal, J.~Berant, A.~D'Amour, J.~Eisenstein, C.~Nagpal, and A.~T. Suresh.
\newblock Theoretical guarantees on the best-of-n alignment policy, 2024.
\newblock URL \url{https://arxiv.org/abs/2401.01879}.

\bibitem[Chao et~al.(2024)Chao, Robey, Dobriban, Hassani, Pappas, and Wong]{pair}
P.~Chao, A.~Robey, E.~Dobriban, H.~Hassani, G.~J. Pappas, and E.~Wong.
\newblock Jailbreaking black box large language models in twenty queries, 2024.
\newblock URL \url{https://arxiv.org/abs/2310.08419}.

\bibitem[Chen et~al.(2023)Chen, Lin, Schärli, and Zhou]{selfdebug}
X.~Chen, M.~Lin, N.~Schärli, and D.~Zhou.
\newblock Teaching large language models to self-debug, 2023.
\newblock URL \url{https://arxiv.org/abs/2304.05128}.

\bibitem[Chen et~al.(2025)Chen, Wang, Cao, Liu, Gao, Cui, Zhu, Ye, Tian, Liu, Gu, Wang, Li, Ren, Chen, Luo, Wang, Jiang, Wang, He, Shi, Zhang, Lv, Wang, Shao, Chu, Tu, He, Wu, Deng, Ge, Chen, Zhang, Wang, Dou, Lu, Zhu, Lu, Lin, Qiao, Dai, and Wang]{chen2025expandingperformanceboundariesopensource}
Z.~Chen, W.~Wang, Y.~Cao, Y.~Liu, Z.~Gao, E.~Cui, J.~Zhu, S.~Ye, H.~Tian, Z.~Liu, L.~Gu, X.~Wang, Q.~Li, Y.~Ren, Z.~Chen, J.~Luo, J.~Wang, T.~Jiang, B.~Wang, C.~He, B.~Shi, X.~Zhang, H.~Lv, Y.~Wang, W.~Shao, P.~Chu, Z.~Tu, T.~He, Z.~Wu, H.~Deng, J.~Ge, K.~Chen, K.~Zhang, L.~Wang, M.~Dou, L.~Lu, X.~Zhu, T.~Lu, D.~Lin, Y.~Qiao, J.~Dai, and W.~Wang.
\newblock Expanding performance boundaries of open-source multimodal models with model, data, and test-time scaling, 2025.
\newblock URL \url{https://arxiv.org/abs/2412.05271}.

\bibitem[DeepSeek-AI et~al.(2024)DeepSeek-AI, :, Bi, Chen, Chen, Chen, Dai, Deng, Ding, Dong, Du, Fu, Gao, Gao, Gao, Ge, Guan, Guo, Guo, Hao, Hao, He, Hu, Huang, Li, Li, Li, Li, Li, Liang, Lin, Liu, Liu, Liu, Liu, Liu, Liu, Lu, Lu, Luo, Ma, Nie, Pei, Piao, Qiu, Qu, Ren, Ren, Ruan, Sha, Shao, Song, Su, Sun, Sun, Tang, Wang, Wang, Wang, Wang, Wang, Wu, Wu, Xie, Xie, Xie, Xiong, Xu, Xu, Xu, Yang, You, Yu, Yu, Zhang, Zhang, Zhang, Zhang, Zhang, Zhang, Zhang, Zhang, Zhao, Zhao, Zhou, Zhou, Zhu, and Zou]{deepseekai2024deepseekllmscalingopensource}
DeepSeek-AI, :, X.~Bi, D.~Chen, G.~Chen, S.~Chen, D.~Dai, C.~Deng, H.~Ding, K.~Dong, Q.~Du, Z.~Fu, H.~Gao, K.~Gao, W.~Gao, R.~Ge, K.~Guan, D.~Guo, J.~Guo, G.~Hao, Z.~Hao, Y.~He, W.~Hu, P.~Huang, E.~Li, G.~Li, J.~Li, Y.~Li, Y.~K. Li, W.~Liang, F.~Lin, A.~X. Liu, B.~Liu, W.~Liu, X.~Liu, X.~Liu, Y.~Liu, H.~Lu, S.~Lu, F.~Luo, S.~Ma, X.~Nie, T.~Pei, Y.~Piao, J.~Qiu, H.~Qu, T.~Ren, Z.~Ren, C.~Ruan, Z.~Sha, Z.~Shao, J.~Song, X.~Su, J.~Sun, Y.~Sun, M.~Tang, B.~Wang, P.~Wang, S.~Wang, Y.~Wang, Y.~Wang, T.~Wu, Y.~Wu, X.~Xie, Z.~Xie, Z.~Xie, Y.~Xiong, H.~Xu, R.~X. Xu, Y.~Xu, D.~Yang, Y.~You, S.~Yu, X.~Yu, B.~Zhang, H.~Zhang, L.~Zhang, L.~Zhang, M.~Zhang, M.~Zhang, W.~Zhang, Y.~Zhang, C.~Zhao, Y.~Zhao, S.~Zhou, S.~Zhou, Q.~Zhu, and Y.~Zou.
\newblock Deepseek llm: Scaling open-source language models with longtermism, 2024.
\newblock URL \url{https://arxiv.org/abs/2401.02954}.

\bibitem[Fu et~al.(2024)Fu, Kim, Kim, Sohn, Logeswaran, Bae, and Lee]{agent1}
Y.~Fu, D.-K. Kim, J.~Kim, S.~Sohn, L.~Logeswaran, K.~Bae, and H.~Lee.
\newblock Autoguide: Automated generation and selection of context-aware guidelines for large language model agents, 2024.
\newblock URL \url{https://arxiv.org/abs/2403.08978}.

\bibitem[Gao et~al.(2023)Gao, Wang, Li, Sun, Qian, Ding, and Zhou]{gao2023text}
D.~Gao, H.~Wang, Y.~Li, X.~Sun, Y.~Qian, B.~Ding, and J.~Zhou.
\newblock Text-to-sql empowered by large language models: A benchmark evaluation.
\newblock \emph{arXiv preprint arXiv:2308.15363}, 2023.

\bibitem[Gao et~al.(2024)Gao, Liu, Li, Shi, Zhu, Wang, Li, Li, Hong, Luo, et~al.]{gao2024xiyan}
Y.~Gao, Y.~Liu, X.~Li, X.~Shi, Y.~Zhu, Y.~Wang, S.~Li, W.~Li, Y.~Hong, Z.~Luo, et~al.
\newblock Xiyan-sql: A multi-generator ensemble framework for text-to-sql.
\newblock \emph{arXiv preprint arXiv:2411.08599}, 2024.

\bibitem[{Google DeepMind}(2024)]{deepmind2024gemini2}
{Google DeepMind}.
\newblock Gemini 2.5: Pushing the frontier with advanced reasoning, multimodality, long context, and next generation agentic capabilities.
\newblock \url{https://storage.googleapis.com/deepmind-media/gemini/gemini_2_report.pdf}, 2024.

\bibitem[Gui et~al.(2025)Gui, G{\^a}rbacea, and Veitch]{gui2025bonbon}
L.~Gui, C.~G{\^a}rbacea, and V.~Veitch.
\newblock Bonbon alignment for large language models and the sweetness of best-of-n sampling.
\newblock \emph{Advances in Neural Information Processing Systems}, 37:\penalty0 2851--2885, 2025.

\bibitem[Jiang et~al.(2023)Jiang, Sablayrolles, Mensch, Bamford, Chaplot, de~las Casas, Bressand, Lengyel, Lample, Saulnier, Lavaud, Lachaux, Stock, Scao, Lavril, Wang, Lacroix, and Sayed]{jiang2023mistral7b}
A.~Q. Jiang, A.~Sablayrolles, A.~Mensch, C.~Bamford, D.~S. Chaplot, D.~de~las Casas, F.~Bressand, G.~Lengyel, G.~Lample, L.~Saulnier, L.~R. Lavaud, M.-A. Lachaux, P.~Stock, T.~L. Scao, T.~Lavril, T.~Wang, T.~Lacroix, and W.~E. Sayed.
\newblock Mistral 7b, 2023.
\newblock URL \url{https://arxiv.org/abs/2310.06825}.

\bibitem[Jiang et~al.(2024)Jiang, Dong, Wang, Fang, Shang, Li, Jin, and Jiao]{jiang2024selfplanningcodegenerationlarge}
X.~Jiang, Y.~Dong, L.~Wang, Z.~Fang, Q.~Shang, G.~Li, Z.~Jin, and W.~Jiao.
\newblock Self-planning code generation with large language models, 2024.
\newblock URL \url{https://arxiv.org/abs/2303.06689}.

\bibitem[Jinnai et~al.(2024)Jinnai, Morimura, Ariu, and Abe]{bon1}
Y.~Jinnai, T.~Morimura, K.~Ariu, and K.~Abe.
\newblock Regularized best-of-n sampling to mitigate reward hacking for language model alignment, 2024.
\newblock URL \url{https://arxiv.org/abs/2404.01054}.

\bibitem[Leviathan et~al.(2023)Leviathan, Kalman, and Matias]{leviathan2023fast}
Y.~Leviathan, M.~Kalman, and Y.~Matias.
\newblock Fast inference from transformers via speculative decoding, 2023.
\newblock URL \url{https://arxiv.org/abs/2211.17192}.

\bibitem[Li et~al.(2024{\natexlab{a}})Li, Hui, Qu, Yang, Li, Li, Wang, Qin, Geng, Huo, et~al.]{li2024can}
J.~Li, B.~Hui, G.~Qu, J.~Yang, B.~Li, B.~Li, B.~Wang, B.~Qin, R.~Geng, N.~Huo, et~al.
\newblock Can llm already serve as a database interface? a big bench for large-scale database grounded text-to-sqls.
\newblock \emph{Advances in Neural Information Processing Systems}, 36, 2024{\natexlab{a}}.

\bibitem[Li et~al.(2024{\natexlab{b}})Li, Zhang, and Yang]{li2024sketch2codeevaluatingvisionlanguagemodels}
R.~Li, Y.~Zhang, and D.~Yang.
\newblock Sketch2code: Evaluating vision-language models for interactive web design prototyping, 2024{\natexlab{b}}.
\newblock URL \url{https://arxiv.org/abs/2410.16232}.

\bibitem[Liu et~al.(2023)Liu, Li, Li, and Lee]{liu2023improvedllava}
H.~Liu, C.~Li, Y.~Li, and Y.~J. Lee.
\newblock Improved baselines with visual instruction tuning, 2023.

\bibitem[Liu et~al.(2024)Liu, Peng, Cao, Bo, Zhang, Zhang, Cheng, Wang, Yin, and Du]{agent2}
Y.~Liu, X.~Peng, J.~Cao, S.~Bo, Y.~Zhang, X.~Zhang, S.~Cheng, X.~Wang, J.~Yin, and T.~Du.
\newblock Tool-planner: Task planning with clusters across multiple tools, 2024.
\newblock URL \url{https://arxiv.org/abs/2406.03807}.

\bibitem[Ma et~al.(2024)Ma, Zhang, Zhu, Yang, Yang, Jin, Lan, Kong, and He]{agentboard}
C.~Ma, J.~Zhang, Z.~Zhu, C.~Yang, Y.~Yang, Y.~Jin, Z.~Lan, L.~Kong, and J.~He.
\newblock Agentboard: An analytical evaluation board of multi-turn llm agents, 2024.
\newblock URL \url{https://arxiv.org/abs/2401.13178}.

\bibitem[Maamari et~al.(2024)Maamari, Abubaker, Jaroslawicz, and Mhedhbi]{maamari2024death}
K.~Maamari, F.~Abubaker, D.~Jaroslawicz, and A.~Mhedhbi.
\newblock The death of schema linking? text-to-sql in the age of well-reasoned language models.
\newblock \emph{arXiv preprint arXiv:2408.07702}, 2024.

\bibitem[Madaan et~al.(2023)Madaan, Tandon, Gupta, Hallinan, Gao, Wiegreffe, Alon, Dziri, Prabhumoye, Yang, Gupta, Majumder, Hermann, Welleck, Yazdanbakhsh, and Clark]{refine1}
A.~Madaan, N.~Tandon, P.~Gupta, S.~Hallinan, L.~Gao, S.~Wiegreffe, U.~Alon, N.~Dziri, S.~Prabhumoye, Y.~Yang, S.~Gupta, B.~P. Majumder, K.~Hermann, S.~Welleck, A.~Yazdanbakhsh, and P.~Clark.
\newblock Self-refine: Iterative refinement with self-feedback, 2023.
\newblock URL \url{https://arxiv.org/abs/2303.17651}.

\bibitem[Mehrabi et~al.(2024)Mehrabi, Goyal, Dupuy, Hu, Ghosh, Zemel, Chang, Galstyan, and Gupta]{flirt}
N.~Mehrabi, P.~Goyal, C.~Dupuy, Q.~Hu, S.~Ghosh, R.~Zemel, K.-W. Chang, A.~Galstyan, and R.~Gupta.
\newblock Flirt: Feedback loop in-context red teaming, 2024.
\newblock URL \url{https://arxiv.org/abs/2308.04265}.

\bibitem[Mroueh(2024)]{mroueh2024informationtheoreticguaranteespolicy}
Y.~Mroueh.
\newblock Information theoretic guarantees for policy alignment in large language models, 2024.
\newblock URL \url{https://arxiv.org/abs/2406.05883}.

\bibitem[Mudgal et~al.(2024)Mudgal, Lee, Ganapathy, Li, Wang, Huang, Chen, Cheng, Collins, Strohman, Chen, Beutel, and Beirami]{decoding1}
S.~Mudgal, J.~Lee, H.~Ganapathy, Y.~Li, T.~Wang, Y.~Huang, Z.~Chen, H.-T. Cheng, M.~Collins, T.~Strohman, J.~Chen, A.~Beutel, and A.~Beirami.
\newblock Controlled decoding from language models, 2024.
\newblock URL \url{https://arxiv.org/abs/2310.17022}.

\bibitem[Nakano et~al.(2021)Nakano, Hilton, Balaji, Wu, Ouyang, Kim, Hesse, Jain, Kosaraju, Saunders, et~al.]{nakano2021webgpt}
R.~Nakano, J.~Hilton, S.~Balaji, J.~Wu, L.~Ouyang, C.~Kim, C.~Hesse, S.~Jain, V.~Kosaraju, W.~Saunders, et~al.
\newblock Webgpt: Browser-assisted question-answering with human feedback.
\newblock \emph{arXiv preprint arXiv:2112.09332}, 2021.

\bibitem[OpenAI et~al.(2024)OpenAI, Achiam, Adler, Agarwal, Ahmad, Akkaya, Aleman, Almeida, Altenschmidt, Altman, Anadkat, Avila, Babuschkin, Balaji, Balcom, Baltescu, Bao, Bavarian, Belgum, Bello, Berdine, Bernadett-Shapiro, Berner, Bogdonoff, Boiko, Boyd, Brakman, Brockman, Brooks, Brundage, Button, Cai, Campbell, Cann, Carey, Carlson, Carmichael, Chan, Chang, Chantzis, Chen, Chen, Chen, Chen, Chen, Chess, Cho, Chu, Chung, Cummings, Currier, Dai, Decareaux, Degry, Deutsch, Deville, Dhar, Dohan, Dowling, Dunning, Ecoffet, Eleti, Eloundou, Farhi, Fedus, Felix, Fishman, Forte, Fulford, Gao, Georges, Gibson, Goel, Gogineni, Goh, Gontijo-Lopes, Gordon, Grafstein, Gray, Greene, Gross, Gu, Guo, Hallacy, Han, Harris, He, Heaton, Heidecke, Hesse, Hickey, Hickey, Hoeschele, Houghton, Hsu, Hu, Hu, Huizinga, Jain, Jain, Jang, Jiang, Jiang, Jin, Jin, Jomoto, Jonn, Jun, Kaftan, Łukasz Kaiser, Kamali, Kanitscheider, Keskar, Khan, Kilpatrick, Kim, Kim, Kim, Kirchner, Kiros, Knight, Kokotajlo, Łukasz Kondraciuk, Kondrich,
  Konstantinidis, Kosic, Krueger, Kuo, Lampe, Lan, Lee, Leike, Leung, Levy, Li, Lim, Lin, Lin, Litwin, Lopez, Lowe, Lue, Makanju, Malfacini, Manning, Markov, Markovski, Martin, Mayer, Mayne, McGrew, McKinney, McLeavey, McMillan, McNeil, Medina, Mehta, Menick, Metz, Mishchenko, Mishkin, Monaco, Morikawa, Mossing, Mu, Murati, Murk, Mély, Nair, Nakano, Nayak, Neelakantan, Ngo, Noh, Ouyang, O'Keefe, Pachocki, Paino, Palermo, Pantuliano, Parascandolo, Parish, Parparita, Passos, Pavlov, Peng, Perelman, de~Avila Belbute~Peres, Petrov, de~Oliveira~Pinto, Michael, Pokorny, Pokrass, Pong, Powell, Power, Power, Proehl, Puri, Radford, Rae, Ramesh, Raymond, Real, Rimbach, Ross, Rotsted, Roussez, Ryder, Saltarelli, Sanders, Santurkar, Sastry, Schmidt, Schnurr, Schulman, Selsam, Sheppard, Sherbakov, Shieh, Shoker, Shyam, Sidor, Sigler, Simens, Sitkin, Slama, Sohl, Sokolowsky, Song, Staudacher, Such, Summers, Sutskever, Tang, Tezak, Thompson, Tillet, Tootoonchian, Tseng, Tuggle, Turley, Tworek, Uribe, Vallone, Vijayvergiya,
  Voss, Wainwright, Wang, Wang, Wang, Ward, Wei, Weinmann, Welihinda, Welinder, Weng, Weng, Wiethoff, Willner, Winter, Wolrich, Wong, Workman, Wu, Wu, Wu, Xiao, Xu, Yoo, Yu, Yuan, Zaremba, Zellers, Zhang, Zhang, Zhao, Zheng, Zhuang, Zhuk, and Zoph]{openai2024gpt4technicalreport}
OpenAI, J.~Achiam, S.~Adler, S.~Agarwal, L.~Ahmad, I.~Akkaya, F.~L. Aleman, D.~Almeida, J.~Altenschmidt, S.~Altman, S.~Anadkat, R.~Avila, I.~Babuschkin, S.~Balaji, V.~Balcom, P.~Baltescu, H.~Bao, M.~Bavarian, J.~Belgum, I.~Bello, J.~Berdine, G.~Bernadett-Shapiro, C.~Berner, L.~Bogdonoff, O.~Boiko, M.~Boyd, A.-L. Brakman, G.~Brockman, T.~Brooks, M.~Brundage, K.~Button, T.~Cai, R.~Campbell, A.~Cann, B.~Carey, C.~Carlson, R.~Carmichael, B.~Chan, C.~Chang, F.~Chantzis, D.~Chen, S.~Chen, R.~Chen, J.~Chen, M.~Chen, B.~Chess, C.~Cho, C.~Chu, H.~W. Chung, D.~Cummings, J.~Currier, Y.~Dai, C.~Decareaux, T.~Degry, N.~Deutsch, D.~Deville, A.~Dhar, D.~Dohan, S.~Dowling, S.~Dunning, A.~Ecoffet, A.~Eleti, T.~Eloundou, D.~Farhi, L.~Fedus, N.~Felix, S.~P. Fishman, J.~Forte, I.~Fulford, L.~Gao, E.~Georges, C.~Gibson, V.~Goel, T.~Gogineni, G.~Goh, R.~Gontijo-Lopes, J.~Gordon, M.~Grafstein, S.~Gray, R.~Greene, J.~Gross, S.~S. Gu, Y.~Guo, C.~Hallacy, J.~Han, J.~Harris, Y.~He, M.~Heaton, J.~Heidecke, C.~Hesse, A.~Hickey,
  W.~Hickey, P.~Hoeschele, B.~Houghton, K.~Hsu, S.~Hu, X.~Hu, J.~Huizinga, S.~Jain, S.~Jain, J.~Jang, A.~Jiang, R.~Jiang, H.~Jin, D.~Jin, S.~Jomoto, B.~Jonn, H.~Jun, T.~Kaftan, Łukasz Kaiser, A.~Kamali, I.~Kanitscheider, N.~S. Keskar, T.~Khan, L.~Kilpatrick, J.~W. Kim, C.~Kim, Y.~Kim, J.~H. Kirchner, J.~Kiros, M.~Knight, D.~Kokotajlo, Łukasz Kondraciuk, A.~Kondrich, A.~Konstantinidis, K.~Kosic, G.~Krueger, V.~Kuo, M.~Lampe, I.~Lan, T.~Lee, J.~Leike, J.~Leung, D.~Levy, C.~M. Li, R.~Lim, M.~Lin, S.~Lin, M.~Litwin, T.~Lopez, R.~Lowe, P.~Lue, A.~Makanju, K.~Malfacini, S.~Manning, T.~Markov, Y.~Markovski, B.~Martin, K.~Mayer, A.~Mayne, B.~McGrew, S.~M. McKinney, C.~McLeavey, P.~McMillan, J.~McNeil, D.~Medina, A.~Mehta, J.~Menick, L.~Metz, A.~Mishchenko, P.~Mishkin, V.~Monaco, E.~Morikawa, D.~Mossing, T.~Mu, M.~Murati, O.~Murk, D.~Mély, A.~Nair, R.~Nakano, R.~Nayak, A.~Neelakantan, R.~Ngo, H.~Noh, L.~Ouyang, C.~O'Keefe, J.~Pachocki, A.~Paino, J.~Palermo, A.~Pantuliano, G.~Parascandolo, J.~Parish, E.~Parparita,
  A.~Passos, M.~Pavlov, A.~Peng, A.~Perelman, F.~de~Avila Belbute~Peres, M.~Petrov, H.~P. de~Oliveira~Pinto, Michael, Pokorny, M.~Pokrass, V.~H. Pong, T.~Powell, A.~Power, B.~Power, E.~Proehl, R.~Puri, A.~Radford, J.~Rae, A.~Ramesh, C.~Raymond, F.~Real, K.~Rimbach, C.~Ross, B.~Rotsted, H.~Roussez, N.~Ryder, M.~Saltarelli, T.~Sanders, S.~Santurkar, G.~Sastry, H.~Schmidt, D.~Schnurr, J.~Schulman, D.~Selsam, K.~Sheppard, T.~Sherbakov, J.~Shieh, S.~Shoker, P.~Shyam, S.~Sidor, E.~Sigler, M.~Simens, J.~Sitkin, K.~Slama, I.~Sohl, B.~Sokolowsky, Y.~Song, N.~Staudacher, F.~P. Such, N.~Summers, I.~Sutskever, J.~Tang, N.~Tezak, M.~B. Thompson, P.~Tillet, A.~Tootoonchian, E.~Tseng, P.~Tuggle, N.~Turley, J.~Tworek, J.~F.~C. Uribe, A.~Vallone, A.~Vijayvergiya, C.~Voss, C.~Wainwright, J.~J. Wang, A.~Wang, B.~Wang, J.~Ward, J.~Wei, C.~Weinmann, A.~Welihinda, P.~Welinder, J.~Weng, L.~Weng, M.~Wiethoff, D.~Willner, C.~Winter, S.~Wolrich, H.~Wong, L.~Workman, S.~Wu, J.~Wu, M.~Wu, K.~Xiao, T.~Xu, S.~Yoo, K.~Yu, Q.~Yuan,
  W.~Zaremba, R.~Zellers, C.~Zhang, M.~Zhang, S.~Zhao, T.~Zheng, J.~Zhuang, W.~Zhuk, and B.~Zoph.
\newblock Gpt-4 technical report, 2024.
\newblock URL \url{https://arxiv.org/abs/2303.08774}.

\bibitem[Ouyang et~al.(2022{\natexlab{a}})Ouyang, Wu, Jiang, Almeida, Wainwright, Mishkin, Zhang, Agarwal, Slama, Ray, Schulman, Hilton, Kelton, Miller, Simens, Askell, Welinder, Christiano, Leike, and Lowe]{rlhf1}
L.~Ouyang, J.~Wu, X.~Jiang, D.~Almeida, C.~L. Wainwright, P.~Mishkin, C.~Zhang, S.~Agarwal, K.~Slama, A.~Ray, J.~Schulman, J.~Hilton, F.~Kelton, L.~Miller, M.~Simens, A.~Askell, P.~Welinder, P.~Christiano, J.~Leike, and R.~Lowe.
\newblock Training language models to follow instructions with human feedback, 2022{\natexlab{a}}.
\newblock URL \url{https://arxiv.org/abs/2203.02155}.

\bibitem[Ouyang et~al.(2022{\natexlab{b}})Ouyang, Wu, Jiang, Almeida, Wainwright, Mishkin, Zhang, Agarwal, Slama, Ray, Schulman, Hilton, Kelton, Miller, Simens, Askell, Welinder, Christiano, Leike, and Lowe]{rlhf2}
L.~Ouyang, J.~Wu, X.~Jiang, D.~Almeida, C.~L. Wainwright, P.~Mishkin, C.~Zhang, S.~Agarwal, K.~Slama, A.~Ray, J.~Schulman, J.~Hilton, F.~Kelton, L.~Miller, M.~Simens, A.~Askell, P.~Welinder, P.~Christiano, J.~Leike, and R.~Lowe.
\newblock Training language models to follow instructions with human feedback, 2022{\natexlab{b}}.

\bibitem[Pourreza and Rafiei(2024)]{pourreza2024din}
M.~Pourreza and D.~Rafiei.
\newblock Din-sql: Decomposed in-context learning of text-to-sql with self-correction.
\newblock \emph{Advances in Neural Information Processing Systems}, 36, 2024.

\bibitem[Pourreza et~al.(2024)Pourreza, Li, Sun, Chung, Talaei, Kakkar, Gan, Saberi, Ozcan, and Arik]{pourreza2024chase}
M.~Pourreza, H.~Li, R.~Sun, Y.~Chung, S.~Talaei, G.~T. Kakkar, Y.~Gan, A.~Saberi, F.~Ozcan, and S.~O. Arik.
\newblock Chase-sql: Multi-path reasoning and preference optimized candidate selection in text-to-sql.
\newblock \emph{arXiv preprint arXiv:2410.01943}, 2024.

\bibitem[Robeyns et~al.(2025)Robeyns, Szummer, and Aitchison]{robeyns2025selfimprovingcodingagent}
M.~Robeyns, M.~Szummer, and L.~Aitchison.
\newblock A self-improving coding agent, 2025.
\newblock URL \url{https://arxiv.org/abs/2504.15228}.

\bibitem[Setlur et~al.(2025)Setlur, Rajaraman, Levine, and Kumar]{setlur2025scalingtesttimecomputeverification}
A.~Setlur, N.~Rajaraman, S.~Levine, and A.~Kumar.
\newblock Scaling test-time compute without verification or rl is suboptimal, 2025.
\newblock URL \url{https://arxiv.org/abs/2502.12118}.

\bibitem[Talaei et~al.(2024)Talaei, Pourreza, Chang, Mirhoseini, and Saberi]{talaei2024chess}
S.~Talaei, M.~Pourreza, Y.-C. Chang, A.~Mirhoseini, and A.~Saberi.
\newblock Chess: Contextual harnessing for efficient sql synthesis.
\newblock \emph{arXiv preprint arXiv:2405.16755}, 2024.

\bibitem[Team et~al.(2024)Team, Georgiev, Lei, Burnell, Bai, Gulati, Vincent, and et~al.]{geminiteam2024gemini15unlockingmultimodal}
G.~Team, P.~Georgiev, V.~I. Lei, R.~Burnell, L.~Bai, A.~Gulati, D.~Vincent, and et~al.
\newblock Gemini 1.5: Unlocking multimodal understanding across millions of tokens of context, 2024.
\newblock URL \url{https://arxiv.org/abs/2403.05530}.

\bibitem[Verdun et~al.(2025)Verdun, Oesterling, Lakkaraju, and Calmon]{verdun2025soft}
C.~M. Verdun, A.~Oesterling, H.~Lakkaraju, and F.~P. Calmon.
\newblock Soft best-of-n sampling for model alignment.
\newblock \emph{arXiv preprint arXiv:2505.03156}, 2025.

\bibitem[Wang et~al.(2023)Wang, Ren, Yang, Liang, Bai, Zhang, Yan, and Li]{wang2023mac}
B.~Wang, C.~Ren, J.~Yang, X.~Liang, J.~Bai, Q.-W. Zhang, Z.~Yan, and Z.~Li.
\newblock Mac-sql: Multi-agent collaboration for text-to-sql.
\newblock \emph{arXiv preprint arXiv:2312.11242}, 2023.

\bibitem[Wang et~al.(2022)Wang, Wei, Schuurmans, Le, Chi, Narang, Chowdhery, and Zhou]{wang2022self}
X.~Wang, J.~Wei, D.~Schuurmans, Q.~Le, E.~Chi, S.~Narang, A.~Chowdhery, and D.~Zhou.
\newblock Self-consistency improves chain of thought reasoning in language models.
\newblock \emph{arXiv preprint arXiv:2203.11171}, 2022.

\bibitem[Xu et~al.(2024)Xu, Su, Xing, Mi, Liu, Shi, Hui, Zhou, Liu, Xie, Cheng, Zhao, Kong, Wang, Xiong, and Yu]{xu2024lemurharmonizingnaturallanguage}
Y.~Xu, H.~Su, C.~Xing, B.~Mi, Q.~Liu, W.~Shi, B.~Hui, F.~Zhou, Y.~Liu, T.~Xie, Z.~Cheng, S.~Zhao, L.~Kong, B.~Wang, C.~Xiong, and T.~Yu.
\newblock Lemur: Harmonizing natural language and code for language agents, 2024.
\newblock URL \url{https://arxiv.org/abs/2310.06830}.

\bibitem[Yang et~al.(2024)Yang, Wang, Lu, Liu, Le, Zhou, and Chen]{llmopt}
C.~Yang, X.~Wang, Y.~Lu, H.~Liu, Q.~V. Le, D.~Zhou, and X.~Chen.
\newblock Large language models as optimizers, 2024.
\newblock URL \url{https://arxiv.org/abs/2309.03409}.

\bibitem[Yang et~al.(2023{\natexlab{a}})Yang, Prabhakar, Narasimhan, and Yao]{intercode}
J.~Yang, A.~Prabhakar, K.~Narasimhan, and S.~Yao.
\newblock Intercode: Standardizing and benchmarking interactive coding with execution feedback, 2023{\natexlab{a}}.
\newblock URL \url{https://arxiv.org/abs/2306.14898}.

\bibitem[Yang et~al.(2023{\natexlab{b}})Yang, Prabhakar, Narasimhan, and Yao]{tryagain}
J.~Yang, A.~Prabhakar, K.~Narasimhan, and S.~Yao.
\newblock Intercode: Standardizing and benchmarking interactive coding with execution feedback, 2023{\natexlab{b}}.
\newblock URL \url{https://arxiv.org/abs/2306.14898}.

\bibitem[Yao et~al.(2023)Yao, Chen, Yang, and Narasimhan]{yao2023webshopscalablerealworldweb}
S.~Yao, H.~Chen, J.~Yang, and K.~Narasimhan.
\newblock Webshop: Towards scalable real-world web interaction with grounded language agents, 2023.
\newblock URL \url{https://arxiv.org/abs/2207.01206}.

\bibitem[Ye et~al.(2023)Ye, Chen, Xu, Zu, Shao, Liu, Cui, Zhou, Gong, Shen, Zhou, Chen, Gui, Zhang, and Huang]{ye2023comprehensivecapabilityanalysisgpt3}
J.~Ye, X.~Chen, N.~Xu, C.~Zu, Z.~Shao, S.~Liu, Y.~Cui, Z.~Zhou, C.~Gong, Y.~Shen, J.~Zhou, S.~Chen, T.~Gui, Q.~Zhang, and X.~Huang.
\newblock A comprehensive capability analysis of gpt-3 and gpt-3.5 series models, 2023.
\newblock URL \url{https://arxiv.org/abs/2303.10420}.

\bibitem[Zheng et~al.(2023)Zheng, Chiang, Sheng, Zhuang, Wu, Zhuang, Lin, Li, Li, Xing, Zhang, Gonzalez, and Stoica]{zheng2023judgingllmasajudgemtbenchchatbot}
L.~Zheng, W.-L. Chiang, Y.~Sheng, S.~Zhuang, Z.~Wu, Y.~Zhuang, Z.~Lin, Z.~Li, D.~Li, E.~P. Xing, H.~Zhang, J.~E. Gonzalez, and I.~Stoica.
\newblock Judging llm-as-a-judge with mt-bench and chatbot arena, 2023.
\newblock URL \url{https://arxiv.org/abs/2306.05685}.

\end{thebibliography}

\clearpage
\appendix

\section{Detailed Environment Description}
\textbf{1. Text-to-SQL} \label{sec:appendix:text2sql}
Text-to-SQL serves as a critical interface between natural language and structured query languages by enabling users to translate natural language queries into executable SQL commands. This functionality empowers individuals without SQL expertise to interact with complex databases, thereby facilitating data exploration, informed decision-making, automated analytics, and advanced feature extraction for machine learning. Generally, a Text-to-SQL system receives a natural language question and any pertinent metadata about the tables and columns, which serves as external knowledge to aid in database comprehension. Consequently, such systems are responsible not only for interpreting user intent and identifying relevant information from a potentially vast set of tables and columns but also for generating SQL queries that may include multiple conditions—a process that is inherently reasoning-intensive. To evaluate our proposed framework, we employ the BIRD benchmark \citep{li2024can}, a challenging and widely used dataset in the Text-to-SQL domain. BIRD comprises an extensive collection of 12,751 unique question-SQL pairs drawn from 95 large databases with a total size of 33.4 GB. The benchmark spans more than 37 professional domains, including blockchain, hockey, healthcare, and education, making it a comprehensive resource for assessing the robustness and generalizability of Text-to-SQL systems. The primary metric for model comparison in this domain is execution accuracy (EX), where the ground truth SQL query and the predicted SQL query are both executed over the target database, if they both generate same sets of results the accuracy for the predicted SQL query is considered as accurate. 

\begin{figure}[!ht]
    \centering
    \begin{subfigure}[b]{0.48\textwidth}
        \centering
        \includegraphics[width=\textwidth]{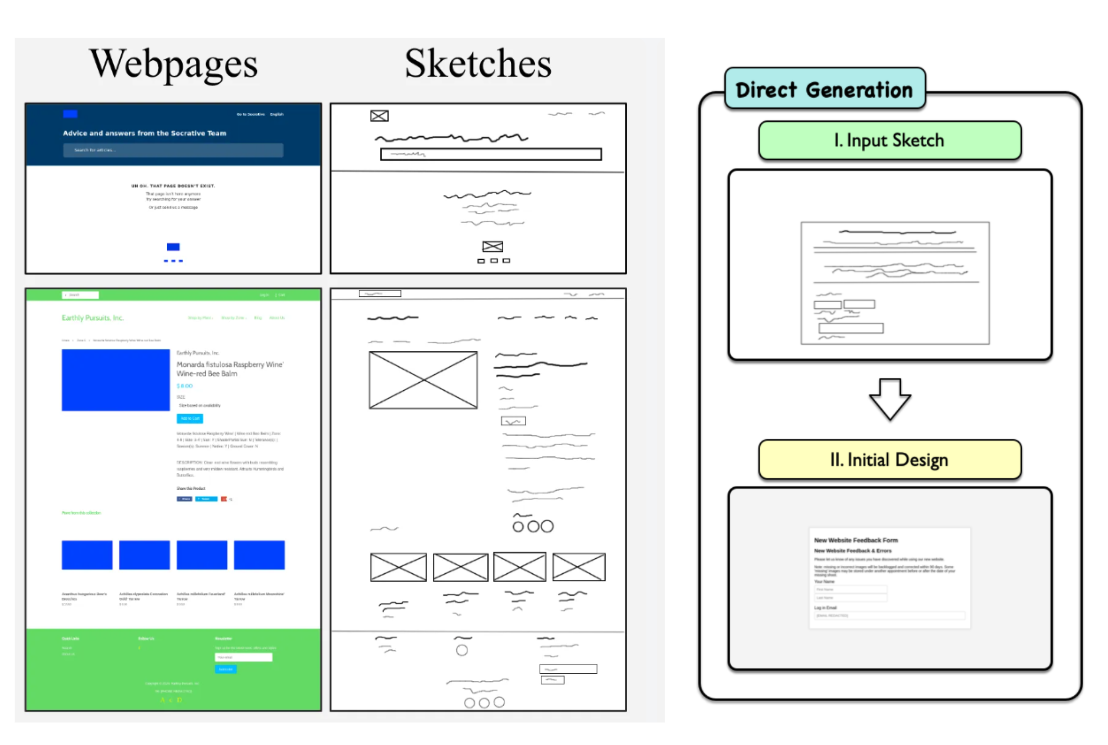} 
        \caption{Sketch2Code Environment}
        \label{fig:sketch2code_env1}
    \end{subfigure}
    \begin{subfigure}[b]{0.48\textwidth}
        \centering
        \includegraphics[width=\textwidth]{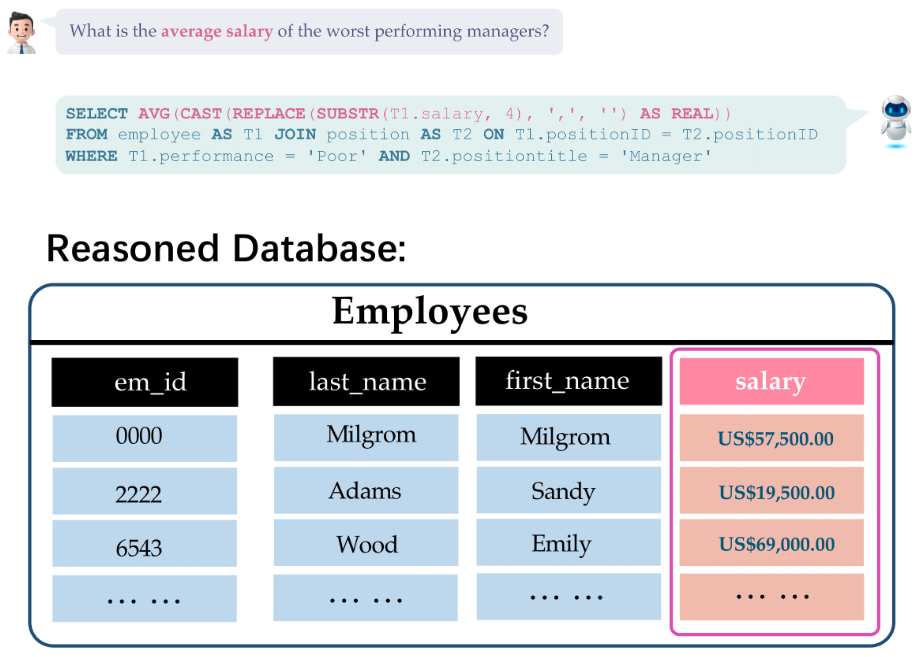} 
        \caption{Text-to-SQL Environment}
        \label{fig:text2sql_env2}
    \end{subfigure}
    \begin{subfigure}[b]{0.47\textwidth}
        \centering
        \includegraphics[width=\textwidth]{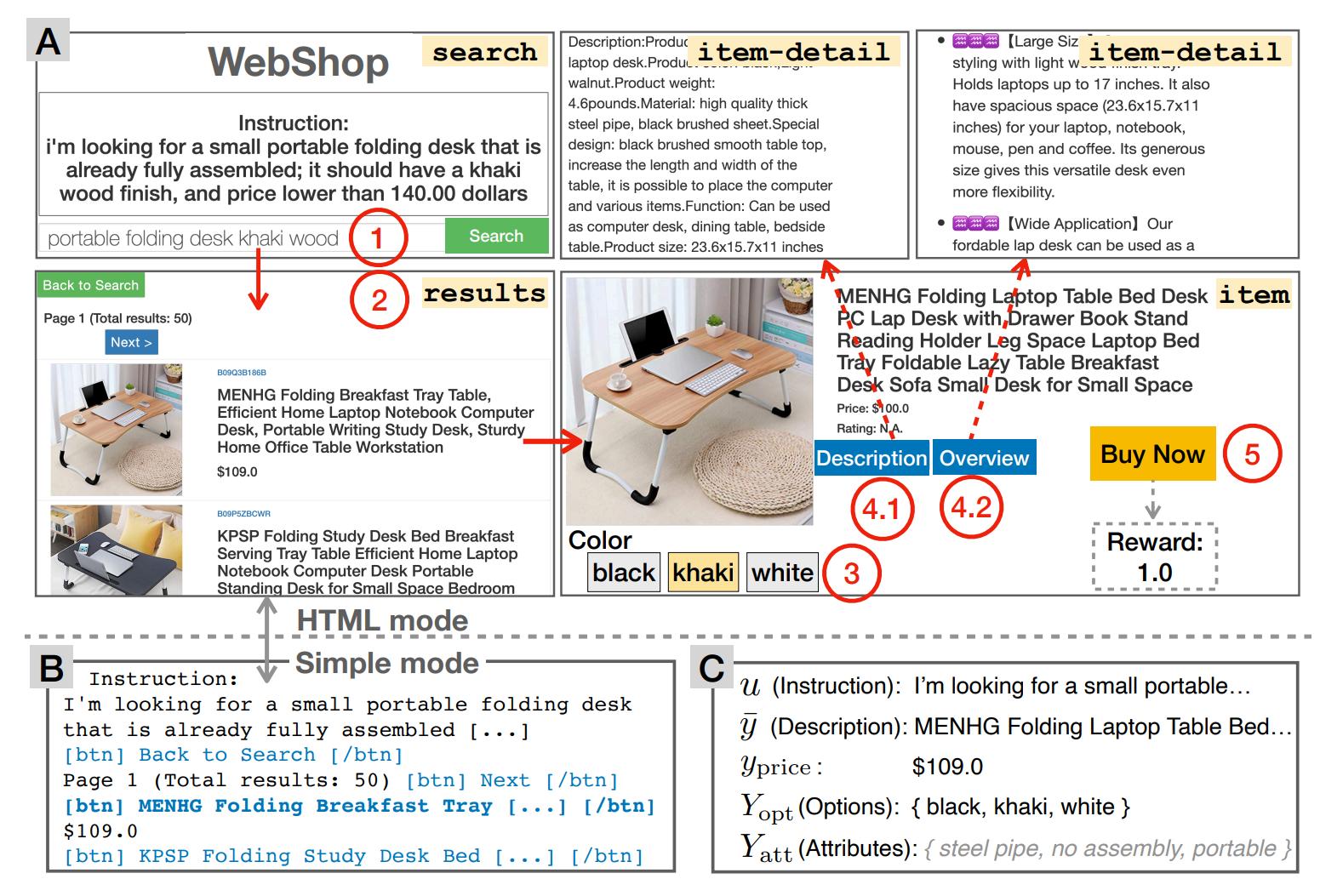} 
        \caption{Webshop Environment}
        \label{fig:webshop_env3}
    \end{subfigure}
    \caption{These three figures given an overview of three diverse and challenging agentic tasks that we consider to evaluate the performance of agents with our proposed approach vs baselines -(a) Sketch2code\citep{li2024sketch2codeevaluatingvisionlanguagemodels} (b)Text2SQL \citep{li2024can} and (c) Webshop \citep{yao2023webshopscalablerealworldweb}}
    \label{fig:mainfig_env}
\end{figure}

\vspace{2mm}
\noindent
\textbf{2. Sketch2code}: Sketch2code \citep{li2024sketch2codeevaluatingvisionlanguagemodels} challenges and evaluates the multi-modal capabilities of agent where the objective is transform wireframe-style rough userk sketches into functional HTML prototypes with embedded CSS. Sketch2Code uniquely tests multi-modality, requiring structured code generation from imprecise visual input, often leading to misaligned text, incorrect spacing, and structural inconsistencies. This leads to challenges such as misaligned text, incorrect spacing, missing components, structural inconsistencies, making it an extremely challenging benchmark for multimodal LLMs. The complexity of this task arises from: ambiguity in hand-drawn sketches, where component boundaries, spacing, and positioning are not precisely defined. The evaluation of the generation is done primarily with three key metrics : \textit{Layout Similarity, Text IOU, Image IOU}. Layout Similarity (IoU-based metrics): Intersection-over-Union (IoU) is computed for different UI components (e.g., buttons, images, text blocks) to measure how well their positions match the reference. Intersection-over-Union (IoU) is computed for different UI components (e.g., buttons, images, text blocks) to measure how well their positions match the reference implementation. Text-IOU similarly measures how accurately the generated text aligns with the reference design. Image IOU uses CLIP embeddings to compare the visual appearance of the generated webpage with the reference design and evaluates color similarity, element positioning, and component rendering. These metrics provide a reliable way to measure the quality of the generated response and strongly correlates with human judgement. Evaluations are also done with LLM as a judge to compare the performance.

\noindent
\textbf{3. Webshop}  is a large-scale, web-based interactive environment designed to test an AI agent's capability to perform sequential decision-making in an online shopping scenario under sparse feedback \citep{yao2023webshopscalablerealworldweb}. The environment is modeled as a partially observable Markov decision process, where the agent navigates a simulated e-commerce platform to fulfill a user's product request based on natural language instructions. At each step, the agent receives an observation in the form of a webpage—such as search results, product details, or checkout options—and must decide on an action, including searching for a product, clicking on an item, or selecting options. The evaluation is based on success rate (SR), which measures whether the agent successfully selects a product that matches all specified criteria (attributes, price, and options), and task score, which represents the overall alignment of the final selection with the given instruction. The WebShop environment presents significant challenges, including sparse rewards (since feedback is only provided at the end of an episode), the need for strategic backtracking and exploration, and handling noisy or ambiguous natural language instructions. This setup makes WebShop a rigorous benchmark for evaluating long-horizon reasoning, language understanding, and decision-making in real-world-like online navigation scenarios.

\subsection{Limitation of Single-turn Approach}
In this section we characterize the performance gap $\Delta$ as the difference between the reward the optimal or ground-truth agent is achieving vs the reward achieved by the reference achieved by the reference agent policy~\citep{mroueh2024informationtheoreticguaranteespolicy}:
\begin{align*}
    \Delta &= \mathbb{E}_{y \sim \pi^*(\cdot|x)}[R(x, y)] - \mathbb{E}_{y \sim \pi_0(\cdot|x)}[R(x, y)] \\ \nonumber
    &\leq \sup_{R \in \mathcal{R}} \mathbb{E}_{y \sim \pi^*(\cdot|x)}[R(x, y)] - \mathbb{E}_{y \sim \pi_0(\cdot|x)}[R(x, y)] \\ \nonumber
    &\leq \|R\|_{\text{max}} d_{\text{TV}}(\pi^*(\cdot|x), \pi_0(\cdot|x)), \nonumber
\end{align*}
where $R(x, y)$ represents the reward function measuring the quality of the generated response, and $d_{\text{TV}}(\pi^*(\cdot|x), \pi_0(\cdot|x))$ is the total variation (TV) distance between the optimal policy $\pi^*(\cdot|x)$ and the reference policy $\pi_0(\cdot|x)$. This result demonstrates that the performance gap $\Delta$ is inherently limited by the quality of the reference agent policy $\pi_0(\cdot|x)$, as measured by its divergence from the optimal policy. Thus, if $\pi_0(\cdot|x)$ is close to $\pi^*(\cdot|x)$ (in terms of TV distance), the performance gap will be small, resulting in near-optimal responses and viceversa.
\begin{figure*}[ht]
    \centering
    \begin{subfigure}[b]{\textwidth}
        \centering
        \includegraphics[width=\textwidth]{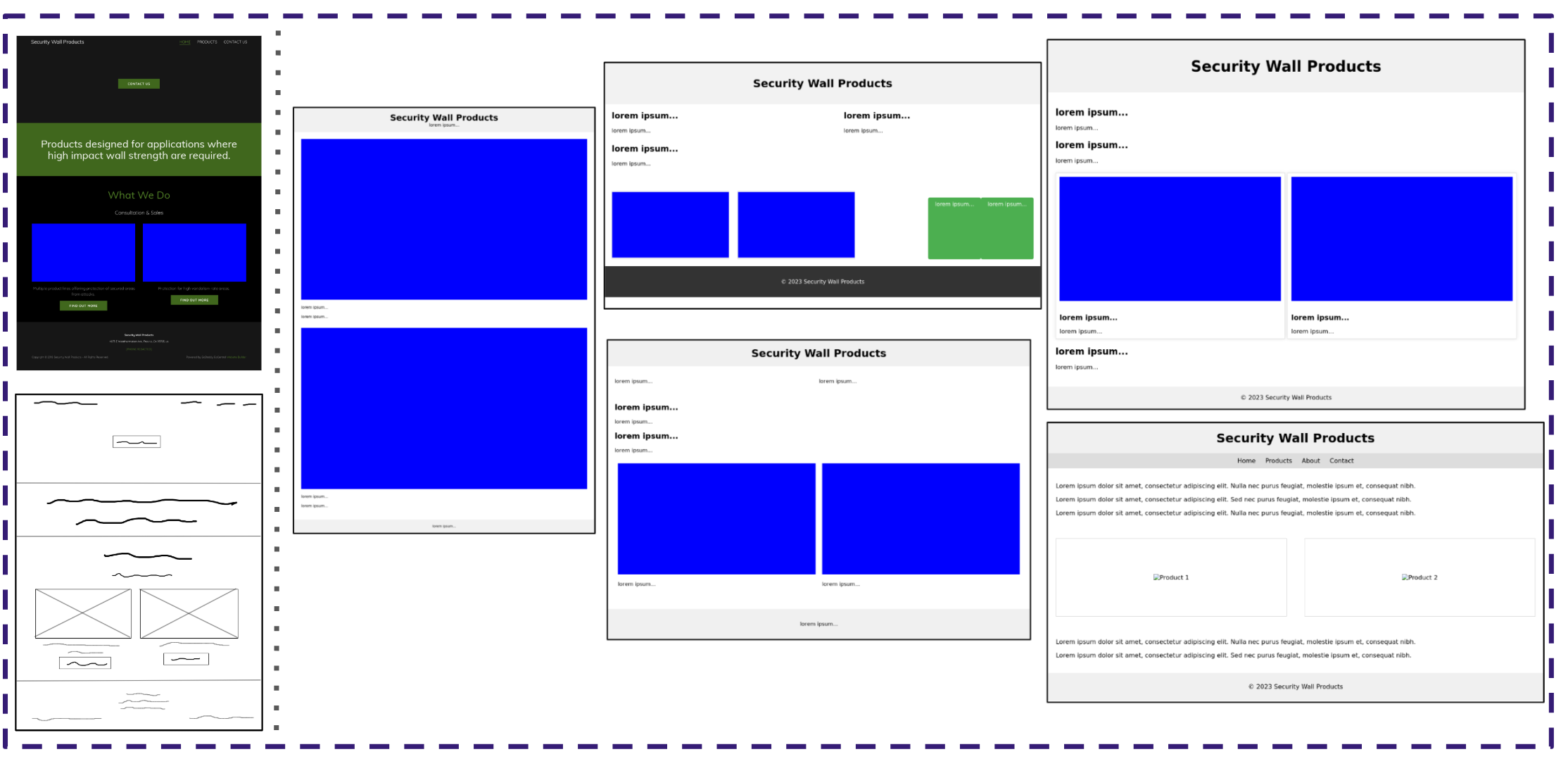} 
        \caption{}
    \end{subfigure}
    \caption{\textbf{Sketch2code}: Qualitative evaluation of the generated HTMLs with BON sampling (N=4) corresponding to the user-sketch (left-bottom) and reference html (left-top). The figures show that BON performs much better in matching the reference HTML but still misses specific properties like rectangular structure, position of text, relative positioning of smaller blocks etc.}
    \label{fig:sketc2code_qual1}
\end{figure*}

\section{Detailed Experimental Analysis}

\subsection{Text-to-SQL Detailed Results and Analysis}

In this section, we detail the experiments conducted on the BIRD text-to-SQL benchmark \citep{li2024can}. For these experiments, we employed the Gemini-1.5-pro and Gemini-1.5-flash models both to generate actions at each state and as judge models to predict the reward. At each state, the LLM is provided with the database schema and the user's query, based on which it generates a draft SQL query. This draft query is then evaluated by the judge model, which also produces feedback on how to improve the draft. The LLM uses this feedback to generate a revised query, establishing a self-correction loop. Finally, the answer with the highest reward value is selected as the candidate output. This process can be repeated to generate multiple candidate SQL queries. We then apply self-consistency \citep{wang2022self} by executing all candidate queries over the database, grouping them based on their execution results, and selecting a query from the largest result cluster as the final answer. We first compare our proposed method with the widely used few-shot prompting approach in terms of Pass@k performance and final accuracy after self-consistency (Majority@K) using execution accuracy as the metric in order to demonstrate that using our method we can generate a pool of candidates with a higher quality. Subsequently, we compare our approach with the best-of-N approach, which is one of the strong baselines as a test-time compute approach to demonstrate the effectiveness of the proposed framework. Finally, we compare our method with all previously proposed test-time methods on the BIRD development set benchmark, excluding works that rely heavily on fine-tuning LLMs \citep{pourreza2024chase, talaei2024chess, maamari2024death, gao2024xiyan} for a fair comparison.

\begin{figure*}[!t]
    \centering
    \includegraphics[width=\textwidth]{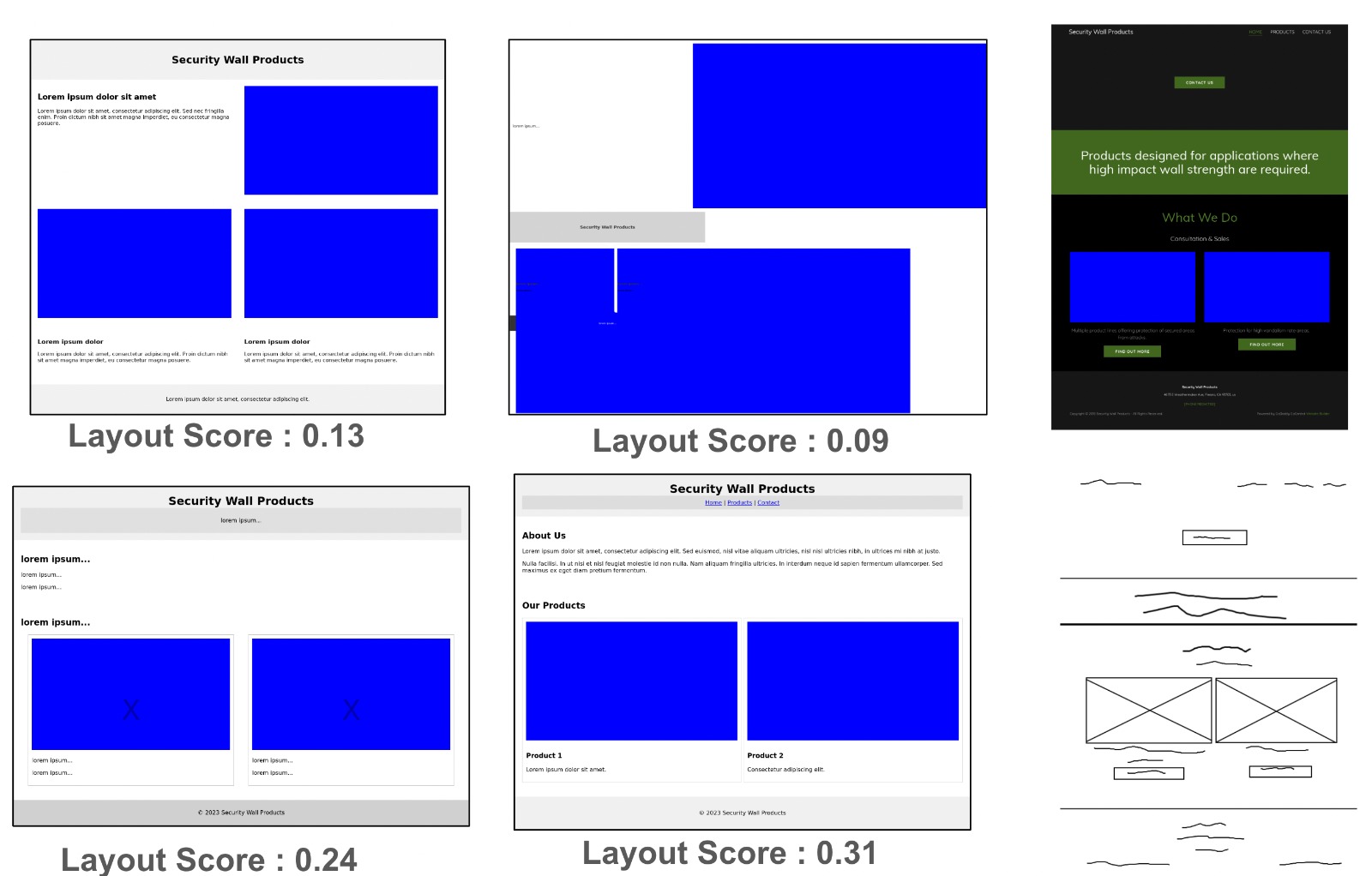}
     \caption{\textbf{Sketch2code}: Provides a qualitative verification of layout score as a metric and corresponding correlation to human judgement. It is evident that HTMLs with higher match with the reference layout (right-top) and user sketch(right-bottom) has higher layout score and vice-versa showing that its a valid metric.} 
    \label{fig:sk2code_metric_ls_eval}
\end{figure*}

\begin{tcolorbox}[floatplacement=t,float, width=\columnwidth, colback=gray!10, colframe=gray, title=\textbf{Sketch2code : Oracle Judge Prompt for providing Feedback}]
    \textbf{Judge Prompt for Sketch2code (similar to \citep{li2024sketch2codeevaluatingvisionlanguagemodels}): Act as if you are a front-end designer working with a code agent to implement an HTML webpage. You are provided with two HTMLs: the first is the reference webpage, and the second one is the current implementation from the code agent. The task is to carefully compare the agent's implementation against the reference webpage and provide feedback to help the agent make its implementation closer to the reference webpage. Your feedback should be specific to the differences in layouts and visual components on the two webpages. Don't focus on the style components too much, and focus on layout similarity and visual match with the reference webpage.}
\end{tcolorbox}

\begin{tcolorbox}[floatplacement=t,float, width=\columnwidth, colback=gray!10, colframe=gray, title=\textbf{Sketch2code : Feedback from LLM as a Judge (Self Verification)}]
    \textbf{Feedback provided from Self-LLM: Iter1: The HTML structure and CSS styling do not reflect the provided wireframe.  Iter2: The layout needs to be revised to accurately represent the sketch's two-column image section and the distribution of text blocks. Iter 3: The layout of the text blocks and image containers does not accurately reflect the provided wireframe.  The layout uses flexbox but doesn't accurately reflect the sketch's proportions and image placement.  The large image should be centered and the smaller images should be positioned to the left of their respective text blocks. Iter4: Implement a more precise grid-based layout using CSS grid or flexbox to achieve the correct positioning and sizing of all elements. Iter 5: The provided HTML closely resembles the wireframe but still needs significant layout adjustments.  Use CSS Grid to precisely position and size all elements according to the wireframe's proportions.}
\end{tcolorbox}

\begin{table}[ht]
\centering
\resizebox{0.49\textwidth}{!}{
\begin{tabular}{lccc}
\toprule
\textbf{Model}  & \textbf{Layout.} & \textbf{Txt IoU} & \textbf{Img IoU} \\
\toprule
\multicolumn{4}{c}{\textbf{Single-Turn Approaches}} \\ 
\toprule
InternVL2-8b$^*$    & 4.01  & 4.89 & 1.41  \\
Llava-1.6-8b$^*$    & 8.01  & 9.26 & 1.95  \\
Claude-3-Sonnet$^*$ & 14.22  & 15.85 & 6.62 \\
GPT-4o-Mini$^*$    & 16.29  & 20.84 & 0.72  \\
Claude-3-Opus$^*$   & 17.11  & 18.09 & 8.32 \\
Claude-3-Haiku$^*$  & 17.52  & 20.60 & 2.72  \\
Gemini-1.5-Flash & 17.9  & 17.50 & 10.77  \\
Gemini-1.5-Pro  & 18.25  & 18.20 & 12.69 \\
GPT-4o$^*$          & 19.20  & 17.12 & 16.19 \\
Gemini-1.5-Flash (CoT) & 19.84  & 19.13 & 10.02  \\
Claude-3.5-Sonnet$^*$  & 22.26  & 25.33 & 9.21 \\
\toprule
\multicolumn{4}{c}{\textbf{Multi-Turn Approaches (\texttt{Gemini-1.5-Flash})}} \\ 
\toprule
Sk2code (N=2)$^{**}$   & 19.41  & 20.45 & 11.81  \\
Self-Refine (N=2)    & 19.51  & 19.35 & 10.71 \\
BON (N=2)          & 21.45  & 20.1 & 13.5  \\
\textbf{IAD (N=2)}   & \textbf{24.78}  & {23.01} & {15.29}  \\
\textbf{IAD-fb (N=2, K=2)}$^{**}$   & \textbf{24.86}  & {23.4} & {14.69} \\
\hline
Self-Refine (N=4)    & 19.97  & 19.11 & 11.74 \\
Sk2code (N=4)$^{**}$  & 20.41  & 21.46 & 12.67  \\
BON (N=4)          & 24.02  & 22.59 & 15.91  \\
\textbf{IAD (N=4)}   & \textbf{25.97}  & {24.13} & {16.98}  \\
\textbf{IAD-fb (N=4, K=4)}$^{**}$   & \textbf{26.61}  & {24.62} & {17.36} \\
\hline
Self-Refine (N=6)    & 19.89  & 18.91 & 11.61 \\
Sk2code (N=6)$^{**}$   & 21.43  & 21.53 & 13.78  \\
BON (N=6)          & 25.75  & 22.91 & 17.67  \\
\textbf{IAD (N=6)}   & \textbf{26.75}  & {24.91} & {19.12}  \\
\textbf{IAD-fb (N=6, K=6)}$^{**}$   & \textbf{27.95}  & {24.99} & {19.01}  \\
\toprule
\end{tabular}
}
\caption{\textbf{Sketch2Code}: Performance comparison between single-turn and multi-response generation approaches. For each of the multi-response generation method Layout score acts as the reference metric (temperature =0.6). Table demonstrate that IAD (Ours) consistently outperform SoTA baseline by >3-4\% margin (absolute). $^{**}$ indicates those approaches leverage LLM-judge (oracle) feedback, although the LLM judges are similar but not exactly the same. $^{*}$ Numbers taken from \citep{li2024sketch2codeevaluatingvisionlanguagemodels}. N denotes the number of LLM calls for generating the HTML (>2000 tokens) and K represents the calls to LLM judge for getting feedback (<200 tokens). We have kept the judge-prompt similar to Sk2code \citep{li2024sketch2codeevaluatingvisionlanguagemodels}}
\label{tab:sketch2code_comparison}. 
\end{table}

\subsection{Sketch2code}
For Sketch2code \citep{li2024sketch2codeevaluatingvisionlanguagemodels}, we provide a detailed comparison of our approach against SoTA baselines on several evaluation criterion and metrics. We used the hyperparameter setting of temperature = 0.6, max tokens = 4096, top p = 1.0, frequency/repetition penalty = 0.0, and presence penalty = 0.0 for all our results. For the metrics, we consider metrics centring \textit{1.Layout Similarity, 2. Visual IoU, 3. Text IoU} with reference HTML following \citep{li2024sketch2codeevaluatingvisionlanguagemodels}. These metrics offer a comprehensive and reliable assessment of HTML generation quality, demonstrating a strong correlation (\~90\%) with human satisfaction, as shown in \citep{li2024sketch2codeevaluatingvisionlanguagemodels} (further details in Appendix). Hence, we use Layout similarity as a verifier along with LLM-as-judge \citep{li2024sketch2codeevaluatingvisionlanguagemodels} to guide the generations for both BON \citep{bon3} and IAD. We report a comparison with baseline single-turn approaches, including SotA models GPT-4o, Claude-3, InternVL2, Gemini-1.5-Flash, CoT, and variants, along with multi-response generation approaches, including BON, Sk2code, and IAD (Ours). \textit{Single turn} approaches even from SoTA models fail to match the layout structure, position of blocks, textual content, size of the blocks etc in the given user-sketch, causing a mismatch w.r.t to the reference layout as can be clearly seen in Figure and achieves a low score in-terms of all the three metrics in-comparison with multi-response generation approaches, even with N=2. Best-of-N sampling (BON) with a weaker model \textit{Gemini-1.5-Flash} improves over single-turn approaches and, with $N = 4$ generations, it outperforms SoTA models with single-turn responses by a margin of 15-18\%, by correctly identifying the block position, title block, overall layout structure, etc. We see monotonic improvement in performance over the number of responses as the layout score improves from $20.41$ to $25.7$ with 6 responses. However, BON struggles in incorporating fine-grained details about layout structure and makes some-times makes repetitive mistakes in the position of block in all the $N$ generation for the prompts (as shown in Fig). Our proposed approach IAD, mitigates this gap by iteratively improving the responses and as shown in Table 1, it achieves a major improvement of $~3-5\%$ (absolute improvement) from BON as well as single-turn SoTA Claude with just 2 iterations (eq : N=2) even with simpler model \textit{Gemini-1.5-Flash}. At each iteration, we pass the best and worst HTML as a context along with instructions for generating the next iteration. We observe IAD is able to learn fine-grained layout components, image semantics over iterations with the context of the Best and Worst HTML. We see that with increased iterations, the performance of IAD improves, reaching a very high layout score of 26.75, outperforming all baselines with the same generations. We also report the Image and Text IoU scores while optimizing with the layout-score, to check for reward-overoptimization of the metric. 

\begin{figure*}[!t]
    \centering
    \includegraphics[width=\textwidth]{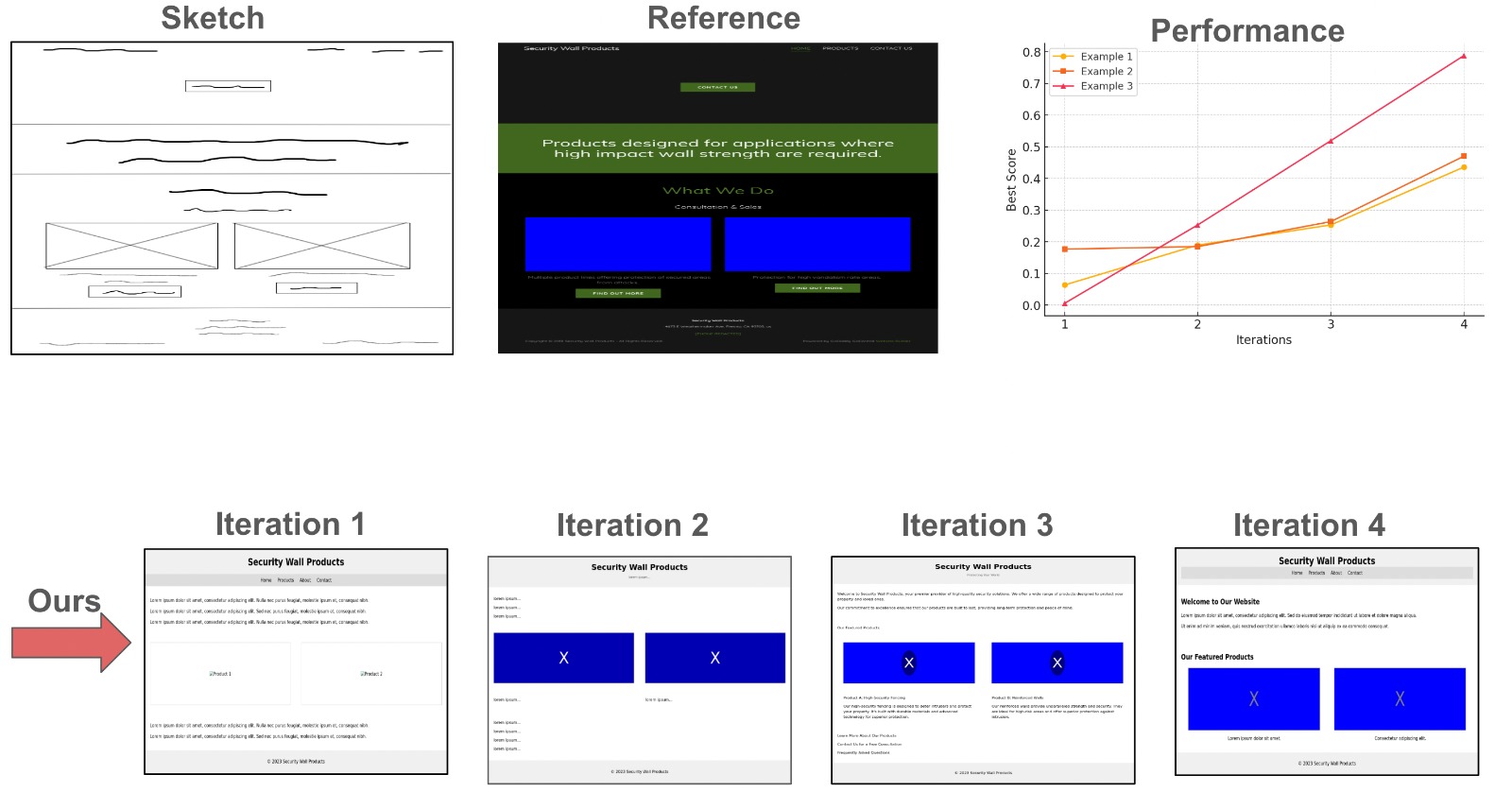}
     \caption{\textbf{Sketch2code} : Top row shows the user sketch, reference image and the performance of IAD over iterations. The figure highlights improvement of IAD over 4-turns w.r.t Layout similarity score (1/100) for 3 examples. It shows clear improvement over iterations. We also qualitatively analyse the snapshots of the HTMLs generated by the agent, which demonstrates that over iteration the qualitative performance improves and matches the input sketch/reference HTML.} 
    \label{fig:iad_qual}
\end{figure*}

\noindent
However, as can be observed in Table 1 and Figure-3, that text and image similarities are also improving over the iterations and our findings regarding comparison with baseline BON are consistent with the same. However, we observe that with increase number of generations the performance gets closer to BON. We also consider sensitivity of the token-length of the context plays a critical role in this case, where providing the entire HTMLs can affect the entropy of the distribution, and thus over-conditioning can hinder structured generation by reducing diversity and exploration (as shown in Figure ). Thus, we provide only the top 200-300 tokens of the best (and worst) HTMLs. However, it is clear that if there would be a judge to highlight which portion of the code needs to be updated that will be more targeted. Hence, we incorporate LLM-judge (Gemini-1.5-Pro) which has the reference policy and it checks with the current response and provide feedback on improvement and sometimes snippets of HTML as well (however, we restrict that to 100 tokens ~ 5-8\% of the original HTML). This leads to an additional improvement of ~2-4\% for the layout score with just two iteration and final score of ~28 with 6 iterations, demonstration the important of iterative approaches for agent performance. We have kept the judge-prompt similar as Sk2code \citep{li2024sketch2codeevaluatingvisionlanguagemodels} for fairness
in evaluation. We admit that its indeed possible to learn more specific prompts to construct a better judge via training/prompt-tuning, however thats outside the main scope of this work. Overall, in all our analyses, our findings remain consistent, where IAD outperforms baselines.

\begin{figure*}[t]
    \centering
    \begin{subfigure}[b]{0.30\textwidth}  
        \centering
        \scalebox{1.0}{\includegraphics[width=\textwidth]{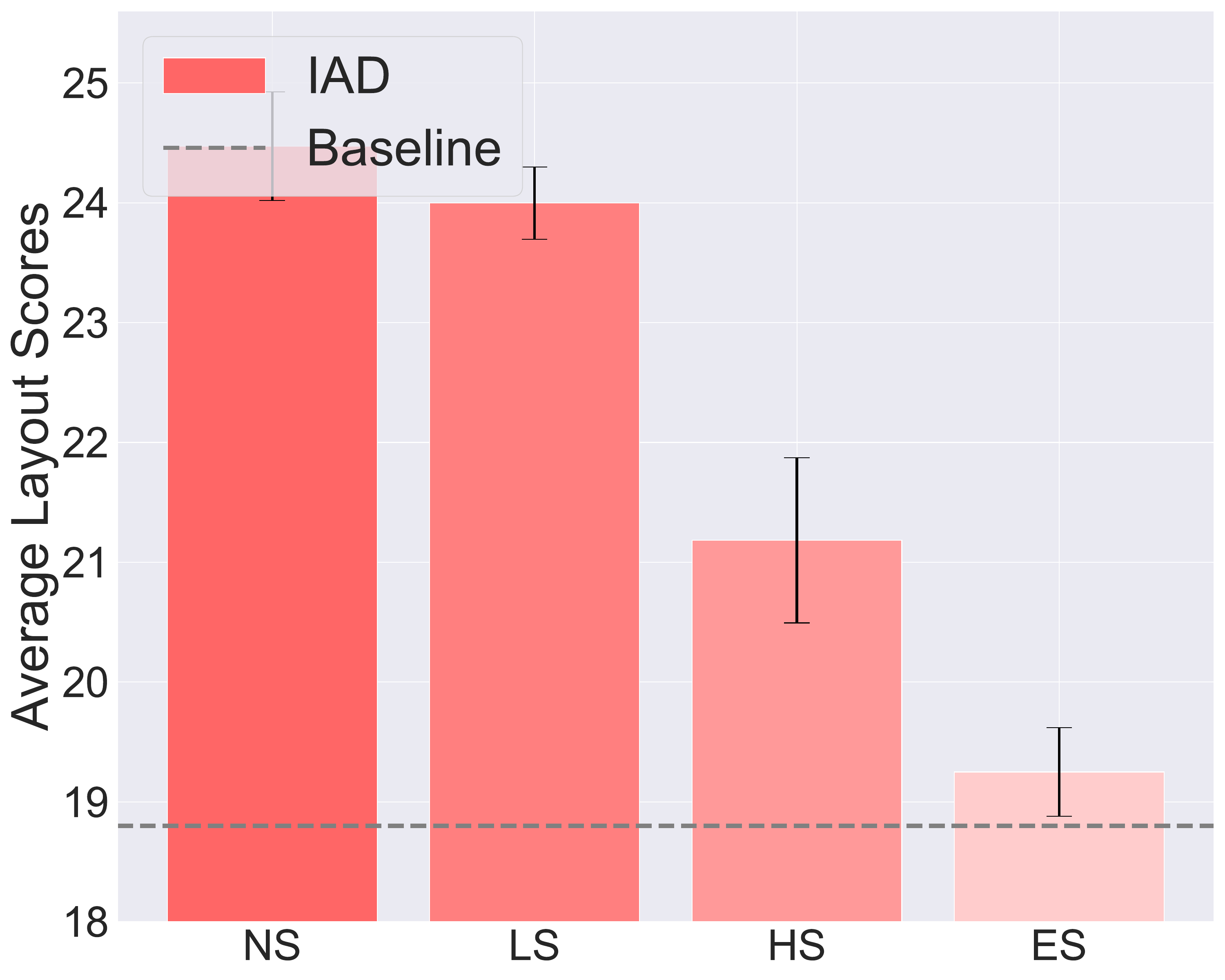}}
        \caption{\centering}
    \end{subfigure}
    \begin{subfigure}[b]{0.30\textwidth}  
        \centering
        \scalebox{1.0}{\includegraphics[width=\textwidth]{sparsity/iad_bon_sparse.pdf}}
        \caption{\centering}
    \end{subfigure}
    \hfill
    \begin{subfigure}[b]{0.35\textwidth}  
        \centering
        \scalebox{1.0}{\includegraphics[width=\textwidth]{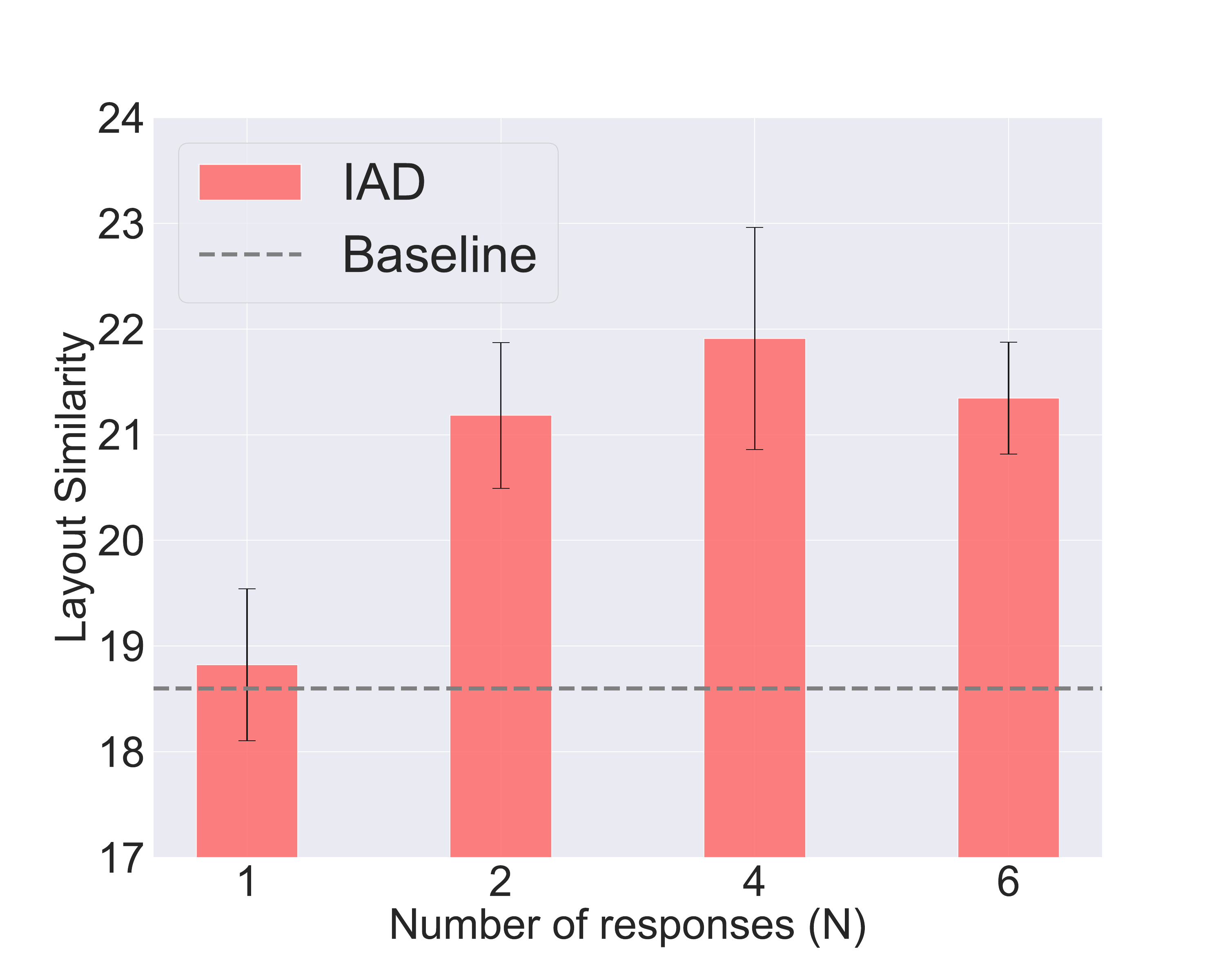}}
        \caption{\centering}
    \end{subfigure}
    \hfill
    \caption{\textbf{Sketch2code}: (a)Represents the performance of IAD w.r.t Layout score with varied sparsity NS (No Sparsity), LS (Low Sparsity) MS (Med Sparsity) and  HS (High Sparsity) (b) Performance comparison with BON via varying sparsity which shows the improvement over BON is less with sparsity (c) Performance of IAD w.r.t Layout similarity with Sparsity via across varying the number of generations.}
\end{figure*}

\subsection{Importance of Verifier and Reward functions}
In this section, we provide a motivation and importance of the verifier in ensuring monotonic improvement with our proposed approach. We define $\pi^*(\cdot|x)$ as the target policy generating $y^* \sim \pi^*(\cdot|x)$. At each step $t$, we sample $y_{t+1} \sim \pi_0(\cdot|x, \hat{y}_t)$, where $\hat{y}_t$ is the best response so far, and update $\hat{y}_{t+1} = \arg\max{y \in (y_t, \hat{y}_t)} R(x, y)$, accepting $y{t+1}$ if $R(x, y_t) - R(x, \hat{y}_t) > 0$. If $R(x, y)$ incorporates the information of $\pi^*(y|x)$ (upto a normalization i.e $R(x, y) = \frac{f(\pi^*(y|x))}{Z}$, $f$ being a monotonic function), we show that our iterative refinement never deteriorates performance. In other words, we assume that for any two responses $y_1, y_2$, if the reward function \citep{setlur2025scalingtesttimecomputeverification} satisfies $R(x, y_1) > R(x, y_2)$ then it implies that the optimal policy assigns a higher probability to $y_1$ than $y_2$, i.e $\pi^*(y_1|x) > \pi^*(y_2|x)$. 
A natural way to measure closeness to the optimal response is by estimating the distance under the true probability distribution (i.e target policy) $\pi^*(\cdot|x)$, defined as
\begin{align}
    d(\hat{y}_{t+1}, y^*) = \pi^*(y^*|x) - \pi^*(\hat{y}_{t+1}|x)
\end{align}
where the difference captures that how good the quality of the response is under optimal policy. If the response $\hat{y}_{t+1}$ is highly optimal, then $d(\hat{y}_{t+1}, y^*)$ will be low and viceversa, when $\hat{y}_{t+1} = y^*$, the gap will be zero.
\begin{align}
    d(\hat{y}_t, y^*) &= \pi^*(y^*|x) - \pi^*(\hat{y}_t|x) \\ \nonumber
    & = \pi^*(y^*|x) - \pi^*(\hat{y}_{t}|x) -(\pi^*(\hat{y}_{t+1}|x) -\pi^*(\hat{y}_{t}|x)) \\ \nonumber
    & \leq \pi^*(y^*|x) - \pi^*(\hat{y}_{t}|x) = d(\hat{y}_{t}, y^*)
\end{align}
where, we first add and subtract the term $\pi^*(\hat{y}_{t}|x)$. Then by definition of our acceptance rule, we ensure that $\pi^*(\hat{y}_{t+1}|x) -\pi^*(\hat{y}_{t}|x) \geq 0$, where equality occurs when  $\hat{y}_{t+1} = \hat{y}_{t}$. Thus we have $d(\hat{y}_t) <= d(\hat{y}_{t-1})$ i.e we ensure that the responses over the iteration are either improving or remain the same over iteration and won't deteriorate over iterations. 
However, it is important to note that this is based on the assumption that the reward function is aligned with the optimal distribution, meaning that selecting responses based on maximizing $R(x,y)$ leads to responses that are increasingly closer to the ground-truth distribution $\pi^*(\cdot|x)$. This highlights the criticality and accuracy of a true verifier in these agentic tasks as also shown in \citep{setlur2025scalingtesttimecomputeverification}

\noindent
\textbf{Verifier and Reward function}: We provide qualitative evaluation of considering layout similarity as a verifier due to its Interpretability and also correlation with human judgements also shown in \citep{li2024sketch2codeevaluatingvisionlanguagemodels}. Additionally, we want to highlight that Sketch2code represents an extremely complex and challenging task for using self-LLM as a judge \citep{refine1} (without significant prompting) to compare between two generated HTMLs (by the agent) with its similarity to the input sketch and prompt. The input sketch has entirely different distribution than the image snapshot of the generated HTML which makes it harder for LLM as a judge to perform which is one of the reason we hypothesize that Self-refine \citep{refine1} type approaches doesn't provide improvements as shown in Table 1. On the other-hand, although LLM judge (oracle) provides more meaningful feedback when it has access to the reference HTML, however needs to be prompted efficiently to generate meaningful responses. We agree the fact that our judge (oracle) for the feedback was allowed to provide more context than the one used in \citep{li2024sketch2codeevaluatingvisionlanguagemodels}. However, the performance improvement in \citep{li2024sketch2codeevaluatingvisionlanguagemodels} feedback approch is very less and we hypothesize major reasons can be not performing IAD type approach, where we take previous best response (HTML) in the context along with specific instructions. Even for LLM-judge (oracle), we leverage feedback along with the previous best and worst HTMLs, which helps in providing more meaningful context to the agent in generating the correct HTML.

\begin{figure*}[!ht]
    \centering
    \includegraphics[width=\textwidth]{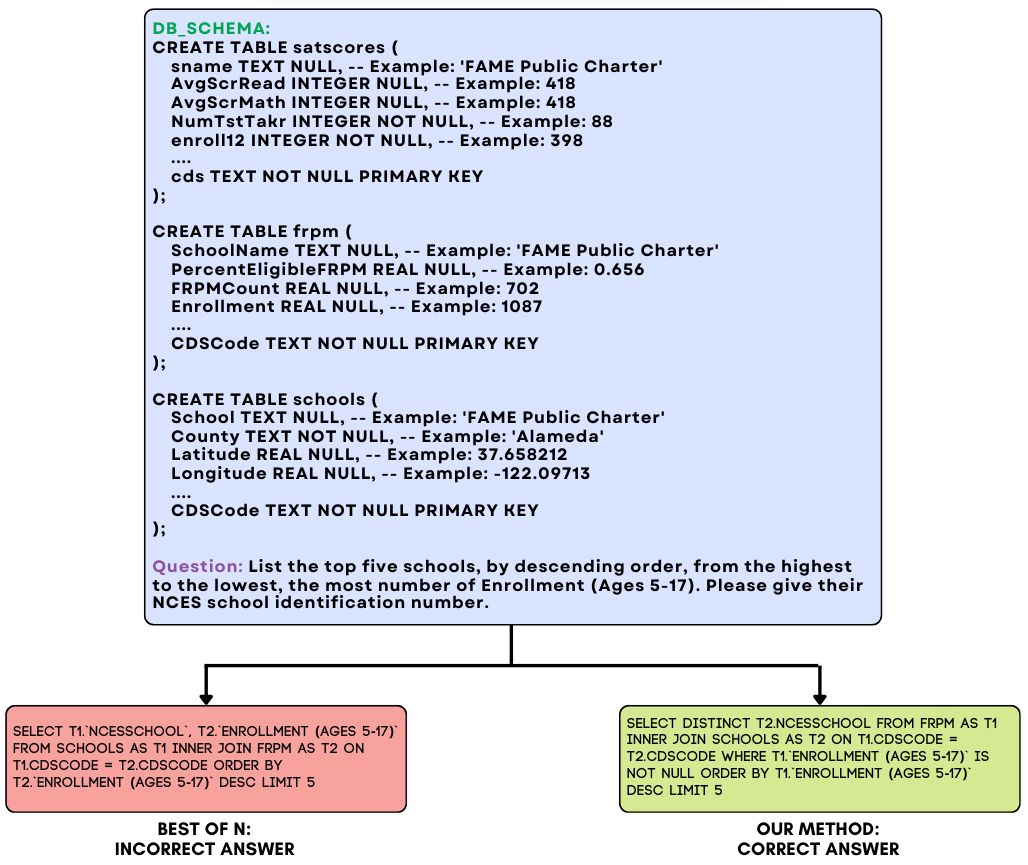}
     \caption{\textbf{Text2SQL}: An example of two responses is presented: the first response, generated using our proposed approach, is correct, while the second response, produced using the best-of-N method, is incorrect.} 
    \label{fig:visual_comparison_txtsql}
\end{figure*}

\begin{table}[h]
    \centering
    \renewcommand{\arraystretch}{1.2}
    \begin{tabular}{p{4cm} p{4cm} p{4cm} p{3cm}}
        \toprule
        \textbf{Query} & \textbf{Search Attempts} & \textbf{Results Found} & \textbf{Final Outcome} \\
        \midrule
        Men's Black Loafers (Size 10.5, Rubber Soles, \textless 60) & Multiple searches, clicked "Next" repeatedly, found unrelated shoes (sneakers, sandals, pumps) & None matched the requirement & Task Failed - No suitable options found (Reward: 0.0) \\
        \midrule
        Blue Diamond Almonds (Gluten-Free, Pecan, 12 Pack) & Repeated searches, encountered "No Search button" error multiple times, retrieved irrelevant snack items & Nut Thins Crackers, Keto Bars, M\&M’s Chocolate & Task Failed - No relevant product found (Reward: 0.0) \\
        \midrule
        Folding Storage Box Ottoman (Faux Leather, 60x40x40cm, \textless 170) & Initial product matched but had incorrect size, next searches returned irrelevant furniture items & Found an ottoman, but wrong size \& overpriced & Task Failed - No exact match found (Reward: 0.0) \\
        \midrule
        Official Cleveland University Drawstring Shorts (Small, Charcoal, Machine Washable, \textless 60) & Search led to incorrect results (Marvel T-Shirts, Women’s Yoga Shorts), agent attempted refinement but couldn't find exact product & No official Cleveland University shorts found & Task Failed - No suitable options found (Reward: 0.0) \\
        \midrule
        Organic Hair Growth Serum Roller Set (For All Hair Types, \textless 60) & Search retrieved some serums but none matched exact request (wrong quantity or expensive) & Found a set, but incorrect product version & Task Failed - No exact match found (Reward: 0.0) \\
        \bottomrule
    \end{tabular}
    \caption{\textbf{Webshop} : Highlights several Failure Cases of the Baseline Agent (Gemini-1.5-Pro) in Retrieving Relevant products given the task. This represents the challenge of current model in performing strategic exploration in Webshop.}
    \label{tab:failure_cases}
\end{table}

\begin{tcolorbox}[floatplacement=t, width=\columnwidth, colback=gray!10, colframe=gray, title=\textbf{Webshop - Task Execution Flow - IAD (Success)}]
    Search: \texttt{"blue color toothbrushes"}  
    \quad $\longrightarrow$ \quad Product List Found  
    \quad $\longrightarrow$ \quad Selected: \texttt{Hoomall Kids U-Shaped Toothbrush (Blue, \$10.95)}  
    \quad $\longrightarrow$ \quad Clicked on Product  
    \quad $\longrightarrow$ \quad Purchased  
    \quad $\longrightarrow$ \quad \textbf{Task Completed (Reward: 1.0)}
\end{tcolorbox}

\begin{tcolorbox}[colback=gray!10, colframe=black, title=\textbf{Webshop - Task: Buy a Folding Storage Box Ottoman- IAD (Success)}]
\textbf{Size:} 60x40x40cm \quad \textbf{Material:} Faux Leather \quad \textbf{Price:} Under \$170

\begin{itemize}
    \item \textbf{Search} $\rightarrow$ "folding storage box ottoman faux leather 60x40x40cm"
    \item \textbf{Product List} $\rightarrow$ Found 50 results
    \begin{itemize}
        \item Ottoman Footstool (40x40x40cm) - \$149.97
        \item Other options did not match size or price
    \end{itemize}
    \item \textbf{Click} $\rightarrow$ Select "Ottoman Footstool"
    \item \textbf{Size Selection} $\rightarrow$ Click "60x40x40cm"
    \item \textbf{Buy Now} $\rightarrow$ Proceed to checkout
    \item \textbf{Task Completed}
\end{itemize}
\end{tcolorbox}

\begin{tcolorbox}[colback=gray!10, colframe=black, title=\textbf{Webshop - Task: Buy a Vegan, Gluten-Free Protein Shake - IAD (Success)}]
\textbf{Requirements:} 100\% Vegan, Gluten-Free, Soy-Free \quad \textbf{Price:} Under \$40

\begin{itemize}
    \item \textbf{Search} $\rightarrow$ "gluten free vegan plant based protein shake"
    \item \textbf{Product List} $\rightarrow$ Found 50 results
    \begin{itemize}
        \item OWYN Protein Shake (Cold Brew Coffee, 12oz) - \$11.07
        \item Other products exceeded price or dietary restrictions
    \end{itemize}
    \item \textbf{Click} $\rightarrow$ Select "OWYN Protein Shake"
    \item \textbf{Buy Now} $\rightarrow$ Proceed to checkout
    \item \textbf{Task Completed}
\end{itemize}
\end{tcolorbox}

\subsection{Text2SQL}
In this section we provide some example responses of Best of N approach and our proposed method on the BIRD development set samples.

\begin{tcolorbox}[floatplacement=t,float, width=\columnwidth, colback=gray!10, colframe=gray, title=\textbf{Text2SQL : Generated queries for best of N approach (Wrong answer)}]
    \textbf{DB schema:} \\ CREATE TABLE satscores ( \\
        enroll12 INTEGER not null, \\
        dname TEXT null, \\
        ...\\
);\\
CREATE TABLE frpm(\\
        IRC INTEGER null, \\  
        `Free Meal Count (K-12)` REAL null, \\
        ...\\
);\\
CREATE TABLE schools(\\
        GSserved TEXT null,\\
        Street TEXT null, \\
        ...  \\
);\\
\textbf{Question:} Which active district has the highest average score in Reading?\\
\textbf{Predicted Query:} SELECT T1.District FROM schools AS T1 INNER JOIN satscores AS T2 ON T1.CDSCode = T2.cds WHERE T1.StatusType = 'Active' GROUP BY T1.District ORDER BY AVG(T2.AvgScrRead) DESC LIMIT 1
\end{tcolorbox}

\begin{tcolorbox}[floatplacement=t,float, width=\columnwidth, colback=gray!10, colframe=gray, title=\textbf{Text2SQL : Generated queries for our method (Correct answer)}]
    \textbf{DB schema:} \\ CREATE TABLE satscores ( \\
        enroll12 INTEGER not null, \\
        dname TEXT null, \\
        ...\\
);\\
CREATE TABLE frpm(\\
        IRC INTEGER null, \\  
        `Free Meal Count (K-12)` REAL null, \\
        ...\\
);\\
CREATE TABLE schools(\\
        GSserved TEXT null,\\
        Street TEXT null, \\
        ...  \\
);\\
\textbf{Question:} Which active district has the highest average score in Reading?\\
\textbf{Predicted Query:} SELECT T1.dname FROM satscores AS T1 INNER JOIN schools AS T2 ON T1.cds = T2.CDSCode WHERE T2.StatusType = 'Active' AND T1.AvgScrRead IS NOT NULL GROUP BY T1.dname ORDER BY MAX(T1.AvgScrRead) DESC LIMIT 1
\end{tcolorbox}

\newpage

\begin{tcolorbox}[floatplacement=t,float, width=\columnwidth, colback=gray!10, colframe=gray, title=\textbf{Text2SQL : Generated queries for best of N approach (Wrong answer)}]
    \textbf{DB schema:} \\ CREATE TABLE satscores ( \\
        enroll12 INTEGER not null, \\
        dname TEXT null, \\
        ...\\
);\\
CREATE TABLE frpm(\\
        IRC INTEGER null, \\  
        `Free Meal Count (K-12)` REAL null, \\
        ...\\
);\\
CREATE TABLE schools(\\
        GSserved TEXT null,\\
        Street TEXT null, \\
        ...  \\
);\\
\textbf{Question:} Among the schools with the average score in Math over 560 in the SAT test, how many schools are directly charter-funded?\\
\textbf{Predicted Query:} SELECT COUNT(T1.CDSCode) FROM schools AS T1 INNER JOIN satscores AS T2 ON T1.CDSCode = T2.CDSCode WHERE T2.AvgScrMath > 560 AND T1.FundingType = 'Directly funded'
\end{tcolorbox}

\begin{tcolorbox}[floatplacement=t,float, width=\columnwidth, colback=gray!10, colframe=gray, title=\textbf{Text2SQL : Generated queries for our method (Correct answer)}]
    \textbf{DB schema:} \\ CREATE TABLE satscores ( \\
        enroll12 INTEGER not null, \\
        dname TEXT null, \\
        ...\\
);\\
CREATE TABLE frpm(\\
        IRC INTEGER null, \\  
        `Free Meal Count (K-12)` REAL null, \\
        ...\\
);\\
CREATE TABLE schools(\\
        GSserved TEXT null,\\
        Street TEXT null, \\
        ...  \\
);\\
\textbf{Question:}Among the schools with the average score in Math over 560 in the SAT test, how many schools are directly charter-funded?\\
\textbf{Predicted Query:} SELECT COUNT(DISTINCT T1.CDSCode) FROM schools AS T1 INNER JOIN satscores AS T2 ON T1.CDSCode = T2.cds INNER JOIN frpm AS T3 ON T3.CDSCode = T2.cds WHERE T2.AvgScrMath > 560 AND T3.`Charter Funding Type` = 'Directly funded' AND T3.`Charter Funding Type` IS NOT NULL
\end{tcolorbox}
\end{document}